\documentclass{article}

\newif\iffinal
\finaltrue

\iffinal
    \usepackage[final, nonatbib]{corl_2019} % Uncomment for the camera-ready ``final'' version
\else
    \usepackage[nonatbib]{corl_2019} % initial submission
\fi

\usepackage[table]{xcolor}% http://ctan.org/pkg/xcolor
\usepackage{tikz,array,collcell}
\usetikzlibrary{calc}
\pgfdeclarelayer{background}
\pgfdeclarelayer{main}
\pgfsetlayers{background,main}

\newcommand*{\maxValue}{1}
\newcommand{\tikzMeL}[1]{\tikzMe{#1}{left}}
\newcommand{\tikzMeC}[1]{\tikzMe{#1}{center}}
\newcommand{\tikzMeR}[1]{\tikzMe{#1}{right}}
\newcommand{\tikzMe}[2]{%
  \tikz[baseline]{
    \node[anchor=base, align=#2, inner xsep=\tabcolsep] (n) {#1};
    \begin{pgfonlayer}{background}
       \pgfmathparse{abs(#1/\maxValue)}
       \fill[color=green!25] (n.north west) rectangle ($(n.south west)!\pgfmathresult!(n.south east)$);
    \end{pgfonlayer}
  }
}
\newcolumntype{Q}{@{}>{\collectcell\tikzMeL}l<{\endcollectcell}@{}}
\newcolumntype{W}{@{}>{\collectcell\tikzMeC}c<{\endcollectcell}@{}}
\newcolumntype{E}{@{}>{\collectcell\tikzMeR}r<{\endcollectcell}@{}}

\usepackage[utf8]{inputenc} % allow utf-8 input
\usepackage[T1]{fontenc}    % use 8-bit T1 fonts
\usepackage{hyperref}       % hyperlinks
\usepackage{url}            % simple URL typesetting
\usepackage{booktabs}       % professional-quality tables
\usepackage{amsfonts}       % blackboard math symbols
\usepackage{nicefrac}       % compact symbols for 1/2, etc.
\usepackage{microtype}      % microtypography
\usepackage{subcaption}
\usepackage{longtable}

\newcommand{\supplementary}[1]{
\par\vspace{6pt}\noindent{\fontsize{9}{9}\selectfont\textbf{Supplementary Materials:} 
\iffinal {#1} \else \textit{Anonymised} \fi \par}}
\renewcommand{\acknowledgments}[1]{
\vspace{6pt}\noindent{\fontsize{9}{9}\selectfont\textbf{Acknowledgements:} 
\iffinal {#1} \else \textit{Anonymised} \fi \par}}
\newcommand{\authorcontributions}[1]{%
\vspace{6pt}\noindent{\fontsize{9}{9}\selectfont\textbf{Author Contributions:} 
\iffinal {#1} \else \textit{Anonymised} \fi \par}}
\newcommand{\funding}[1]{
\vspace{6pt}\noindent{\fontsize{9}{9}\selectfont\textbf{Funding:} 
\iffinal {#1} \else \textit{Anonymised} \fi \par}}
\newcommand{\conflictsofinterest}[1]{%
\vspace{6pt}\noindent{\fontsize{9}{9}\selectfont\textbf{Conflicts of Interest:} {#1}\par}}

\usepackage{booktabs}
\usepackage{array}
\newcolumntype{L}[1]{>{\raggedright\let\newline\\\arraybackslash\hspace{0pt}}m{#1}}
\newcolumntype{C}[1]{>{\centering\let\newline\\\arraybackslash\hspace{0pt}}m{#1}}
\newcolumntype{R}[1]{>{\raggedleft\let\newline\\\arraybackslash\hspace{0pt}}m{#1}}
\usepackage{multicol}
\usepackage{multirow}

\usepackage[
  backend=bibtex,
  sorting=none,%ydnt,
  citestyle=numeric%numeric-comp%verbose%-tcomp
  ]{biblatex}
\bibliography{bibliography}

\usepackage[nonumberlist,nogroupskip]{glossaries}
\setglossarypreamble{We present the terms and notations used in this literature review as needed. For completeness, we provide a list of the most used terms and notations here.}

 \newacronym{2d}{2D}{2--dimensional}
 \newacronym{3d}{3D}{3--dimensional}
 \newacronym{ann}{ANN}{artificial neural network}
 \newacronym{ai}{AI}{artificial intelligence}
 \newacronym{acktr}{ACKTR}{actor-critic using Kronecker-factored trust region}
 \newacronym{acer}{ACER}{actor-critic with experience replay}
 \newacronym{a2c}{A2C}{advantage actor-critic}
 \newacronym{a3c}{A3C}{asynchronous advantage actor-critic}
 \newacronym{cnn}{CNN}{convolutional neural network}
 \newacronym{cas}{CAS}{continuous action space}
 \newacronym{ci}{CI}{confidence interval}
 \newacronym{dnn}{DNN}{deep neural network}
 \newacronym{dl}{DL}{deep learning}
 \newacronym{drl}{DRL}{deep reinforcement learning}
 \newacronym{dti}{DTI}{Danish Technological Institute}
 \newglossaryentry{dof}
{
  name={dof},
  description={degree of freedom},
  first={\glsentrydesc{dof} (\glsentrytext{dof})},
  plural={DoFs},
  firstplural={degrees of freedom (\glsentryplural{dof})}
}

 \newacronym{dqn}{DQN}{deep Q-network}
 \newacronym{ddpg}{DDPG}{deep deterministic policy gradients}
 \newacronym{edf}{EDF}{empirical distribution function}
 \newacronym{ga}{GA}{genetic algorithm}
 \newacronym{hrl}{HRL}{hierarchical reinforcement learning}
 \newacronym{irl}{IRL}{inverse reinforcement learning}
 \newacronym{kl}{KL}{Kullback–Leibler}
 \newacronym{ks}{KS}{Kolmogorow-Smirnow}
 \newacronym{lr}{LR}{learning rate}
 \newacronym{ml}{ML}{machine learning}
 \newacronym{mdp}{MDP}{Markov decision process}
 \newacronym{nn}{NN}{neural network}
 \newacronym{ppo}{PPO}{proximal policy optimization}
 \newacronym{pg}{PG}{policy gradient}
 \newacronym{rl}{RL}{reinforcement learning}
 \newacronym{rnn}{RNN}{recurrent neural network}
 \newacronym{soft-q}{soft-Q}{soft Q-learning}
 \newacronym{smbo}{SMBO}{sequential model-based optimisation}
 \newacronym{se}{SE}{standard error}
 \newacronym{tpe}{TPE}{tree-structured parzen estimator}
 \newacronym{trpo}{TRPO}{trust region policy optimization}
 \newacronym{ur}{UR}{Universal Robots}
\usepackage[super]{nth}

\usepackage[page]{appendix}

\usepackage{csquotes}

\usepackage{amsfonts}

\usepackage{comment}
\usepackage{todonotes}

\usepackage{listings}
\newcommand{\etal}{\emph{et al}.\ }

\title{A Survey on Reproducibility by Evaluating Deep Reinforcement Learning Algorithms on Real-World Robots}

\author{
  Nicolai A.~Lynnerup$^{1,2,\dagger, *}$\\
  \texttt{nily@dti.dk} \\
  \texttt{nia@mmmi.sdu.dk} \\
  \And
  Laura~Nolling$^{1,2,*}$\\
  \texttt{lauj@dti.dk} \\
  \texttt{lnj@mmmi.sdu.dk} \\
  \AND
  Rasmus Hasle$^{1}$\\
  \texttt{raha@dti.dk} \\
  \And
  John Hallam$^{2}$\\
  \texttt{john@mmmi.sdu.dk} \\
  \AND
  {\normalfont $^{1}$Robot Technology, Danish Technological Institute (DTI) }\\
  {\normalfont $^{2}$Embodied Systems for Robot Learning, University of Southern Denmark (SDU) }\\
  {\normalfont $^{\dagger}$Correspondence: \texttt{nily@dti.dk}; Tel.: +45-7220 2713}\\
  {\normalfont $^{*}$Equal contributions}
}

\hypersetup{draft}
\begin{document}
\maketitle

%===============================================================================

\begin{abstract}
  As \gls*{rl} achieves more success in solving complex tasks, more care is needed to ensure that \gls*{rl} research is reproducible and that algorithms therein can be compared easily and fairly with minimal bias. \Gls*{rl} results are, however, notoriously hard to reproduce due to the algorithms' intrinsic variance, the environments' stochasticity, and numerous (potentially unreported) hyper-parameters. In this work we investigate the many issues leading to irreproducible research and how to manage those. We further show how to utilise a rigorous and standardised evaluation approach for easing the process of documentation, evaluation and fair comparison of different algorithms, where we emphasise the importance of choosing the right measurement metrics and conducting proper statistics on the results, for unbiased reporting of the results.
\end{abstract}

\keywords{CoRL, Robots, Learning, Reinforcement Learning, Reproducibility, Statistics} 

%===============================================================================

\section{Introduction}
The ability critically to assess and evaluate the claims made by other scientists in published research is fundamental and a cornerstone of science. The impartial and independent verification of others' research serves the purpose of credibility-confirmation and allows for building on top of a ``body of knowledge,'' referred to as \textit{extensible research}. Research in robotics and \gls*{ml} is not excluded from this strict scientific requirement, even though it is notoriously hard to ensure reproducibility in computational studies of this nature \cite{sandve2013ten}. 

In the domain of robotic \gls*{rl}, algorithms such as \gls*{trpo} \cite{schulman2015trust}, \gls*{ppo} \cite{schulman2017proximal}, \gls*{ddpg} \cite{lillicrap2015continuous} and Soft Q-Learning (Soft-Q) \cite{haarnoja2017reinforcement} have gained popularity due to their success in simulated robotic tasks \cite{duan2016benchmarking}; but as Mahmood \etal \cite{mahmood2018setting, mahmood2018benchmarking} shows, setting up tasks and evaluating \gls*{rl} algorithms on real-world robots is seldom straightforward and requires many practical considerations in order to ensure reproducibility.

One of the most common issues when replicating \gls*{ml} research is omission of one or more hyper-parameter choices in the manuscript. Often hyper-parameters have significant impacts on how the algorithm performs so it is critical to report their values including how they were obtained \cite{henderson2018deep,islam2017reproducibility}. One reason for neglecting to report hyper-parameters is that they are simply forgotten, which may be due to multiple reasons, e.g.; a) they are not considered important or b) their value is simply the default value, specified by the underlying implementation used. Both reasons are obviously important challenges to handle but are often hard to discover and subsequently enforce.

To add to the complexity, real-world robotic \gls*{rl} methods require large amounts of data. Results from the real world are thus expensive to obtain. Further, many industrial researchers are forced by their company's legal department to omit specific details to remain in front of their competitors. In combination with the ``publish or perish'' pressure on academic researchers, this seems to result in deviations from the standards of good science \cite{henderson2018deep}. To maintain progress in \gls{drl}, research must be reproducible and comparable so that improvements can be verified and built upon.

Differences in evaluation metrics and the lack of significance testing in the field of \gls{drl} potentially cause misleading reporting of results \cite{khetarpal2018re}. With no statistical evaluation of the results, it is difficult to conclude if there are meaningful improvements. If results are to be trusted, complete and statistically correct evaluations of proposed methods are needed.

We find inspiration from a pipeline proposed by Khetarpal \etal \cite{khetarpal2018re}, which provides common interfaces for both environments, algorithms, and evaluation schemes. We build on this idea by adding \textit{experiment configuration files}, to create a unified experiment framework. The configuration file contains all (hyper-)parameters related to specific experiments. With the configuration files, we aim to ease the process of a) running new experiments with new hyper-parameters during the tuning process, b) keeping track of past experiments and their hyper-parameters, and c) reporting all the hyper-parameters in the scientific paper. 

We evaluate our pipeline on the SenseAct framework\footnote{\url{https://github.com/kindredresearch/SenseAct/}} which provides an interface letting \gls*{rl} agents interact with the real world through \gls*{ur}' 6 DoF robotic manipulators \cite{mahmood2018benchmarking}. We utilise the proposed benchmarking task \textit{UR-Reacher-2D} in which the agent is to reach arbitrarily chosen target points in a 2D plane with its wrist joint (end-effector).

\textbf{Our key contributions} to the field are the demonstration of a rigorous method for easy documentation of parameters; reproducing \gls*{rl} results and determining significance by a statistics-based evaluation of common \gls*{rl} baseline algorithms on real-world robots; and our suggestions on ways to ensure correct choices of measurement metrics based on the task at hand.

%===============================================================================

\section{A Reproducibility Taxonomy}
The taxonomy of reproducible research is widely discussed \cite{plesser2018reproducibility, barba2018terminologies}. Here we present the terminology that we conform to:

    \paragraph{Repeatability} (\textit{same} team, \textit{same} experimental setup) refers to the same team with the same experimental setup re-running the experiment. This procedure is needed when wanting to report statistically sound results. The procedure implies the exact same team and the same code.
    
    \paragraph{Reproducibility} (\textit{different} team, \textit{same} experimental setup) refers to a different team conducting the same experiment with the same setup, achieving results within marginals of experimental error. The setup includes both library code, experimental code, data and environments. This procedure can be viewed as software testing at the level of a complete study.
    
    \paragraph{Replicability} (\textit{different} team, \textit{different} experimental setup) refers to teams, attempting to obtain the same (or similar enough) results as reported in the original work, who do not have access to either code, data, environment or all of them.

Note that the Claerbout, Donoho, Peng \cite{plesser2018reproducibility, claerbout1992electronic, buckheit1995wavelab, peng2006reproducible} convention omits the \textit{repeatability} term, but as we show in appendix \ref{app:taxonomy}, the ACM, Drummond convention's take on this term is applicable, as it is not contradictory. In our work we conform to the Claerbout, Donoho, Peng Convention \cite{claerbout1992electronic, buckheit1995wavelab, peng2006reproducible} and add the \textit{repeatability} term.

In addition to the taxonomy, we present a practical view of reproducibility in appendix \ref{app:practical_reproducibility}.

%===============================================================================

\section{Methods}
We propose a simple method for uniformly \textit{configuring} \gls*{rl} experiments by collecting all \mbox{(hyper-)parameters} in a configuration file using an open data format, in our case YAML. These configuration files contain the comprehensive list of parameters used for the specific experiments ranging from environment and agent specifics, such as robot kinematics, to algorithm specifics such as number of hidden layers in the function approximator\footnote{For a complete example of a configuration file, see: {\iffinal \url{https://github.com/dti-research/SenseActExperiments/blob/master/code/experiments/ur5/trpo_kindred_example.yaml} \else \textit{URL anonymised} \fi}}.

We additionally separate the algorithms from the environments and metric collection routines to ease the evaluation of additional algorithms, as proposed by Khetarpal \etal \cite{khetarpal2018re}.

In practice, the configuration files contain module specifications for each of the aforementioned, meaning that changing out e.g. the algorithm can be done directly by changing the module path in the YAML file. The same applies to environments and logging routines.

\subsection{Reporting the Evaluation}
The most common way of reporting results obtained by \gls*{rl} is to present a plot of the average cumulative reward (average returns). However, the performance of an algorithm can vary to a great extent due to the stochasticity of the algorithms and environments, and the average returns alone will not depict an algorithm’s range of performance. Instead, a proper evaluation requires multiple runs with different preset randomisation seeds \cite{henderson2018deep}. Further, proper statistics are needed to determine if a higher return in fact does represent better performance, such as \glspl*{ci} on the mean or probability values of obtaining a certain threshold performance value. 

In general, the more trials, the easier it will be to support the conclusions with the proper statistics. However, as in life sciences, robotics suffers from the fact that samples are expensive to obtain so cost-effective methods such as \textit{bootstrapping} are of particular interest. Bootstrapping is a method used for estimating a population distribution from a small sample by sampling with replacement.  The empirical distribution obtained by bootstrapping allows for statistical inference. Utilising this can further verify that no errors have occurred across trials. We conduct multiple trials to obtain a statistical significance measure of our results.

In the context of reporting the evaluation, we acknowledge that some cases are so time-consuming that multiple runs are not an option, not even for the small number of samples needed for bootstrapping. We encourage researchers who find themselves in such cases to state why they could not conduct proper statistics.

%===============================================================================

\section{Experimental Protocol}\label{sec:experimental_setup}

%===============================================================================================

\subsection{Task Description}
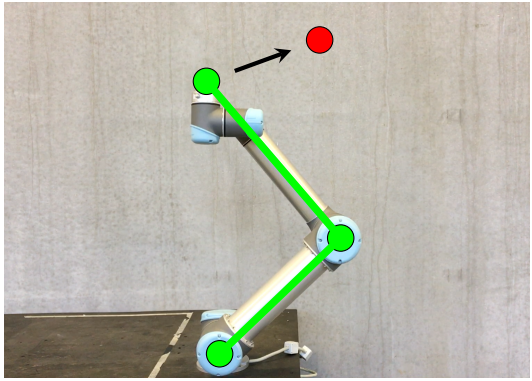
\begin{figure}
    \centering
    \begin{tikzpicture}
        \node[anchor=south west,inner sep=0] (image) at (0,0) {\includegraphics[width=0.5\textwidth,trim={5cm 0 5cm 0},clip]{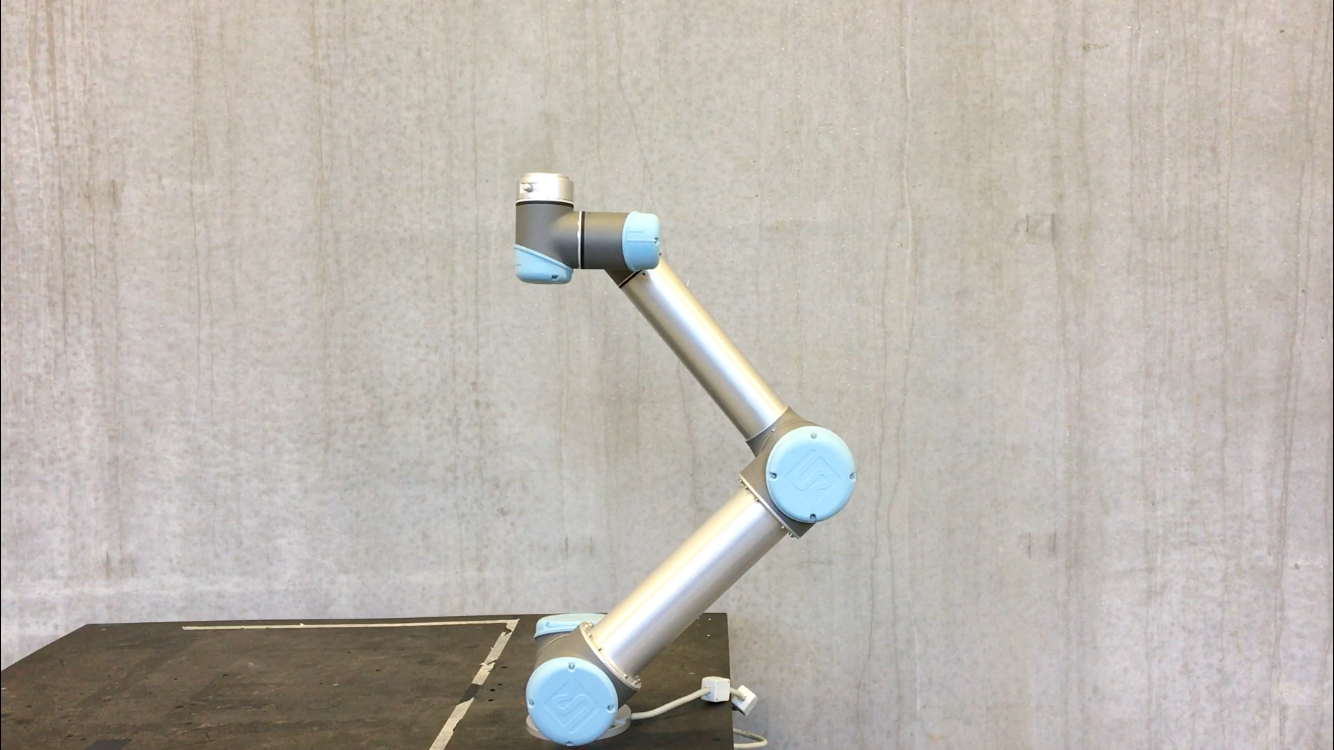}};
        \begin{scope}[x={(image.south east)},y={(image.north west)}]
            % inside the scope, (0,0) is at the lower left of the picture and (1,1) is at the upper right
            \draw[black, fill=red] (0.6,0.9) circle (5pt);
            \draw[black, fill=green] (0.41,0.065) circle (5pt);
            \draw[black, fill=green] (0.64,0.373) circle (5pt);
            \draw[black, fill=green] (0.385,0.79) circle (5pt);
            \draw[green, line width=1mm] (0.41,0.065) -- (0.64,0.373);
            \draw[green, line width=1mm] (0.64,0.373) -- (0.385,0.79);
            \draw [shorten >=4mm,shorten <=4mm,->, >=stealth, ultra thick] (0.385,0.79) -- (0.6,0.9);
        \end{scope}
    \end{tikzpicture}
    \caption{
        \textbf{Robot used for evaluation:}
        The 6 DOF robotic manipulator used for evaluating our proposed methodology is the UR5 robot. For the UR-Reacher-2D task, the robot is limited to actuate its second and third joints to reach the arbitrary points (red circle) with its tool.
    }\label{fig:ur5}
\end{figure}

To evaluate our proposed method we use the task by Mahmood \etal \cite{mahmood2018setting}, the \textit{UR-Reacher-2D}. In this task, the agent's objective is to reach arbitrary target positions by low-level control where the real-world UR5 (figure \ref{fig:ur5}) is restricted to only move its \nth{2} and \nth{3} axis. The reward function is defined as $R_{t}=-d_{t}+\exp\left(-100 d_{t}^{2}\right)$, where $d_t$ is the Euclidean distance between the point target and flange pose of the robot. The observation vector consists of the robot joint angles, joint velocities, its previous action, and the vector difference between the target and the flange coordinates. We keep the episodes to be 4 seconds, as in \cite{mahmood2018benchmarking}. A list of all hyper-parameter values used to conduct our evaluation is in appendix \ref{app:param_values}. 

%===============================================================================================

\subsection{RL Algorithms}
As in \cite{mahmood2018benchmarking} we use the OpenAI Baselines implementations for the two model-free policy-gradient algorithms \gls*{trpo} and \gls*{ppo}, to ensure that the code-base used is not a source of difference \cite{henderson2018deep}. It should be noted that the work by Mahmood \etal \cite{mahmood2018benchmarking} benchmarks two additional algorithms, \gls*{ddpg} \cite{lillicrap2015continuous} and Soft Q-learning \cite{haarnoja2017reinforcement}, which are not included in our work as they require inline modifications to the underlying code-bases (for extracting the evaluation metrics) which were not made publicly available by \cite{mahmood2018benchmarking} for legal reasons\footnote{Based on correspondence with the lead author}. The choice of \gls*{rl} algorithm has received little attention through this work, as it is beyond the scope of our study.

%===============================================================================================

\subsection{Evaluation}
We evaluate the \gls*{rl} algorithms on the UR-Reacher-2D task partly to investigate our proposed methodology, and partly the reproducibility of the original authors work \cite{mahmood2018benchmarking}.

We take the top 5 performing hyper-parameter configurations, shown in appendix \ref{app:param_values}, from the 30 randomly selected ones in \cite{mahmood2018benchmarking} for each of the two \gls*{rl} algorithms and evaluate their performance on the UR-Reacher-2D task.  We repeat each of the experiments (1 hyper-parameter configuration of 1 algorithm) 10 times to determine the statistical significance of the performance of each hyper-parameter configuration, which means running the experiments with different randomisation seeds that result in different network initialisations, target positions, and action selections. We approximate the \gls*{edf} using bootstrapping, to which we fit theoretical distributions. From the theoretical distributions we determine the probability of obtaining \textit{at least} the performance reported in the original work \cite{mahmood2018benchmarking}.  

\textbf{Evaluation metrics} are computed the same way as in \cite{mahmood2018benchmarking} to allow for comparison. The computations consist of a rolling average using a window size of $5,000$ steps, calculated every $1,000$ steps. The metrics are collected during training.

\textbf{Processing the results:} first, the average returns for each of the ten runs using the same configuration are calculated. Then we perform bootstrapping using 10k resamples on the ten average return values to find the \gls*{edf}. We then test for normality, which appears to be a common assumption in the field, and further explore a general approach for when normality cannot be assumed: fitting a theoretical distribution to the \gls*{edf}. In section \ref{sec:results} we show the results of fitting 100 theoretical distributions\footnote{For the comprehensive list see: \url{https://docs.scipy.org/doc/scipy/reference/stats.html}} to our \glspl*{edf}, while the plots are presented in appendix \ref{app:dist_fitting}.  Using a significance level of $\alpha = 0.05$, we determine if we successfully replicated the results reported in \cite{mahmood2018benchmarking}.

%===============================================================================

\section{Results}\label{sec:results}
\subsection{Repeatability of Code Base}
\begin{figure}
    \centering
    \begin{minipage}{0.49\textwidth}
        \centering
        \includegraphics[width=1.0\textwidth,trim={10pt 0 35pt 0},clip]{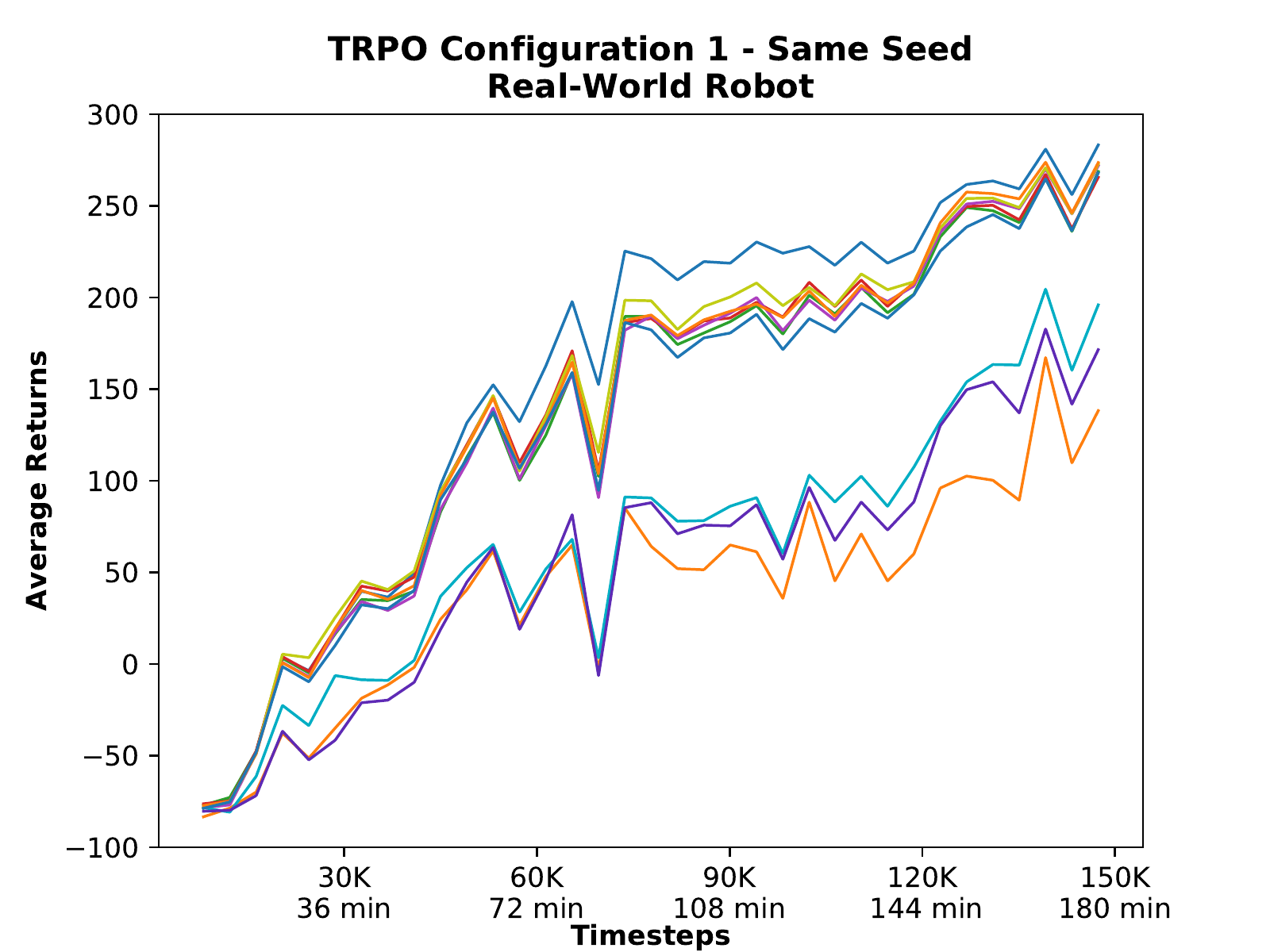}
    \end{minipage}\hfill
    \begin{minipage}{0.49\textwidth}
        \centering
        \includegraphics[width=1.0\textwidth,trim={10pt 0 35pt 0},clip]{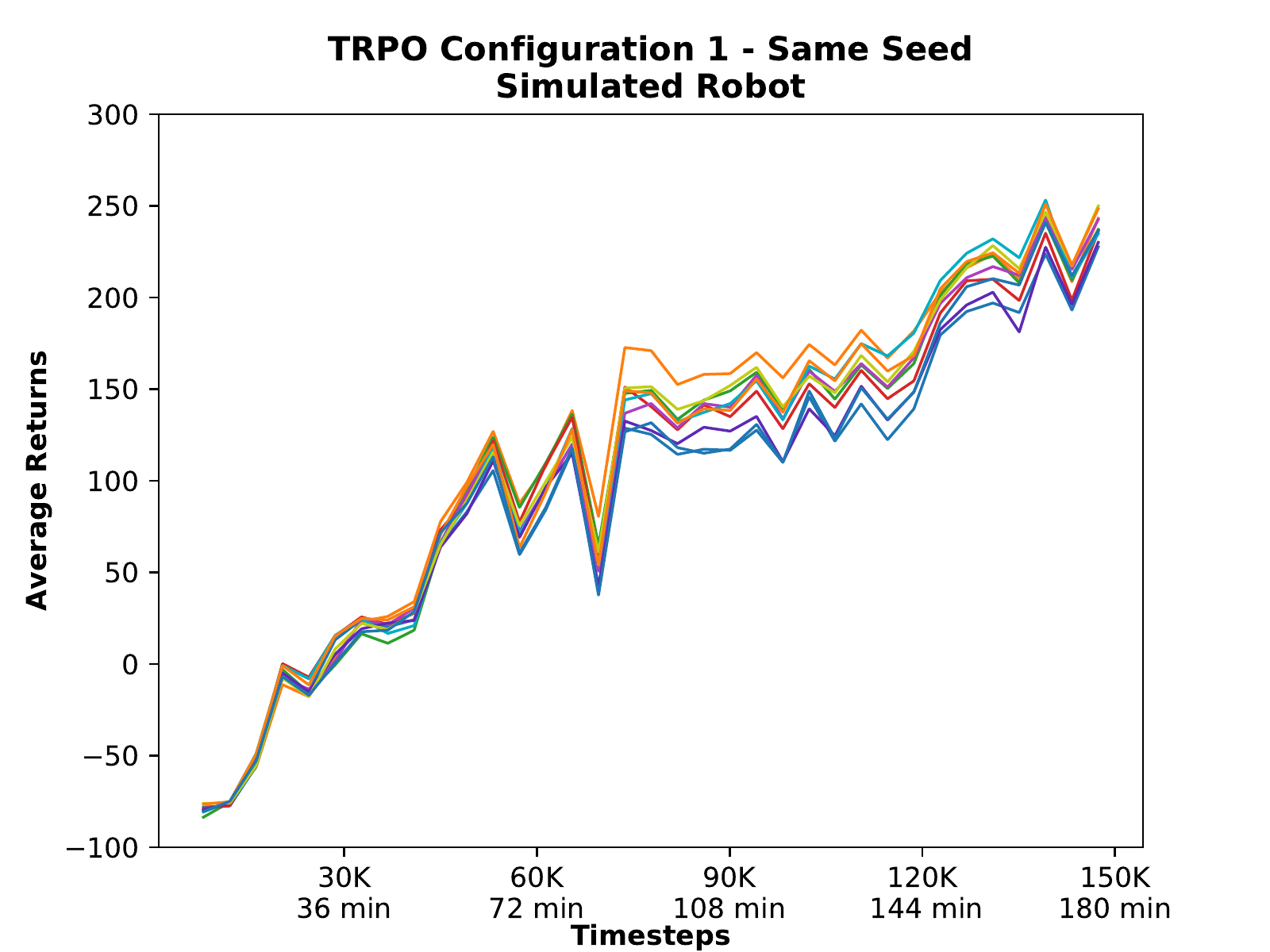}
    \end{minipage}
    \caption[footnotemark]{
        \textbf{Repeatability of learning:} The plots show the average return obtained over time during training, computed as a rolling average with a window size of 5.000 steps, calculated every 1.000 steps. 
        (\textit{left}) Ten runs of hyperparameter configuration 1 for \gls*{trpo} using same seed and evaluated on the real world robot, 
        (\textit{right}) Ten runs using the same \gls*{rl} algorithm with same code base using the same seed but evaluated in \gls*{ur}'s simulator: URSim v. 3.9.1\footnotemark.
    }\label{fig:repeatability}
\end{figure}

To verify that our changes to the code bases did not interfere with the ability to repeat experiments reported in \cite{mahmood2018benchmarking}, we performed ten runs of \gls*{trpo} on the real-world robot using the same seed and same experiment configuration. Initial results were promising but, after the first seven runs, something unknown happens and the last (bottom) three runs diverge from the seven previous. We speculate that the deviations on the real-world robot might suggest physical issues, such as delayed sensor readings, rather than stochasticity in the code-base. Therefore we decided to run the same experiment using the simulator to exclude the stochasticity of the real world. We can use the simulator provided by \gls*{ur} as it uses the exact same robot controller, code base, inverse kinematics solver, etc. We find that the resulting learning curves are in close proximity to one another. Thus we conclude that the ability to repeat experiments is upheld. The results are visualised in figure \ref{fig:repeatability}. 

%trim={<left> <lower> <right> <upper>}

\footnotetext{Available for download here: \url{https://www.universal-robots.com/download/?option=51823}}

\subsection{Evaluation of Baseline Algorithms}
\begin{figure}
    \centering
    \begin{minipage}{0.49\textwidth}
        \centering
        \includegraphics[width=1.0\textwidth,trim={10pt 0 35pt 0},clip]{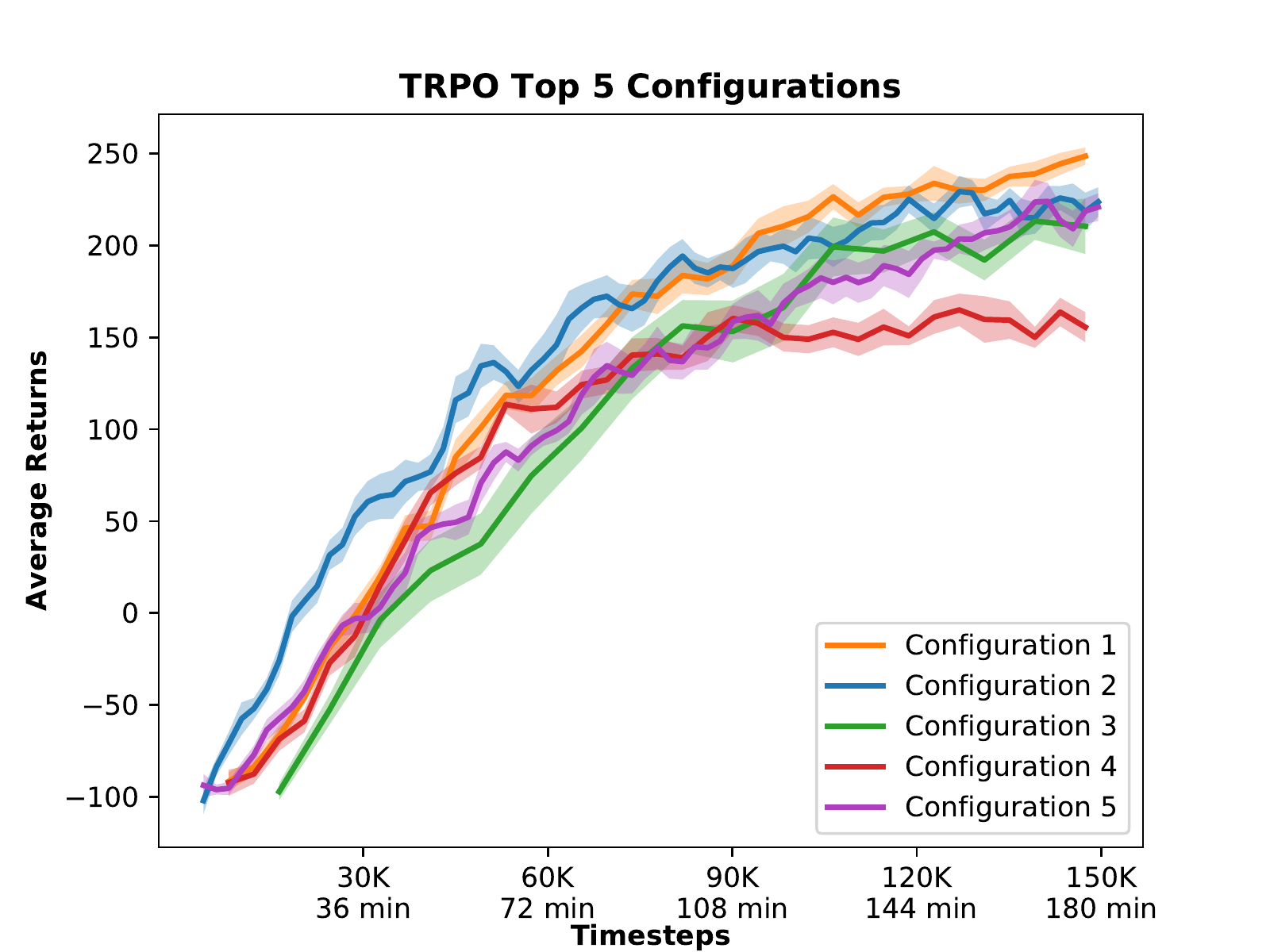}
    \end{minipage}\hfill
    \begin{minipage}{0.49\textwidth}
        \centering
        \includegraphics[width=1.0\textwidth,trim={10pt 0 35pt 0},clip]{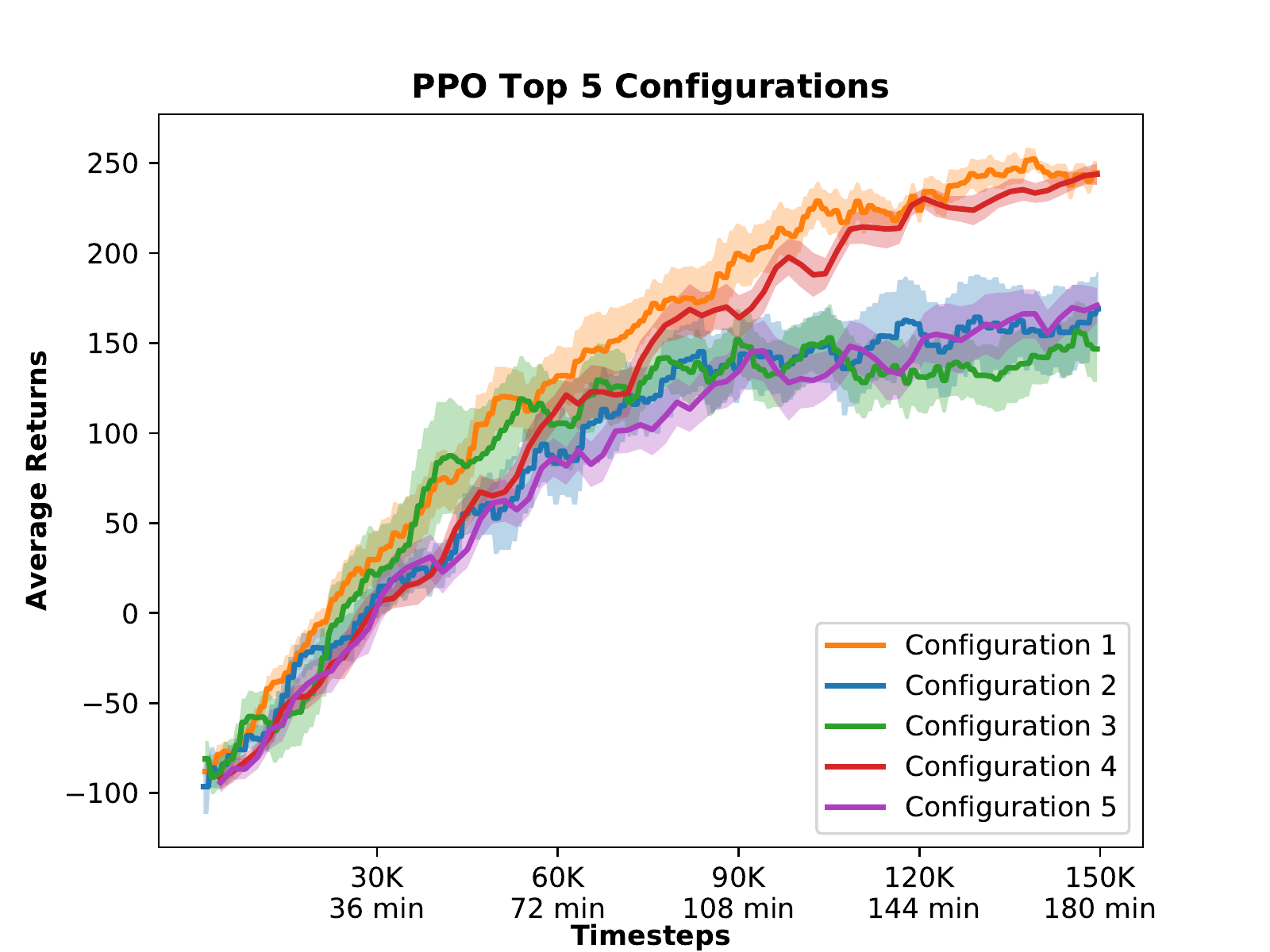}
    \end{minipage}
    \caption{
        \textbf{Top-5 hyperparameter configurations from the random search:} The mean average reward is plotted with its \gls*{se}, computed from the ten runs conducted for each of the five hyperparameter configurations for each of the two algorithms;  
        (\textit{left}) \gls*{trpo} and
        (\textit{right}) \gls*{ppo}. The average return is computed by a rolling average with a window size of 5000, and computed every 1.000 steps. 
    }\label{fig:top-5}
\end{figure}

\begin{figure}
    \centering
    \includegraphics[width=0.49\textwidth,trim={10pt 0 35pt 0},clip]{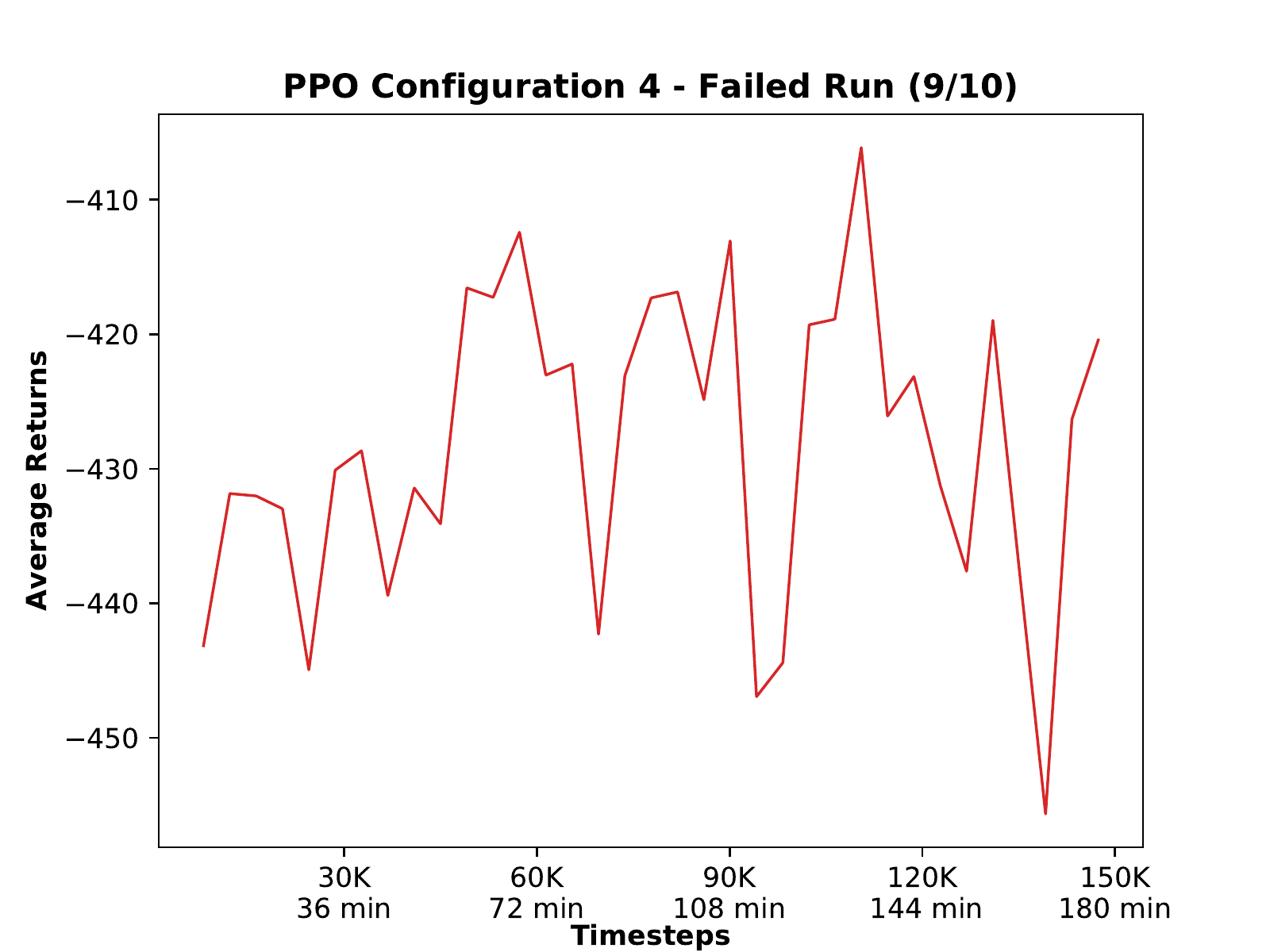}
    \caption{
        \textbf{The failed run of \gls*{ppo}:} The \nth{9} run of the \nth{4} configuration for \gls*{ppo} failed for some unknown reason. We assume that this run is an outlier and not part of the true population. Thus, we exclude it from the statistical analysis.
    }\label{fig:ppo_conf3_failed_run9}
    %\vspace{-1cm}
\end{figure}

The resulting learning curves from evaluating the top 5 configurations are plotted in figure \ref{fig:top-5}, which consists of the mean rewards and their \gls*{se}. Through the evaluation of \gls*{trpo}, we observed that the worst performing configuration was configuration 4, which is different from the original work \cite{mahmood2018benchmarking}. Our evaluation of \gls*{ppo} results in better performance for configurations 1 and 4. Note that for \gls*{ppo}, the second last run of configuration 4 failed. Figure \ref{fig:ppo_conf3_failed_run9} shows the return over time for this run. We assume it is an outlier, not representative of the true population, and exclude it from the statistical analysis.

\textbf{To test the significance of our results} we perform bootstrapping on the original 10 observations for each configuration. The resulting sample statistics (means and \glspl*{ci}) are presented in table \ref{tab:mean_ci}, where we obtain a different order of performance for the five configurations from that originally reported in \cite{mahmood2018benchmarking}. For \gls*{trpo}, we find that even though configuration 1 appears to show the best performance in figure \ref{fig:top-5}, the mean performance of configuration 2 is higher, suggesting a faster increase in performance. For \gls*{ppo}, we obtain the largest mean values from configurations 1 and 4. While it appears that the \glspl*{ci} are large for \gls*{ppo}, suggesting a greater range of performance, we recall that Henderson \etal \cite{henderson2018deep} speculates that exceedingly large confidence bounds might suggest an insufficient sample size.

\begin{table}[h!]
      \centering
      \resizebox{\textwidth}{!}{%
      \begin{tabular}{c|c|rl|c|rl}
      \toprule
        \multirow{4}{3cm}{Hyper-parameter configuration}   & \multicolumn{6}{c}{Algorithms} \\
                                        %\cline{2-7}
                                        & \multicolumn{3}{c|}{TRPO} & \multicolumn{3}{c}{PPO}\\
                                        \cline{2-7}
                                        & \cite{mahmood2018benchmarking} & \multicolumn{2}{c|}{Ours} & \cite{mahmood2018benchmarking} & \multicolumn{2}{c}{Ours} \\
                                        & $\hat{\mu}$ & \multicolumn{1}{c}{$\bar{\mu}$} & \multicolumn{1}{c|}{$95\%$ CI} & $\hat{\mu}$ & \multicolumn{1}{c}{$\bar{\mu}$} & \multicolumn{1}{c}{$95\%$ CI} \\
      \midrule
        c1                               & 158.56 & $135.78 $ & $(127.31,144.78)$ & 176.62 & $137.08 $ & $(116.64,157.73)$     \\% PPO 176.62 150.25 137.92 137.26 136.09
        c2                               & 138.58 & $139.65 $ & $(128.04,153.28)$ & 150.25 & $86.51  $ & $(58.48,115.48)$      \\ % TRPO: 158.56 138.58 131.35 123.45 122.60
        c3                               & 131.35 & $112.37 $ & $(91.38,134.72)$  & 137.92 & $90.12  $ & $(64.28,118.38)$      \\
        c4                               & 123.45 & $98.03  $ & $(93.34,103.18)$  & 137.26 & $119.43 $ & $(107.98,130.31)$     \\
        c5                               & 122.60 & $106.62 $ & $(95.57,118.60)$  & 136.09 & $82.42  $ & $(62.58,104.15)$      \\
      \bottomrule
      \end{tabular}
      }
      \caption{\textbf{Overall performance achieved by the two baselines:} The mean and 95\% \gls*{ci} for each hyper-parameter configuration is computed from the 10k samples obtained by bootstrapping from our original ten samples. All trials are conducted using the \gls*{ur} environment which corresponds to 300 hours of robot wall time. The reported average return by \cite{mahmood2018benchmarking} is denoted $\hat{\mu}$, while $\bar{\mu}$ denotes the empirical mean we obtained from bootstrapping.}
      \label{tab:mean_ci}
      \vspace{-0.5cm}
\end{table}

\textbf{The empirical distributions} we obtain from bootstrapping are presented in appendix \ref{app:norm_empirical_dists}, with a theoretical normal distribution fitted to the \gls*{edf}. To determine if the data was normally distributed we performed a normality test and, even though our data initially appears normally distributed, we found that 8 of 10 configurations would reject our null-hypothesis using a significance level of $\alpha=0.05$. Thus the data cannot be assumed normally distributed and other theoretical distributions must be considered.

\textbf{Fitting theoretical distributions to the \glspl*{edf}} is performed to determine which distribution fits the empirical data best. We tested 100 theoretical distributions, of which 52 converged successfully. On these 52, we compute a goodness of fit by performing a \gls{ks} test (see appendix \ref{app:dist_fitting}) from which we choose the most promising distributions determined from their $p$-values. This results in the six distributions presented in appendix \ref{app:dist_fitting}.

\textbf{Verifying reproducibility.} To verify the claims made by the original authors, we compute the probability that we would obtain a mean performance \textit{at least as good} as the originally reported average rewards (shown in appendix \ref{app:param_values}). We do this for each of the best-fit distributions and the results are shown in table \ref{tab:reproducibility_p_values}. If we have the probabilities $P_d$ that the distributions match the underlying \glspl{edf} (reported in appendix \ref{app:dist_fitting}) described as $P\{\mathrm{dist}=d|\mathrm{data}\}=P_d$, and the probability $P_v$ that we can get a value, $v$, at least as good (from that specific distribution), then we have $P\{v\geq\hat{\mu}|\mathrm{dist}=d, \mathrm{data}\} = P_v$. Thus $P\{v\geq \hat{\mu}|\mathrm{data}\} = P_d\cdot P_v$, where $\hat{\mu}$ denotes the single value of average return reported in \cite{mahmood2018benchmarking} and table \ref{tab:mean_ci}.

\begin{table}
      \resizebox{\textwidth}{!}{%
      \begin{tabular}{l|*5{C{1.15cm}}|*5{C{1.15cm}}}
      \toprule
        \multirow{3}{*}{Distributions}   & \multicolumn{10}{c}{Algorithms} \\
                                         & \multicolumn{5}{c}{TRPO} & \multicolumn{5}{c}{PPO}\\
                                         & c1 & c2 & c3 & c4 & c5 & c1 & c2 & c3 & c4 & c5\\
      \midrule
        beta       & 0.0000 & 0.5990$\star$ & 0.0436 & 0.0000 & 0.0022    &    0.0000 & 0.0000 & 0.0000 & 0.0015 & 0.0000 \\
        johnsonsb  & 0.0000 & 0.5985$\star$ & 0.0246 & 0.0000 & 0.0022    &    0.0000 & 0.0000 & 0.0000 & 0.0015 & 0.0000 \\
        johnsonsu  & 0.0000 & 0.3749$\star$ & 0.0417 & 0.0000 & 0.0015    &    0.0000 & 0.0000 & 0.0000 & 0.0011 & 0.0000 \\
        loggamma   & 0.0000 & 0.3894$\star$ & 0.0424 & 0.0000 & 0.0015    &    0.0000 & 0.0000 & 0.0000 & 0.0004 & 0.0000 \\
        powernorm  & 0.0000 & 0.2798$\star$ & 0.0407 & 0.0000 & 0.0013    &    0.0000 & 0.0000 & 0.0000 & 0.0012 & 0.0000 \\
        skewnorm   & 0.0000 & 0.2518$\star$ & 0.0411 & 0.0000 & 0.0014    &    0.0000 & 0.0000 & 0.0000 & 0.0010 & 0.0000 \\
      \bottomrule
      \end{tabular}
      }
      \caption{\textbf{Verification of reproducibility:} The table depicts the probabilities of obtaining a new sample \textit{at least as good} as the one reported in \cite{mahmood2018benchmarking} under the distributions listed. A failure to reject our hypothesis is indicated by $\star$.}
      \label{tab:reproducibility_p_values}
      \vspace{-0.5cm}
\end{table}

%===============================================================================

\section{Discussions and Conclusions}
\textit{Experimental Conclusions}

From table \ref{tab:reproducibility_p_values} we reject 55 out of 60 of our hypotheses stating that it would be possible to obtain a single sample at least as good as the one reported in \cite{mahmood2018benchmarking}. Our probability values conclude that the values reported in \cite{mahmood2018benchmarking} do not depict the range of performance very well. From the $p$-values we conclude that one run is not a representative sample of the underlying distribution, rather than dismissing the original work as irreproducible. The ideal way to compare results is to compare the resulting \glspl*{edf} from statistical analysis, which was not possible in our case as \cite{mahmood2018benchmarking} only reports one value of average return per configuration.

The key takeaway is that it is -- indeed -- very challenging to create reproducible robotic \gls{rl} research, and that having systems of uniform testing and ways to describe experiments can assist researchers in reporting their results without missing those important hyperparameters.

%===============================================================================================

\ \\\textit{Which recommendations can we draw from the experiments?}

\textbf{Reproducing \gls*{rl} results on a real-world robot} is not trivial and introduces many practical issues. The main issue we encountered was that the robot systematically stopped every four runs as \gls*{ur}'s controller interface (PolyScope) lost its connection to the real-time controller. It was challenging to figure out where and why the error occurred, and we never found the cause, but eventually worked around it by shutting down and restarting the robot controller after each trial. We conclude that physical issues are difficult to debug since it is hard to predict and solve bugs in the internal software of the physical robot.

\textbf{Managing software dependencies} is essential to creating reproducible experiments. We encourage reducing the number of dependencies to the minimum necessary to run the experimental code. Further, dependencies should be easy to install and use, which means that authors should, as a bare minimum, provide a list of dependencies and versions used to carry out their experimental work. We used the software container platform Docker\footnote{\url{https://www.docker.com/}} to manage our dependencies.

\textbf{Presetting and reporting the random seeds used} is key for ensuring that results are reproducible. However, when dealing with real-world robotics, where the sensor readings can be delayed and contain real measurement error, stochasticity will prevail. From our experiments, we found that if the PC used cannot process the traffic across the TCP socket fast enough, the sensor readings will not occur at the same time across different runs, leading to continuous stochasticity even if all parts of the software are seeded. This is visualised in Fig. \ref{fig:repeatability}. We argue that presetting (and reporting) the random seeds should be considered good practice, and we admit that we did not preset the random seeds for obtaining the results presented in this work. Not presetting and reporting the random seeds is an error we wish to highlight for others to avoid.

\textbf{Distinguishing between experimental code and library code} helps others to run the code more easily. One of our initial discoveries when attempting to reproduce the work of Mahmood \etal \cite{mahmood2018benchmarking} was that the experimental code they used for obtaining the reported results was not available.
The open-sourced code\footnote{\url{https://github.com/kindredresearch/SenseAct}} was library code and example code. We reached out to the lead author multiple times requesting both missing hyperparameters and experimental code, but did not obtain the code due to legal reasons. This restriction meant that we were forced to \textit{replicate}, rather than \textit{reproduce}, the original work.

\textbf{Logging of return values} is done by using the callback interface provided by the OpenAI implementations \cite{brockman2016openai}, where return values are only available at the end of each iteration which, in turn, is dependent on the algorithm's batch size. In order to obtain comparable results, all computed rewards during training should be logged and used for statistical inference. We did not do this as it would require us to diverge more significantly from the original code-base. 

\textbf{Hyper-parameters} have a significantly different effect across algorithms and environments, as shown in \cite{henderson2018deep}. Reporting both how the selected parameters were obtained, and all parameters themselves, is essential for allowing others to replicate work. 

%===============================================================================================

\ \\\textit{Which general recommendations can we draw from our work?}

\textbf{Separating evaluation from training} as highlighted by Khetarpal \etal \cite{khetarpal2018re} is significant as \gls*{rl} agents are shown to overfit quite robustly to training instances \cite{zhang2018study}. The use of different preset seeds for training and evaluation also is to be encouraged. While we wished to follow and extend the pipeline in \cite{khetarpal2018re}, the current states of the proposed pipeline and OpenAI Baselines are incompatible and are thus subject for future work.

\textbf{Thorough documentation} is essential to ensure reproducibility and comparability, as even small details of an experiment can be critical. One may argue that documentation can become very time consuming and take valuable time away from potential progress. We believe that it is a question of establishing a culture in the field where the documentation of the experimental work is an integrated part of conducting research. Many tools can be utilised to ease the process. Our proposed method, including the configuration files, presents an attempt to ease the documentation process.

\textbf{Measurement metrics} are one of the key issues when comparing \gls*{rl} algorithms. Often, \textit{how} and \textit{when} performance is recorded remains unreported \cite{khetarpal2018re}. Different implementations of algorithms collect, process and store the performance metrics differently, making comparisons difficult if not impossible.

When choosing the right metric when reporting the results of conducted experiments, the designer must first make clear what makes an algorithm good. Next, the designer must determine how to measure that. In general, there are two aspects to consider:
1) \textit{Good performance}
describes how well the algorithm works in the specific task addressed, and 
2) \textit{Efficient use} 
describes the cost of achieving a \textit{satisfactory performance} and might include the time spent optimising hyper-parameter choices, the effort required to compute the algorithm's output, and/or the cost of the data obtained during training. Metrics can measure either of these, and it is the designer's choice according to what problem is sought solved, but in order to ensure comparability in scientific papers the authors should check which standard metrics are reported and report those, as well as potential additional ones selected by the authors. Further, the metrics chosen should be reported, explained, and argued for, including how they are computed.

\textbf{Open-Source Research}
Through this work we advocate for open-sourcing all aspects of research, including source code, as we believe it to be paramount to ensure the continued progress of the scientific fields. As a bare minimum, a study or experiment should be advertised with enough details so any third-party scientist, with sufficient skills, can obtain the same results within the marginals of experimental error \cite{plesser2018reproducibility}. Many researchers fail even to meet these simple though crucial requirements due to all sorts of reasons \cite{sandve2013ten}.

\textbf{The idea of a trade-off between fast continued progress and rigorous analysis} is one we have met from multiple sources. We believe that irreproducible research is not actually research, merely data -- and poor quality of data at that -- and that there is not really a trade-off between going somewhere carefully and possibly nowhere fast.

%===============================================================================

\newpage
\supplementary{All library code, hence adaptations to the SenseAct framework, is publicly available at: \texttt{\url{https://github.com/dti-research/SenseAct}} and the docker image we have used throughout this work is available at: \texttt{\url{https://hub.docker.com/r/dtiresearch/senseact}}. All the experimental code and data to regenerate the figures in this manuscript is available at \texttt{\url{https://github.com/dti-research/SenseActExperiments}}. If you find something missing or not working, please feel free to contact the lead author.}
\authorcontributions{All authors have made substantial contributions to the conception and design of the work. Nicolai A. Lynnerup and Laura Nolling wrote the original draft of this manuscript while John Hallam and Rasmus Hasle have substantively revised it. Laura Nolling made the changes to the SenseAct framework in order to make it callable by external programs and added logging functionality. Further, Laura Nolling devised the bootstrapping and evaluation scripts, while Nicolai A. Lynnerup programmed the distribution-fitting script and reviewed and edited evaluation and bootstrapping scripts. Nicolai A. Lynnerup created the Docker images and adapted the SenseAct framework to work with all versions of the CB-series UR robots, based on the work of Oliver Limoyo\footnote{\url{https://github.com/kindredresearch/SenseAct/pull/29}}, Ph.D. Student at the University of Toronto. Further, Nicolai A. Lynnerup conducted the literature review on reproducibility. Rasmus Hasle and John Hallam provided critical feedback and helped shape the research and analyses. Nicolai A. Lynnerup, Rasmus Hasle, and John Hallam secured the research funding, while all authors have approved the submitted version.}
\funding{The work presented in this paper is partially funded by the Danish Technological Institute and partially Innovation Fund Denmark through their Industrial Researcher Program, grant 8053-00057B.}
\acknowledgments{This work benefited from the help of many people beyond the authors. We want to show our gratitude to Jens-Jakob Bentsen, and Thomas Mosgaard Giselsson, specialists at Danish Technological Institute (DTI) for numerous discussions on different aspects of the reproducibility terminology. We further thank Jens-Jakob Bentsen for all his help debugging the UR robot's communication interface and underlying controller functionality. Next, we would like to thank Rasmus Lunding Henriksen, specialist at DTI, for his comments that significantly improved the manuscript. Additionally, we would like to show our gratitude to Kasper Stoy, Professor at ITU Copenhagen, for his comments on our work, which helped us see more perspectives on reproducibility and reporting in science. We would also like to show our gratitude to the two anonymous reviewers for their insights and constructive comments. Any errors are our own and should not tarnish the reputations of these persons nor the institutions.
}
\conflictsofinterest{The authors declare no conflict of interest. The funding sponsors had no role in the design of the research; in the collection, analyses or interpretation of data; in the writing of this manuscript, nor in the decision to, or where to, publish the results.}

%===============================================================================

\printbibliography

%===============================================================================

  \newpage
  \appendix
  \section*{Appendices}
  % Section numbering
  \renewcommand{\thesection}{A\arabic{section}}

  \section{A Brief Overview of a Confused Taxonomy}\label{app:taxonomy}
  \subsection{Two Perspectives on Terminology}

Unfortunately there exists some confusion on the meaning of \textit{repeatability}, \textit{reproducibility} and \textit{replicability} which in turn negatively affects the overall development of science. Barba \cite{barba2018terminologies} group a series of papers into 2 groups; A. Those who \textbf{do not} distinguish between the words; \textit{reproducibility} and \textit{replicability}, and B. Those who \textbf{do} distinguish between the two words. Group B is then divided into two additional groups who's contradicting conventions are shown below.

\subsection*{B.1. The Claerbout, Donoho, Peng Convention}
From the pioneering work of Claerbout \cite{claerbout1992electronic} Buckheit and Donoho \cite{buckheit1995wavelab} and Peng \etal \cite{peng2006reproducible} the following convention has been derived, but as the papers are somewhat dated it might be beneficial to read the more recent work by Schwab \etal \cite{schwab2000making}, Donoho \cite{donoho2009reproducible} and Peng \etal \cite{peng2011reproducible} instead.

\paragraph{Reproducibility} describes a study where the original authors has provided all the necessary observations and potentially computer code to run the method again, allowing a third party scientist to reproduce the same results. Hence; \textit{different} team with \textit{same} experimental setup.
\paragraph{Replicability} is used to describe a third party study that arrives at the same conclusions as an original study, where new observations are collected and the method is implemented based on the published paper. Hence; \textit{different} team with \textit{different} experimental setup.

\subsection*{B.2. The ACM, Drummond Convention}
Drummond \cite{drummond2009replicability} published his article at the International Conference on Machine Learning (ICML) in 2006 where he unfortunately switched the definitions of \textit{reproducibility} and \textit{replicability} around, which according to Professor of Linguistics Mark Liberman should be rejected \cite{liberman2015replicability} as the term was coined much earlier by Claerbout \cite{claerbout1992electronic}. Prior to the suggestion of rejecting the re-coining of terms, Drummond's ``new definitions'' spread through several scientific papers. Fang \etal \cite{casadevall2010reproducible} and Mende \etal \cite{mende2010replication} seems to have picked up the confusion of the terms from Drummond. Further the Association for Computing Machinery (ACM) is also using the terms wrong in their badging of artifacts system \cite{ACM2017artifact}, and they seem to stick with the definition, as it apparently is revised latest in 2017, two years after Mark Liberman published his findings. The ACM, Drummond convention is as follows.

\paragraph{Repeatability} (\textit{same} team, \textit{same} experimental setup) The observations can be obtained with the stated precision by the same researchers using the same method and the same hardware under the same conditions in the same location.
\paragraph{Replicability}  (\textit{different} team, \textit{same} experimental setup) The observations can be obtained with the stated precision by third party researchers using the same method, the same hardware under the same conditions in the same or different location.
\paragraph{Reproducibility} (\textit{different} team, \textit{different} experimental setup) The observations can be obtained with the stated precision by third party researchers using different hardware on a different location.

% advocate vs. plead?
We advocate that all researchers refrain from using the terms the ACM, Drummond way as the re-coining of the terms is not justified.

\paragraph{The conflicting terminologies} are annoying at least and at worst a blockade for the progress of science.

\subsection{Expanding the Terminologies}
In addition to re-coining terms, there exists various papers suggesting coining new more descriptive terms as a way out of the heated discussions regarding \textit{repeatability}, \textit{reproducibility} and \textit{replicability}. As the number of papers re-coining the terms are very large we present only a few of them here as a complete review is beyond the scope of this work.

Goodman \etal \cite{goodman2016does} proposes a lexicon for reproducibility by differentiating between \textit{methods reproducibility}, \textit{results reproducibility} and \textit{inferential reproducibility} to solve the terminology confusion. The three new \textit{reproducibility} terms are defined below, where Goodmann \etal \cite{goodman2016does} argues that the definitions should be clear in principle but operationally elusive, why they provide many operational examples specific to certain scientific fields.

\begin{itemize}
    \item \textit{Methods reproducibility:} provide enough detail about study procedures and data so the same procedures could, in theory or in actuality, be exactly repeated.
    \item \textit{Results reproducibility:} obtain same results from the conduct of an independent study whose procedures are as closely matched to the original experiment as possible.
    \item \textit{Inferential reproducibility:} draw similar conclusions from either an independent study replication or a a reanalysis of the original study. See Gilbert \etal \cite{gilbert2016comment} for a debate on the difference between results and inferential reproducibility.
\end{itemize}

The coining of these three terms is an attempt to make it more specific what aspects of ``trustworthiness'' we focus on, when analysing a study. Thus avoiding the ambiguity caused by everyday language's indifferent use of the words; \textit{repeatable}, \textit{reproducible} and \textit{replicable}.

Tatman \etal \cite{tatman2018practical} proposes a practically oriented taxonomy for \textit{reproducibility} more or less specific for \gls*{ml} research. As Goodman \etal \cite{goodman2016does}, they are attempting to make the principally clear definitions of the terms less operationally elusive by comprehensive descriptions of to what extend details, code and data are shared. While we applaud the author's take on a simple and practical taxonomy we discourage the seemingly indifferent use of the term \textit{reproducibility} \cite{tatman2018practical} as it contradicts the Claerbout, Donoho, Peng convention, which could have been easily avoided.

\begin{itemize}
    \item \textit{Low reproducibility: Finished paper only} is essentially \textit{replicability} in the Claerbout, Donoho, Peng convention where a well-written publication -- in theory -- should be enough for a third party scientist to \textit{replicate} the study or experiment. However, as the authors argue, this is often impractical and even impossible given time constraints and/or missing information.
    \item \textit{Medium reproducibility: Code and data, no environment} relates to \textit{reproducibility} in the Claerbout, Donoho, Peng convention. The original paper from Claerbout \cite{claerbout1992electronic} does not explicitly describe whether the environment should be a part of the scholarship or not as the tools making this possible simply weren't available at the time. With today's tools such as Docker, we argue that it should be a part of it. LeVeque \cite{leveque2013top}, argues that as long as the code is present alongside the publication then it is not critical whether the code runs or not, as the code itself contains a wealth of details that may not appear in the description. The bare minimum requirement should hold that information regarding the environment -- versions, etc. -- should be a critical part of the publication.
    \item \textit{High reproducibility: Code, data and environment} also relates to \textit{reproducibility} in the Claerbout, Donoho, Peng convention where, as discussed above, the environment is included, making the \textit{reproducibility} process easier for the reviewer or reader as this researcher does not have to mess around with installing all kinds of libraries in certain versions on top of his/hers functioning system.
\end{itemize}

\subsection{Discussion \& Conclusion}

Firstly, we believe that we should honor the often cited quote from Buckheit \& Donoho \cite{buckheit1995wavelab} paraphrasing Jon Claerbout \cite{claerbout1992electronic} and submit our code alongside our publications.

\begin{displayquote}
``An article about computational science in a scientific publication is not the scholarship itself, it is merely advertising of the scholarship. The actual scholarship is the complete software development environment and the complete set of instructions which generated the figures.''
\end{displayquote}

Secondly we encourage the community to find some common ground regarding the terminology, so focus once again can be on evolving the scientific field instead of terminology. The bare minimum requirement is that researchers at least state what they mean when they use the terms.

We wish to highlight that \textit{reproducibility} is not a ``free-pass'' for the readers to use without properly investigating the submitted code. Consider the case where a researcher is to \textit{reproduce} the results of a published scientific paper and has hence obtained the original code and data. This third-party researcher can now re-run the code and (hopefully) obtain the same results. The problem occurs when the code is run without it being understood, making it possible for the third-party researcher to obtain the results without him finding the potential bugs in the code from the original author. This is likewise the case when someone commits fraud and makes the fraud reproducible. Our claim is that when other researchers tries to build on top of faulted experiments and thus transfer the methodologies to other domains the fraud or bugs will -- often, if not always -- become apparent. This, as we assume that it must take some significant hand-engineering to make the fraud reproducible to the specific problem, making it non-generic.

  \section{Practicalities for Ensuring Reproducibility}\label{app:practical_reproducibility}
  As discussed in \cite{mahmood2018setting} there exists many practical issues when setting up \gls*{rl} tasks on real-world robots, especially when attempting to ensure reproducibility. Below we outline some of the most important practices that should be encouraged in order to ensure reproducibility.

\paragraph{Seed the Random Number Generator}
    A simple, yet effective, approach is to seed the random number generator, with a set of predefined seeds, so experiments can be repeated and reproduced. For an overview of the different sources of non-determinism in \gls*{ml} see appendix \ref{app:non_determinism}.

\paragraph{Avoiding the Dependency Hell}
    To avoid (most of) the problems related to the colloquial term \textit{dependency hell}, researchers can advantageously utilise tools such as Docker. For this work, we adapted the SenseAct framework so we could build it into a Docker image.

\paragraph{Code Reviews}
    considers how to ensure the integrity of code by letting others review it prior to conducting experiments. This indirectly enforces developers to write understandable code (including comments). This might take a little longer than simply throwing pieces of code together, but the benefits of code reviewing prior to testing should motivate most development teams to conform to this practice. One of the largest benefits is the increased probability for avoiding bad experiments due to bugs in the code, as two sets of eyes are commonly known to be better than one.

\paragraph{Open-Sourcing Gathered Artefacts}
    Gathered data should, as well as code, be open-sourced. This helps verifying reproductions and potential replications. All data used for describing the methods shown in this work is available at our GitHub repository\footnote{{\iffinal \url{https://github.com/dti-research/SenseActExperiments} \else \textit{URL anonymised} \fi}}.

  \newpage
  \section{Comprehensive List of Hyper-parameter Ranges and Values}\label{app:param_values}
  This appendix presents all hyper-parameter values that we have used to conduct our experiments. The hyper-parameter configurations used in this work are presented in table \ref{tab:trpo_param_configs} and table \ref{tab:ppo_param_configs}. The hyperparameters that the authors of the original work chose not to tune have been kept fixed throughout our experiments and are presented in table \ref{tab:app_param_values}. We wish to highlight that for;
  a) \gls*{trpo} the step-size is denoted \lstinline{vf_stepsize}, 
  b) \gls*{ppo} the step-size is denoted \lstinline{optim_stepsize}. This information is obtained through correspondence with the original authors of \cite{mahmood2018benchmarking}.

%%%%%%%%
% TRPO %
%%%%%%%%
\begin{table}[h!]
\centering
\begin{tabular}{R{0.25cm} R{1.2cm} *{7}{R{1.15cm}}}
\toprule
\multicolumn{9}{c}{\textbf{TRPO}}\\
\# & Average Return  & Hidden Layers & Hidden Size &Batch Size & Step Size & $\gamma$ & $\lambda$ & $\delta_{KL}$\\
 \midrule
1 & 158.56  &   2    &   64     &   4096   &    0.00472  &  0.96833    &  0.99874  &   0.02437  \\
2 & 138.58  &   1    &   128    &   2048   &    0.00475  &  0.99924    &  0.99003  &   0.01909  \\
3 & 131.35  &   4    &   64     &   8192   &    0.00037  &  0.97433    &  0.99647  &   0.31222  \\
4 & 123.45  &   4    &   128    &   4096   &    0.00036  &  0.99799    &  0.92958  &   0.01952  \\
5 & 122.60  &   4    &   32     &   2048   &    0.00163  &  0.96801    &  0.96893  &   0.00510  \\
\bottomrule
\end{tabular}
\caption{\textbf{Hyper-parameter configurations found by random search:} The table presents the top-5 configurations found for \gls{trpo} in \cite{mahmood2018benchmarking}.}
\label{tab:trpo_param_configs}
\end{table}

%%%%%%%
% PPO %
%%%%%%%
\begin{table}[h!]
\centering
\begin{tabular}{R{0.25cm} R{1.2cm} *{7}{R{1.15cm}}}
\toprule
\multicolumn{9}{c}{\textbf{PPO}}\\
\# & Average Return  & Hidden Layers & Hidden Size & Batch Size & Step Size & $\gamma$ & $\lambda$ & Opt. Batch Size\\
 \midrule
1 & 176.62  &   3   &   64    &    512   &  0.00005   &     0.96836     &   0.99944   &    16    \\
2 & 150.25  &   1   &   16    &    256   &  0.00050   &     0.99926     &   0.98226   &    64    \\
3 & 137.92  &   1   &   2048  &    512   &  0.00011   &     0.99402     &   0.90185   &    8     \\
4 & 137.26  &   4   &   32    &    2048  &  0.00163   &     0.96801     &   0.96893   &    1024  \\
5 & 136.09  &   1   &   128   &    2048  &  0.00280   &     0.99924     &   0.99003   &    32    \\
\bottomrule
\end{tabular}
\caption{\textbf{Hyper-parameter configurations found by random search:} The table presents the top-5 configurations found for \gls{ppo} in \cite{mahmood2018benchmarking}.}
\label{tab:ppo_param_configs}
\end{table}

\begin{table}[!ht]
      \centering
      \begin{tabular}{L{3cm}|*{2}{C{2.25cm}}}
      \toprule
        Hyperparameter                         & \multicolumn{2}{c}{Fixed Values} \\
                                               & TRPO                          & PPO       \\
      \midrule
        Max. Timesteps                         & $150,000$                     & $150,000$ \\
        Entropy Coef.                          & 0.0                           & 0.0       \\
        CG Iterations                          & 10                            & -         \\
        CG Damping                             & 1e-2                          & -         \\
        VF Iterations                          & 3                             & -         \\
        Clip Parameter                         & -                             & 0.2       \\
        Optim. Epochs                          & -                             & 10        \\
        Adam $\epsilon$                        & -                             & 1e-5      \\
      \bottomrule
      \end{tabular}
      \caption{\textbf{Fixed hyperparameter values:} The table shows the fixed hyperparameter values that are not included in the random search.}
      \label{tab:app_param_values}
\end{table}

  \newpage
  \section{Fitting of Theoretical Distributions}\label{app:dist_fitting}
  We fit 100 theoretical distributions\footnote{For the comprehensive list see: \url{https://docs.scipy.org/doc/scipy/reference/stats.html}} to our \glspl*{edf} and use the \gls*{ks} test to determine the goodness of fit. From this, we can determine which statistic to use in order to conduct the hypothesis testing.

\def \trpocaption {
        \textbf{Theoretical distribution fitting on empirical data from \gls*{trpo}:}   
        (\textit{top}) 52 theoretical distributions fitted to the empirical data of the \nth{1} configuration of \gls*{trpo},
        (\textit{left}) top-6 theoretical distributions (chosen based on $p$-value),
        (\textit{right}) best fitted theoretical distributions, each row corresponds to the hyperparameter configuration.
        }

\begin{figure}[h!]
    \centering
    \begin{minipage}{0.49\textwidth}
        \centering
        \includegraphics[width=1.0\textwidth,trim={10pt 0 35pt 0},clip]{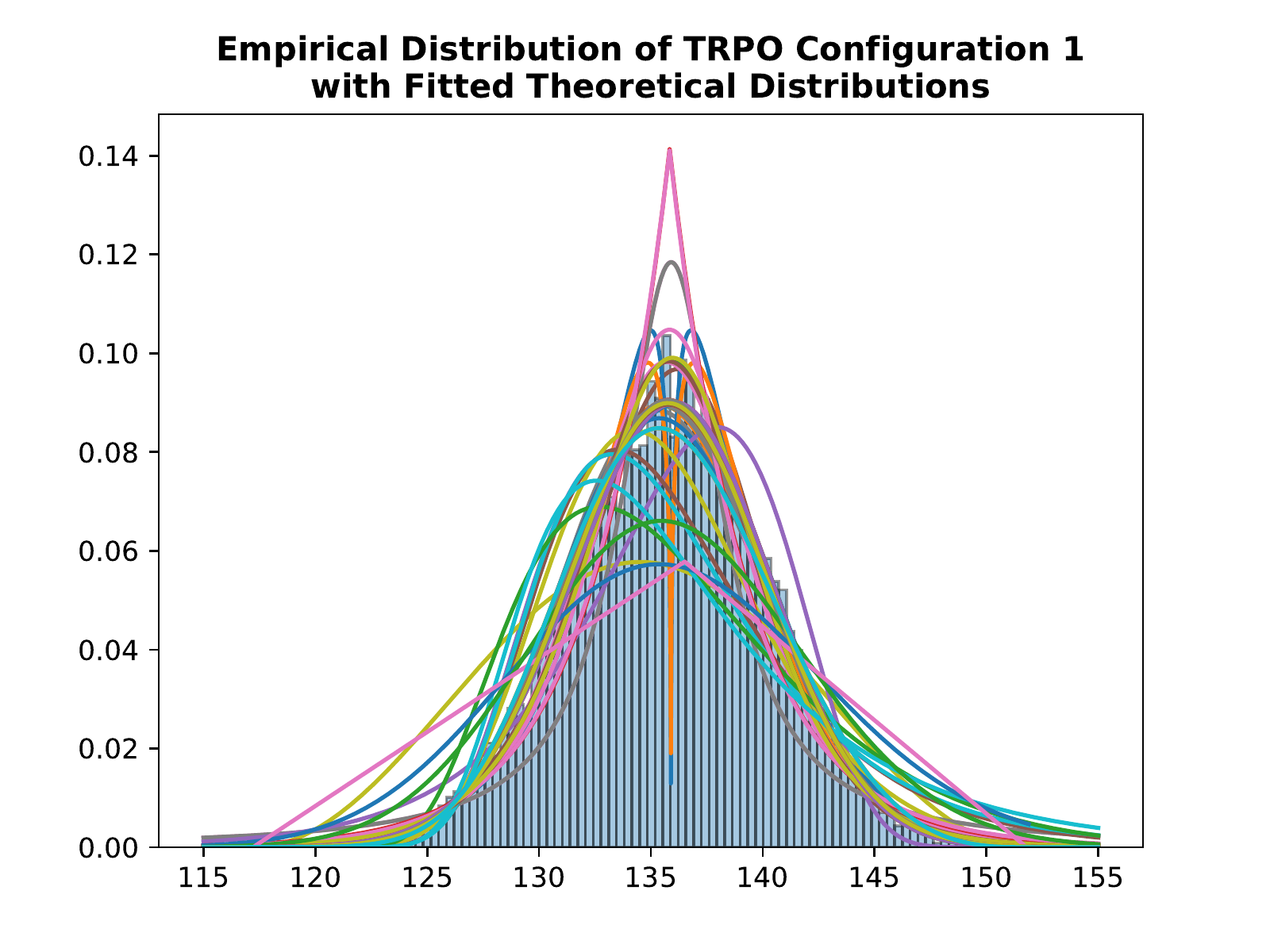}
    \end{minipage}\\
    
    \begin{minipage}{0.49\textwidth}
        \centering
        \includegraphics[width=1.0\textwidth,trim={10pt 0 35pt 0},clip]{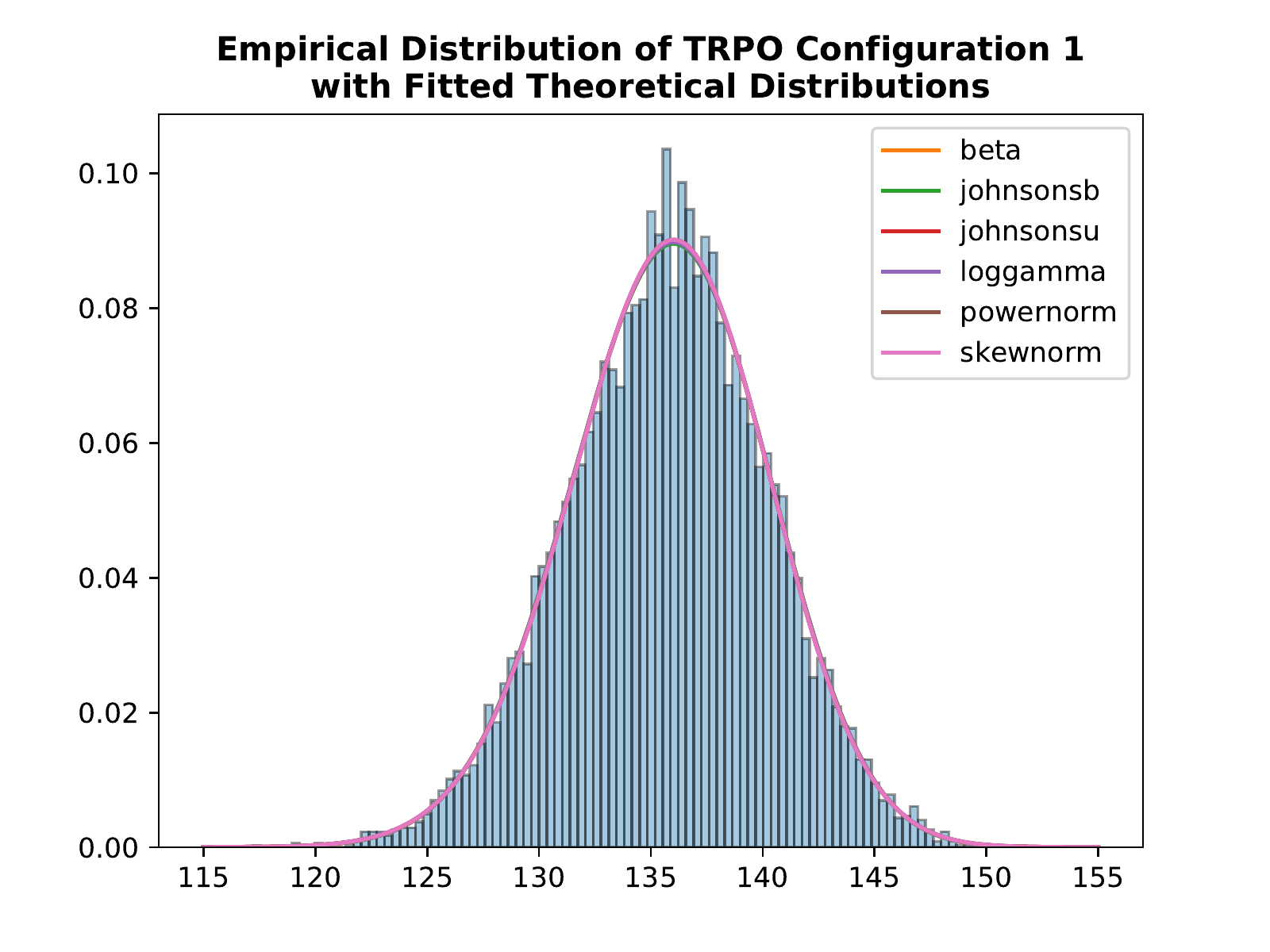}
    \end{minipage}
    \begin{minipage}{0.49\textwidth}
        \centering
        \includegraphics[width=1.0\textwidth,trim={10pt 0 35pt 0},clip]{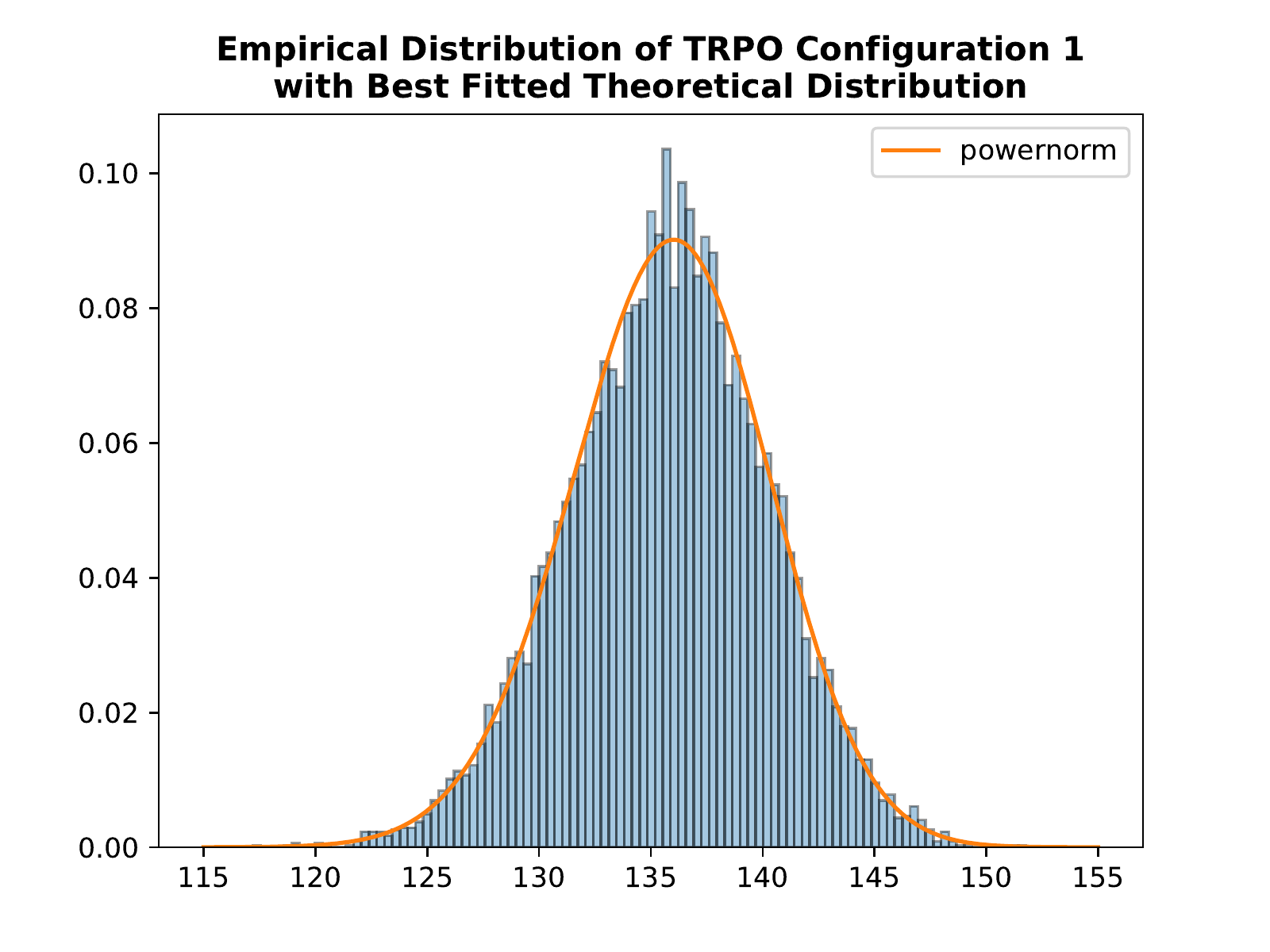}
    \end{minipage}
    \begin{minipage}{0.49\textwidth}
        \centering
        \includegraphics[width=1.0\textwidth,trim={10pt 0 35pt 0},clip]{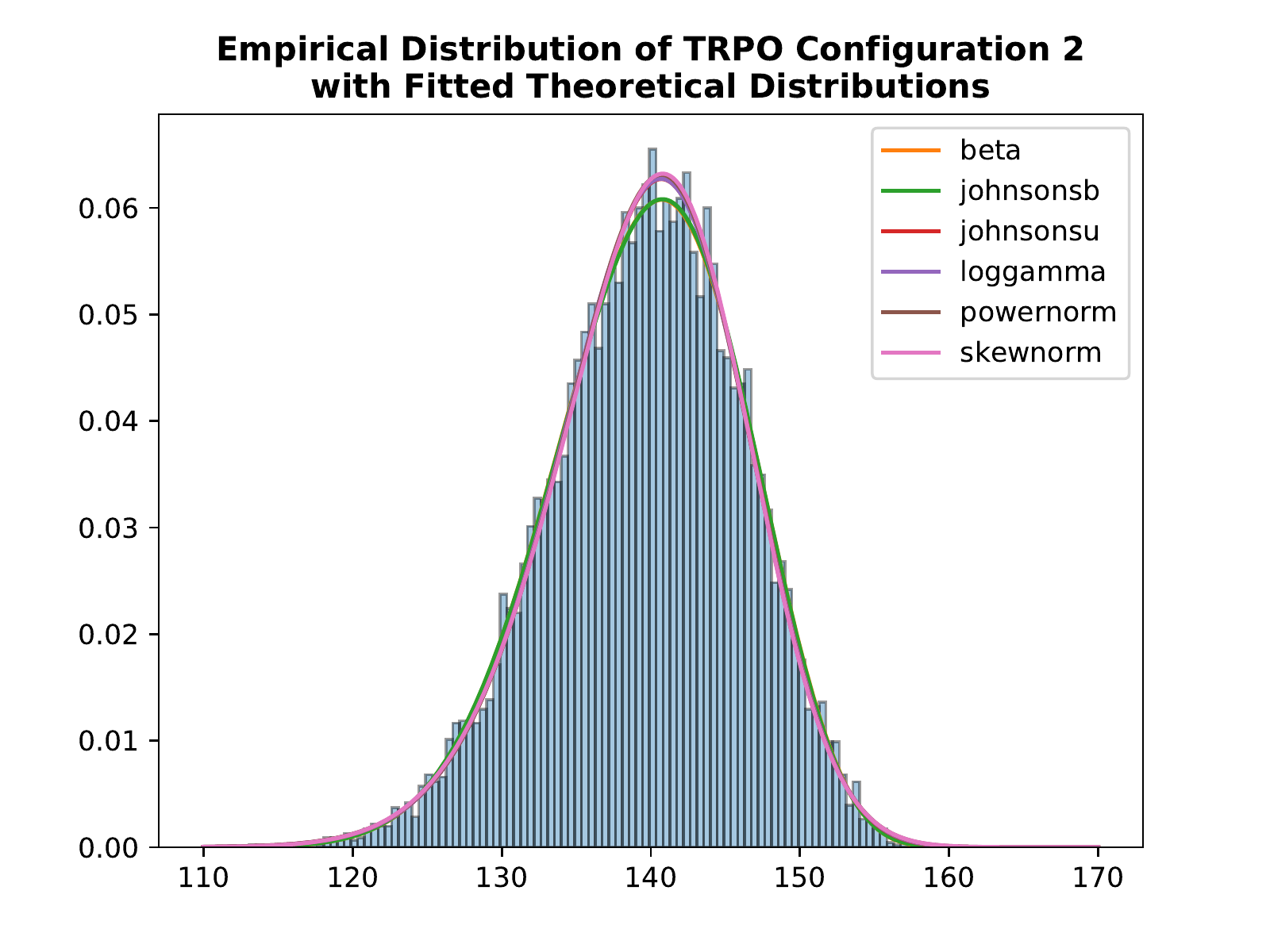}
    \end{minipage}
    \begin{minipage}{0.49\textwidth}
        \centering
        \includegraphics[width=1.0\textwidth,trim={10pt 0 35pt 0},clip]{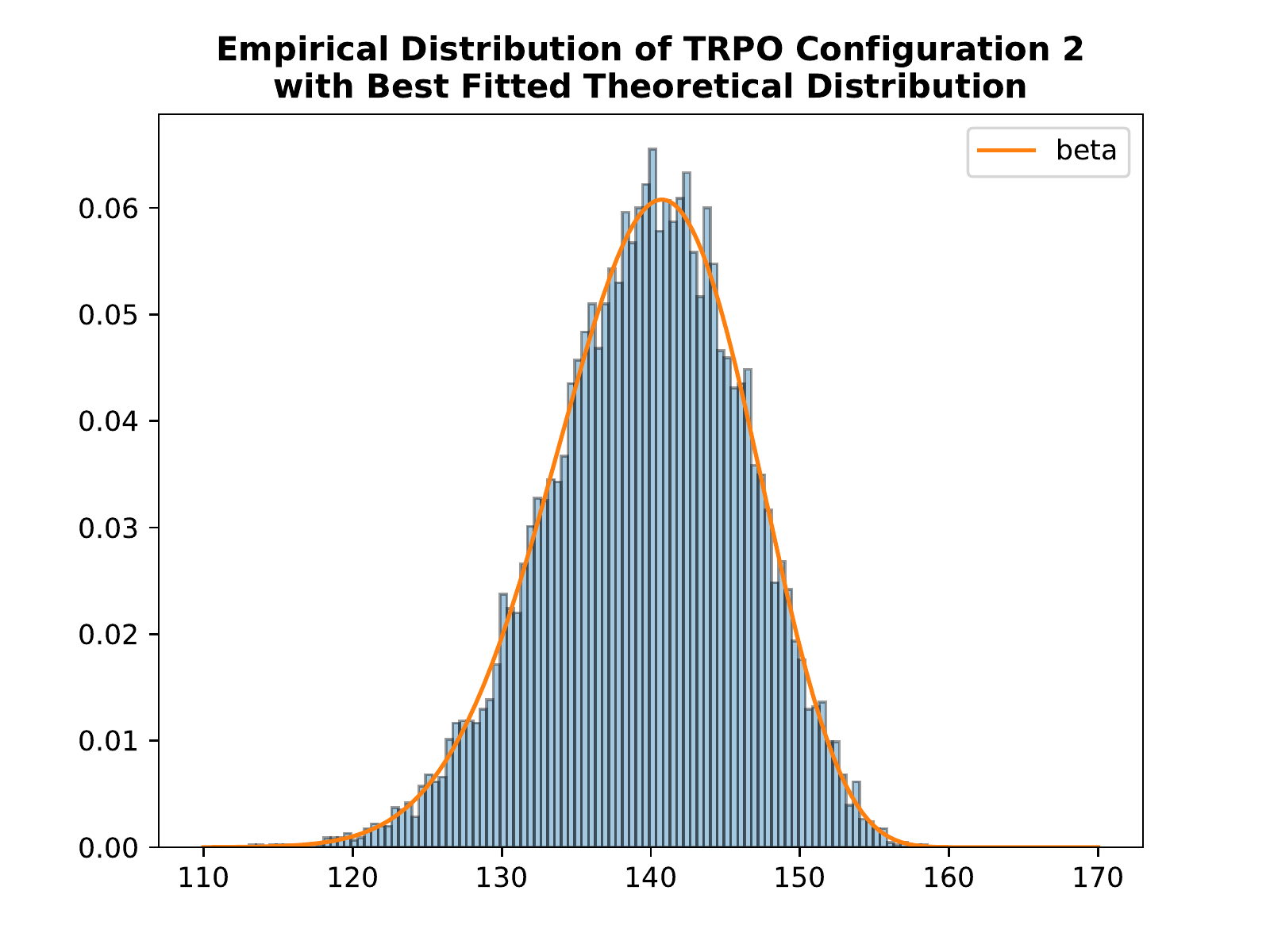}
    \end{minipage}
    \caption{\trpocaption {\tiny(figure continues on next page)}}
\end{figure}

\begin{figure}[p]\ContinuedFloat
    \centering
    \begin{minipage}{0.49\textwidth}
        \centering
        \includegraphics[width=1.0\textwidth,trim={10pt 0 35pt 0},clip]{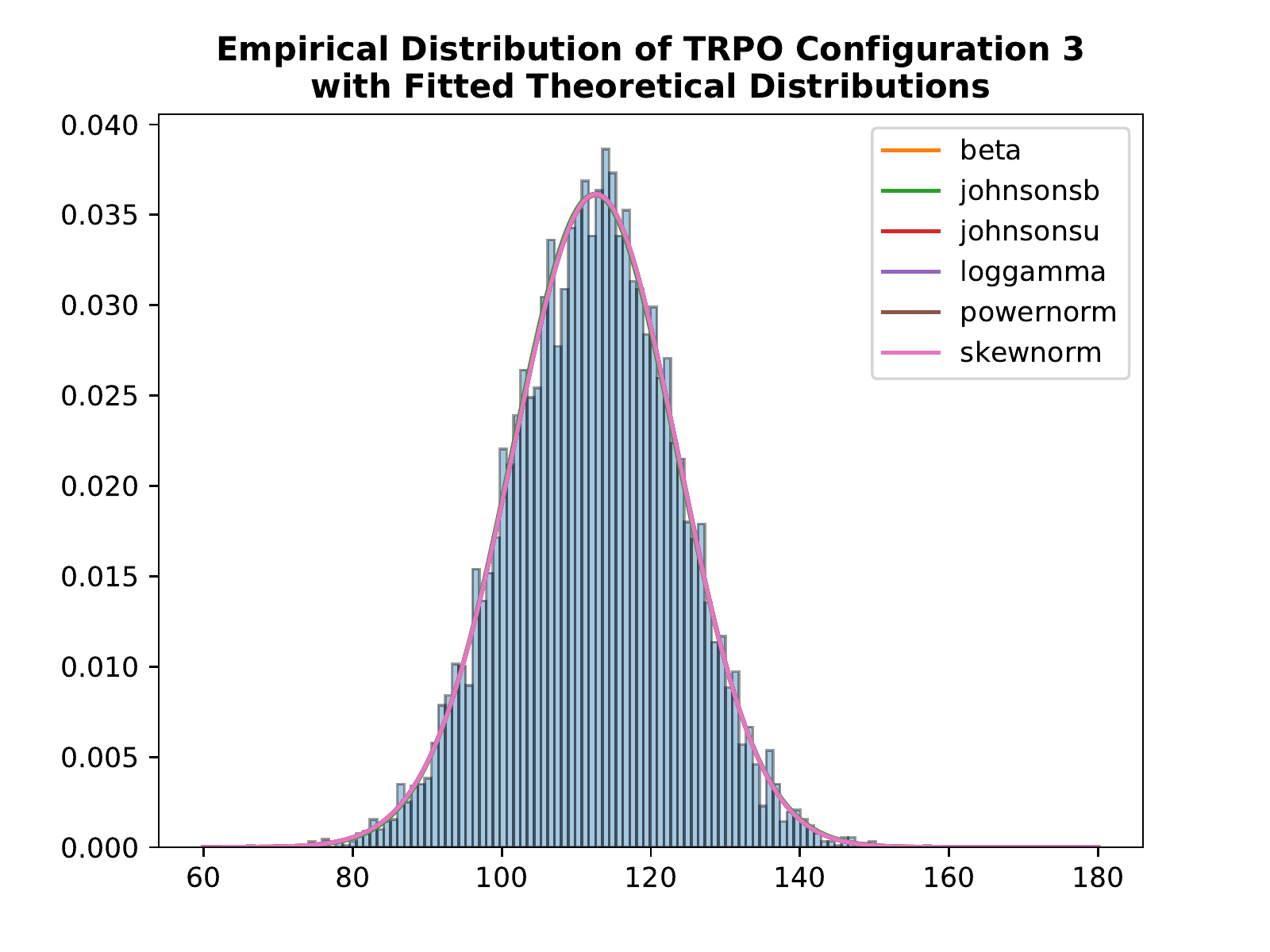}
    \end{minipage}
    \begin{minipage}{0.49\textwidth}
        \centering
        \includegraphics[width=1.0\textwidth,trim={10pt 0 35pt 0},clip]{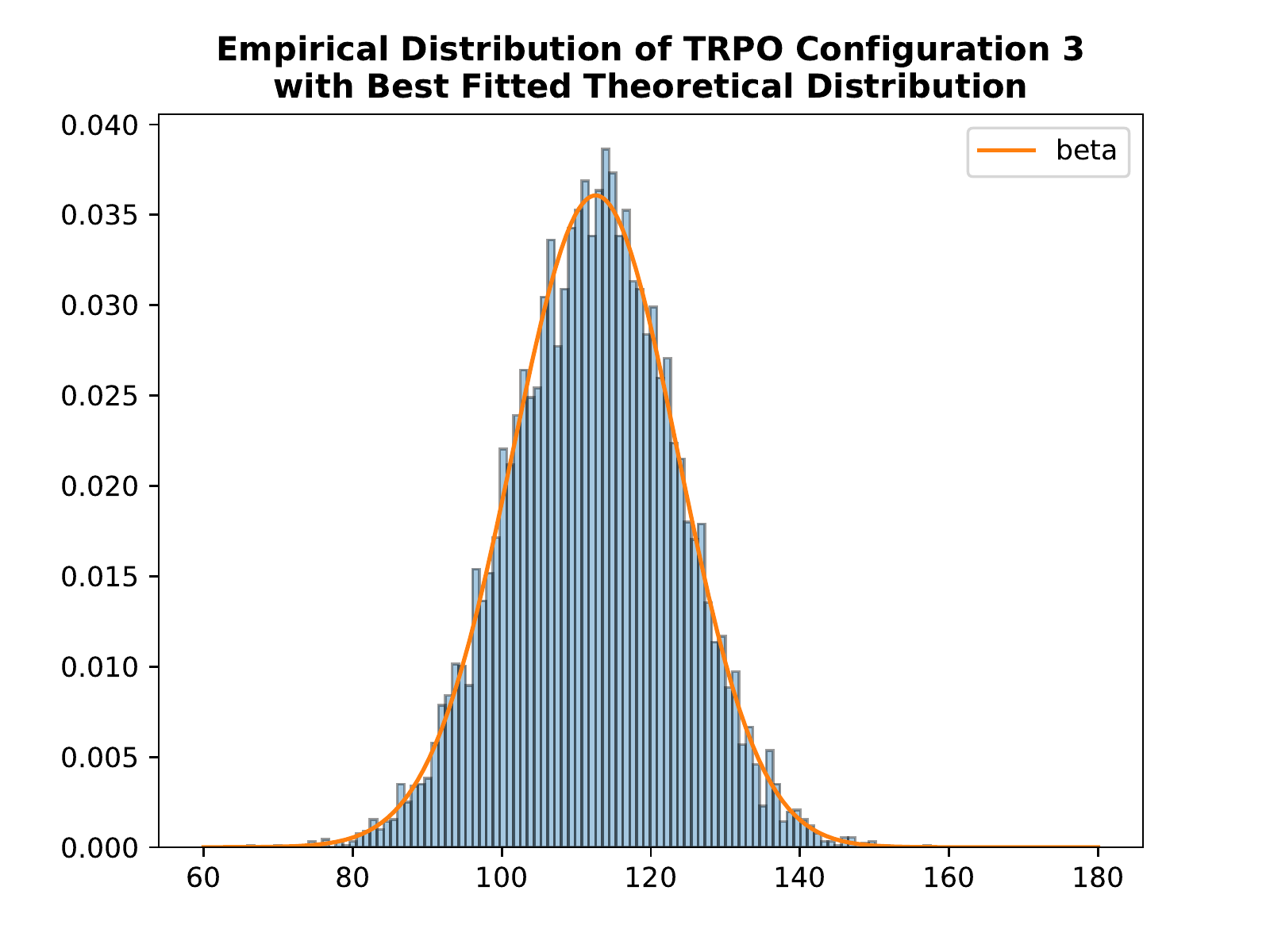}
    \end{minipage}
    \begin{minipage}{0.49\textwidth}
        \centering
        \includegraphics[width=1.0\textwidth,trim={10pt 0 35pt 0},clip]{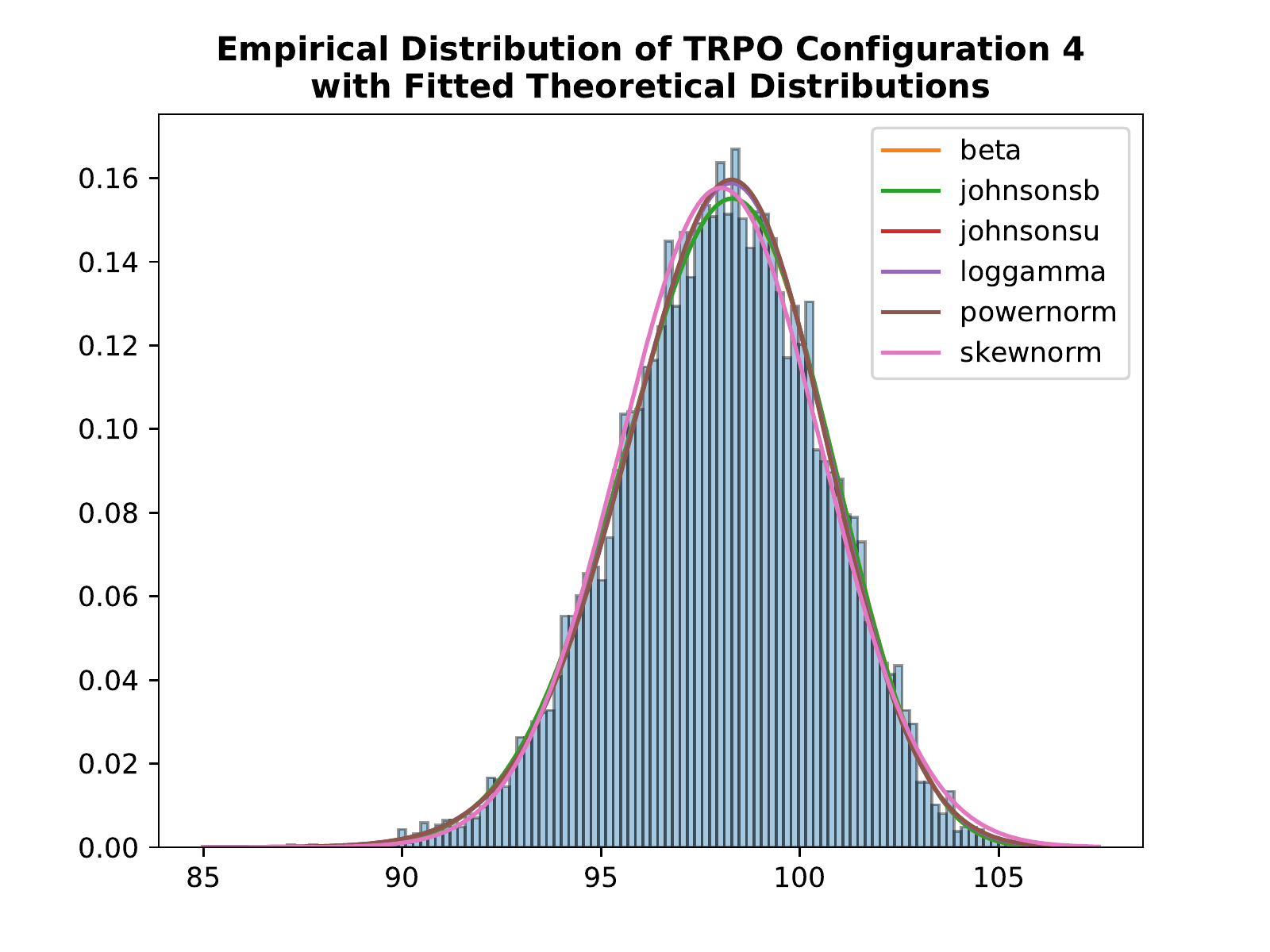}
    \end{minipage}
    \begin{minipage}{0.49\textwidth}
        \centering
        \includegraphics[width=1.0\textwidth,trim={10pt 0 35pt 0},clip]{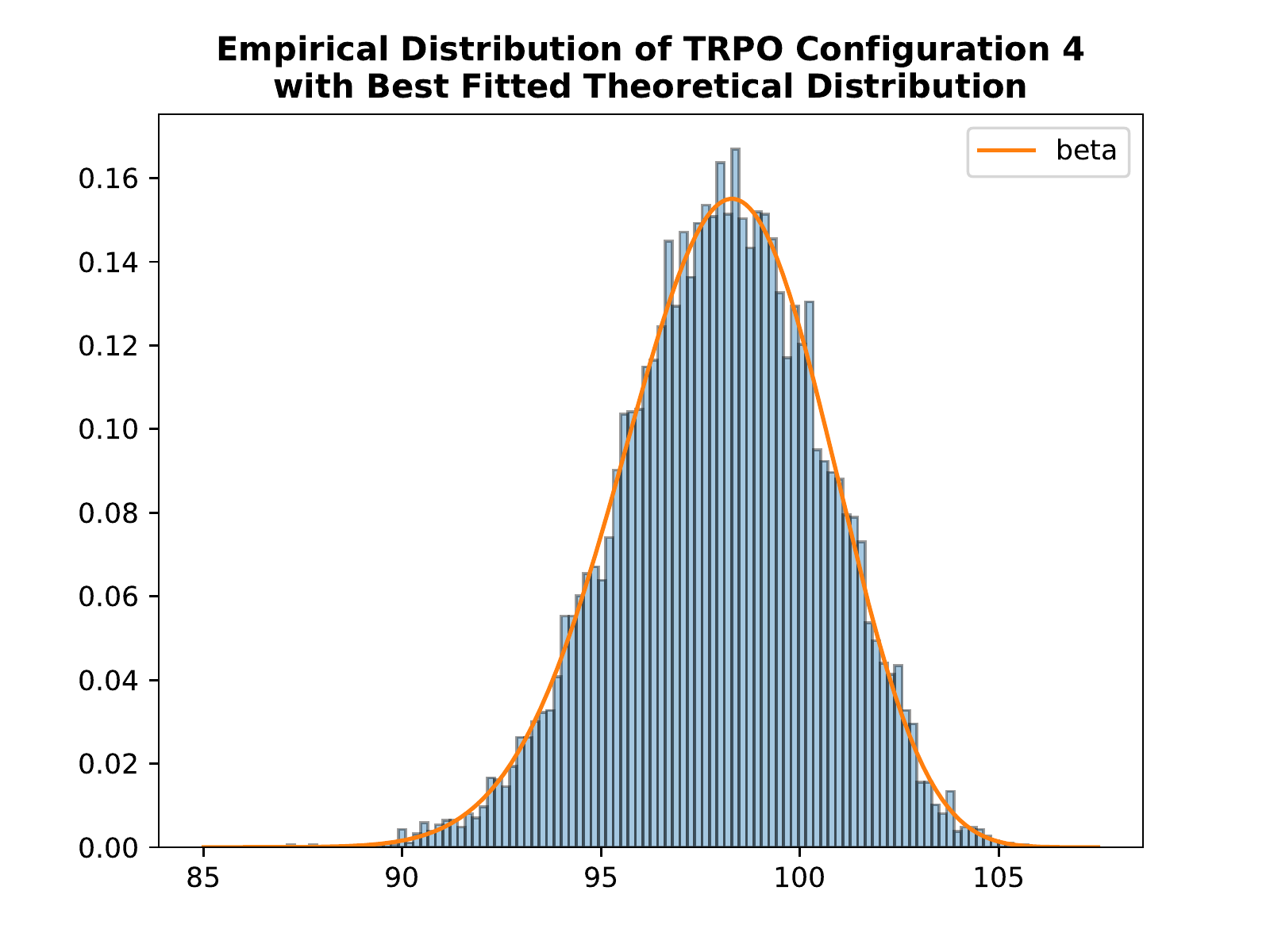}
    \end{minipage}
    \begin{minipage}{0.49\textwidth}
        \centering
        \includegraphics[width=1.0\textwidth,trim={10pt 0 35pt 0},clip]{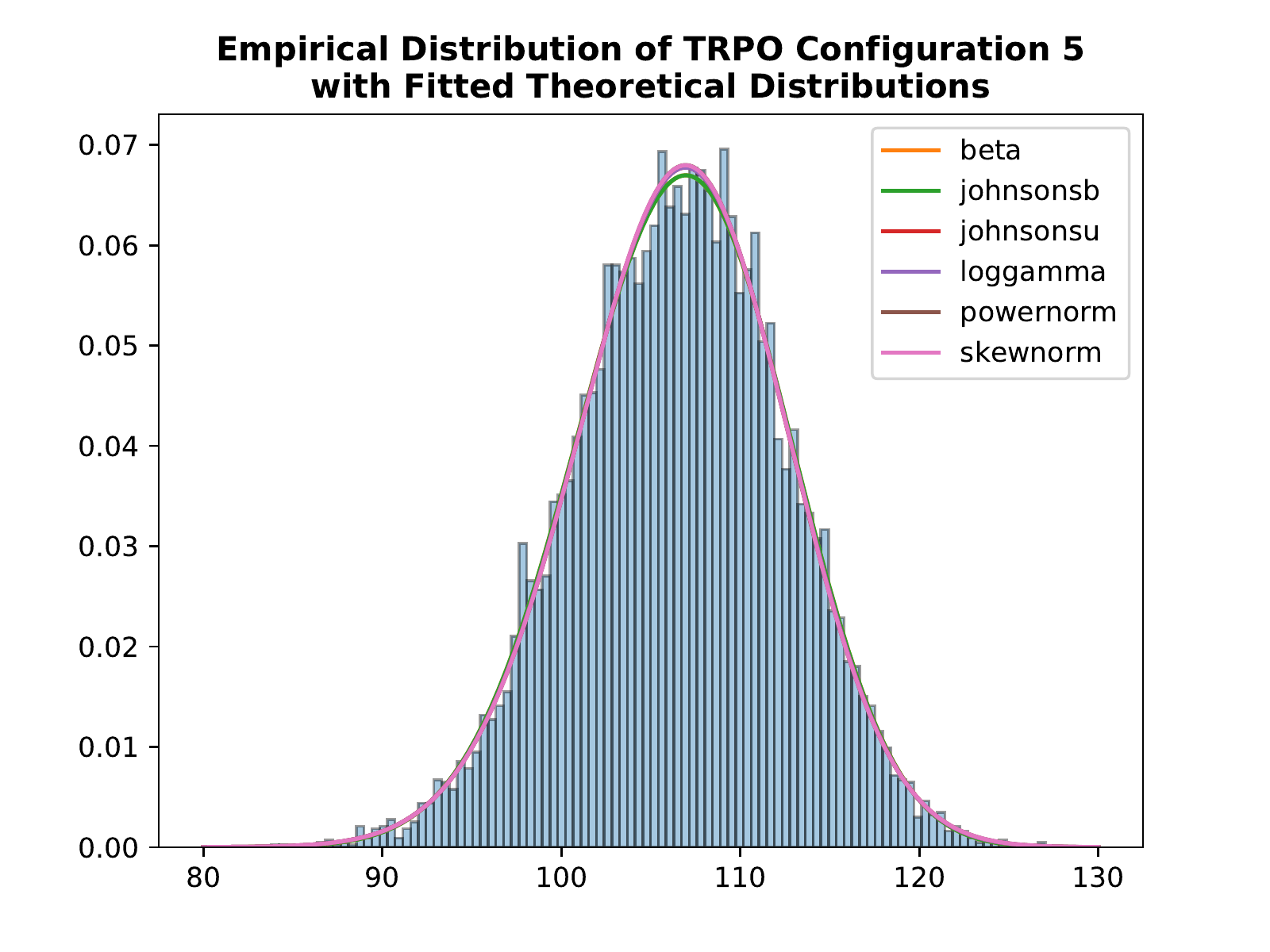}
    \end{minipage}
    \begin{minipage}{0.49\textwidth}
        \centering
        \includegraphics[width=1.0\textwidth,trim={10pt 0 35pt 0},clip]{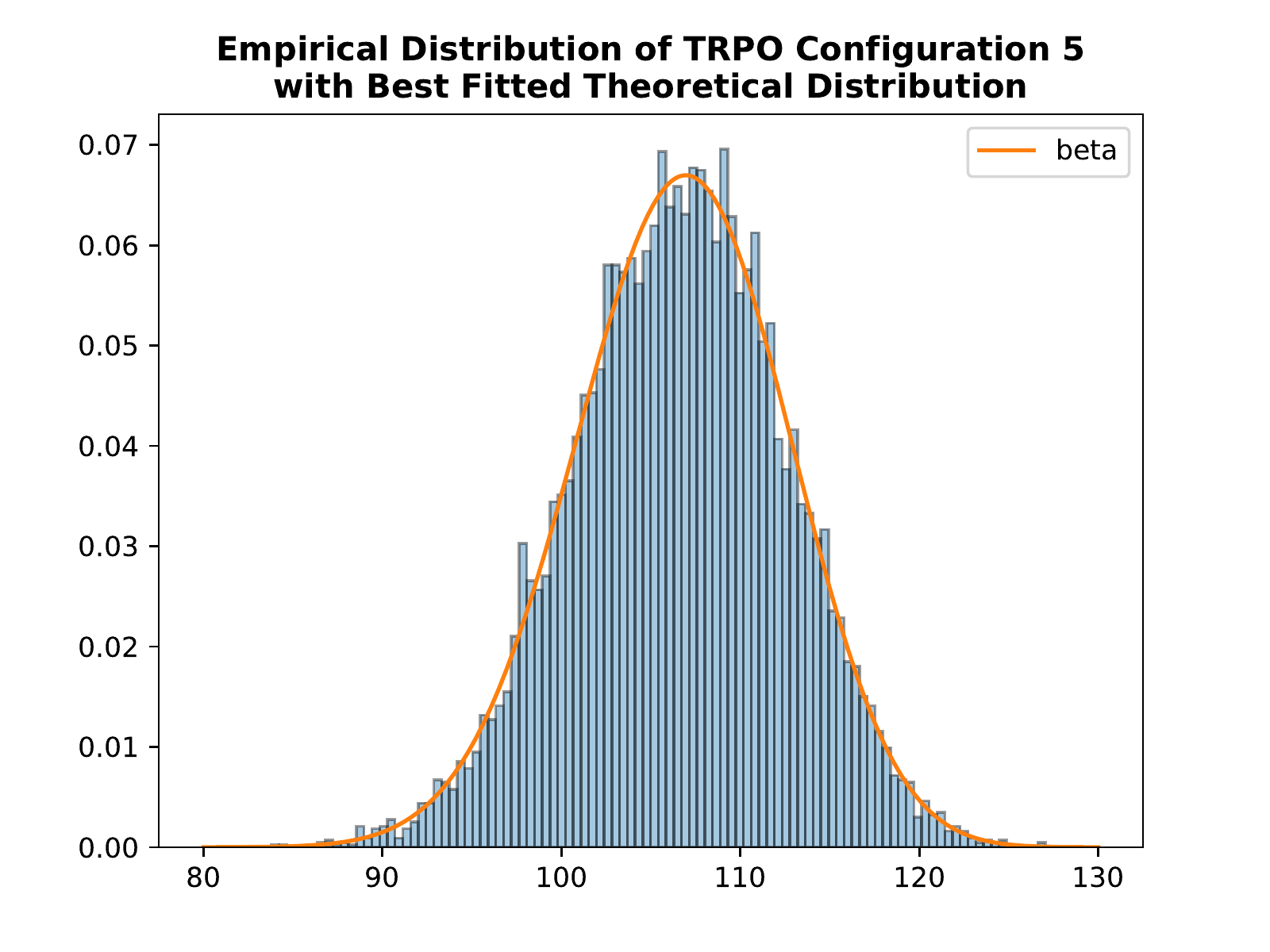}
    \end{minipage}
    \caption{\trpocaption}\label{fig:dist_fitting_trpo}
\end{figure}

%===============================================================================

\def \ppocaption {
        \textbf{Theoretical distribution fitting on empirical data from \gls*{ppo}:}   
        (\textit{top}) 52 theoretical distributions fitted to the empirical data of the \nth{1} configuration of \gls*{ppo},
        (\textit{left}) top-6 theoretical distributions (chosen based on $p$-value),
        (\textit{right}) best fitted theoretical distributions, each row corresponds to the hyperparameter configuration.
        }

\begin{figure}[p]
    \centering
    \begin{minipage}{0.49\textwidth}
        \centering
        \includegraphics[width=1.0\textwidth,trim={10pt 0 35pt 0},clip]{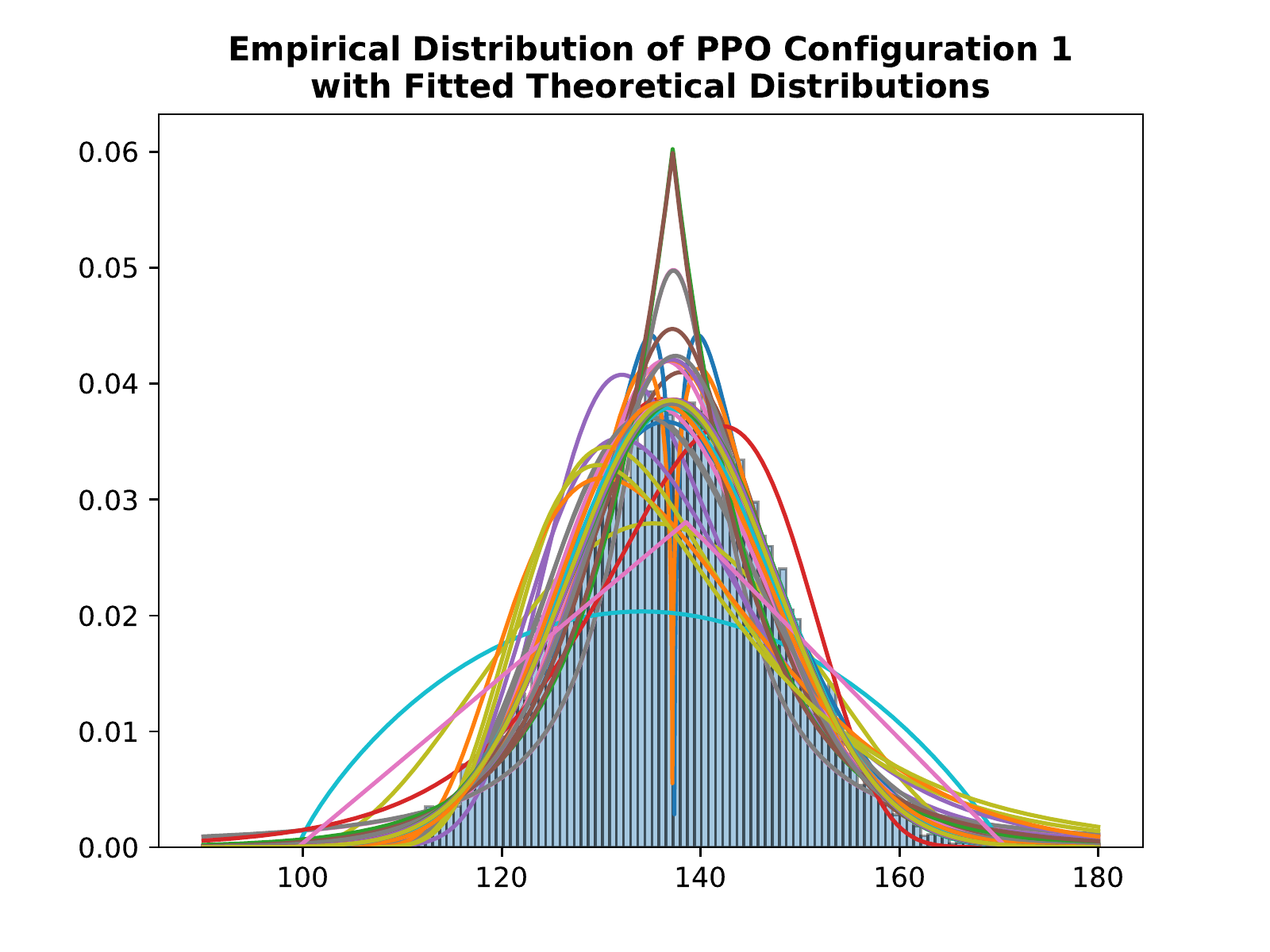}
    \end{minipage}\\
    
    \begin{minipage}{0.49\textwidth}
        \centering
        \includegraphics[width=1.0\textwidth,trim={10pt 0 35pt 0},clip]{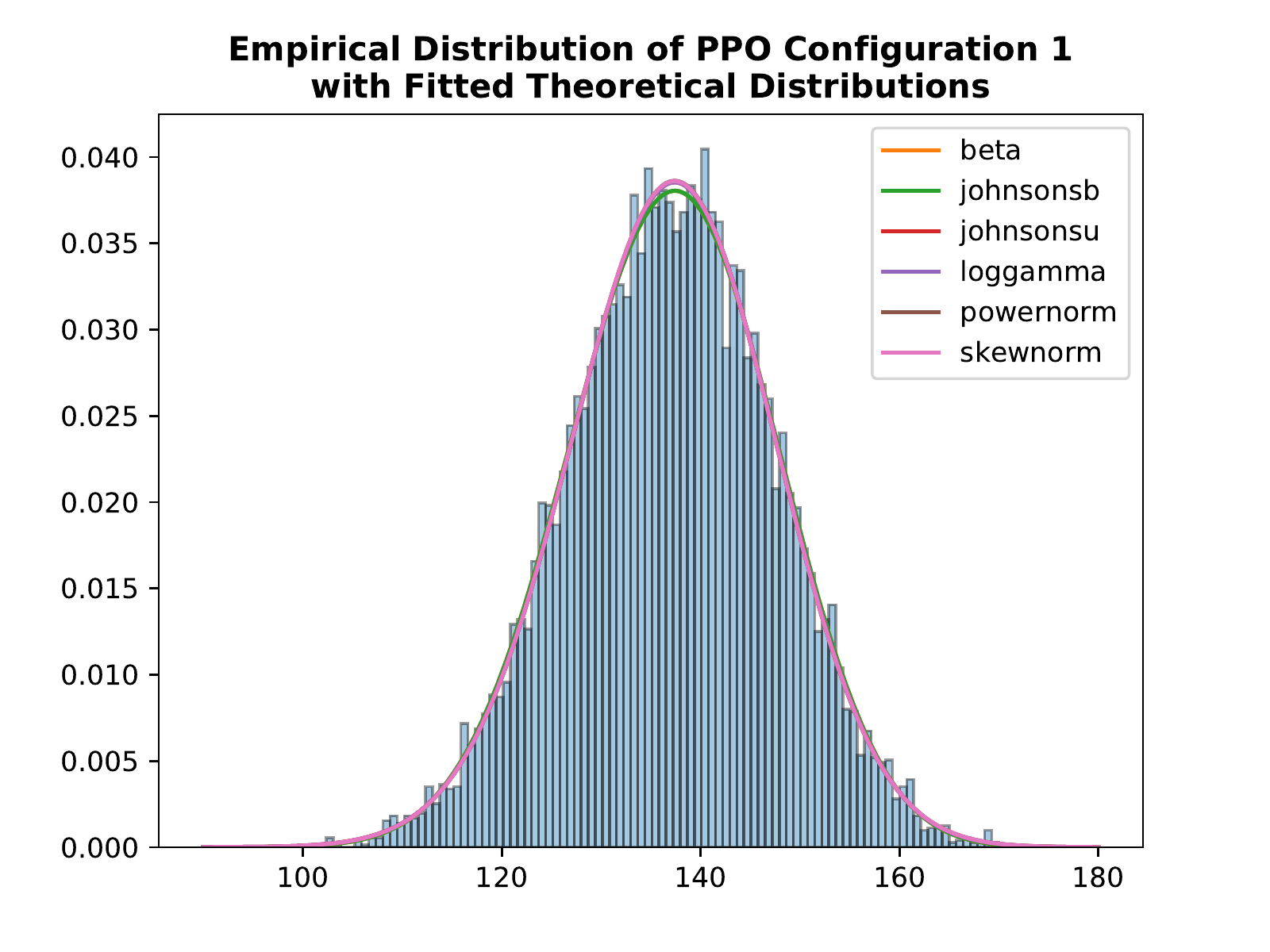}
    \end{minipage}
    \begin{minipage}{0.49\textwidth}
        \centering
        \includegraphics[width=1.0\textwidth,trim={10pt 0 35pt 0},clip]{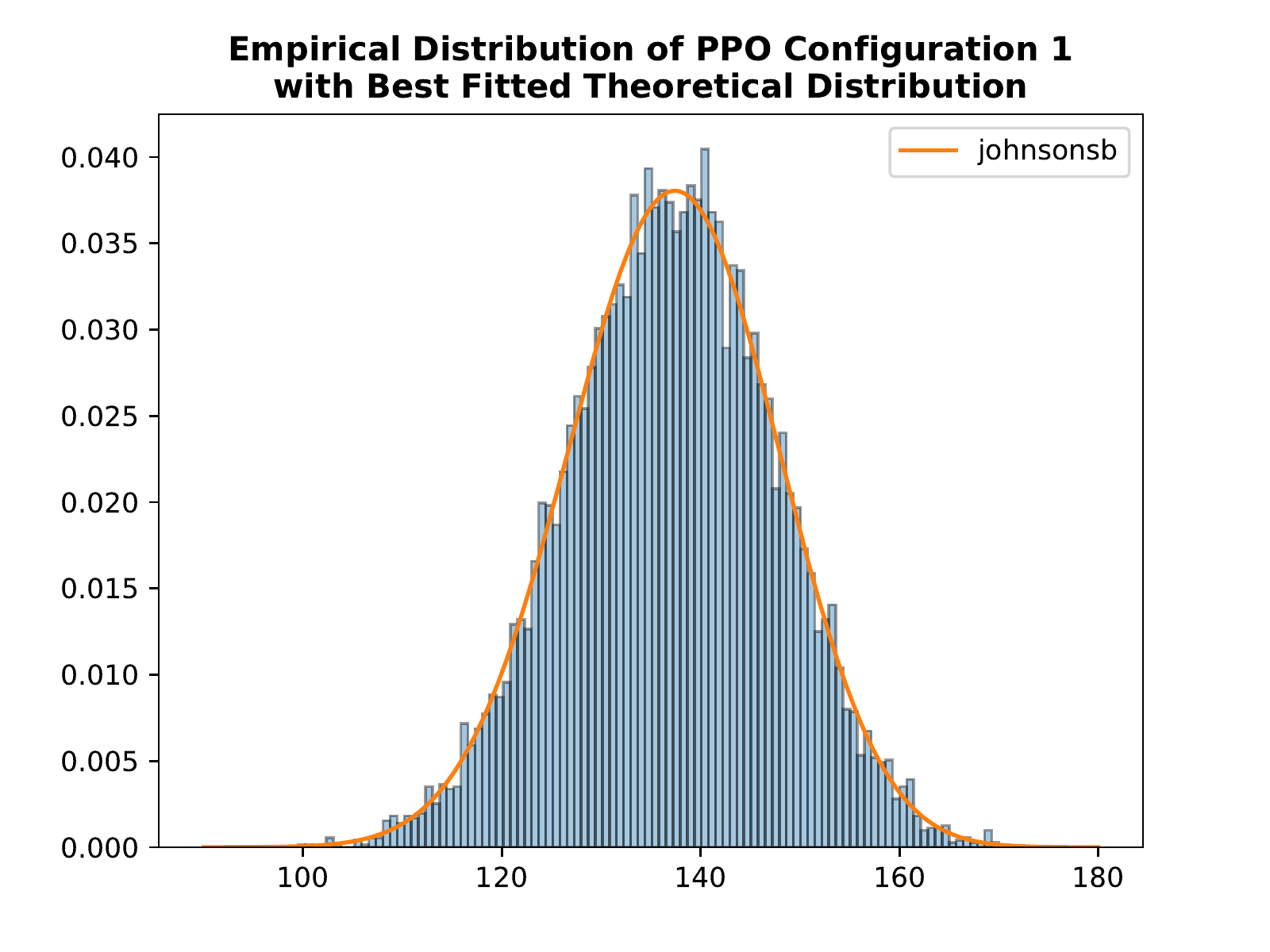}
    \end{minipage}
    \begin{minipage}{0.49\textwidth}
        \centering
        \includegraphics[width=1.0\textwidth,trim={10pt 0 35pt 0},clip]{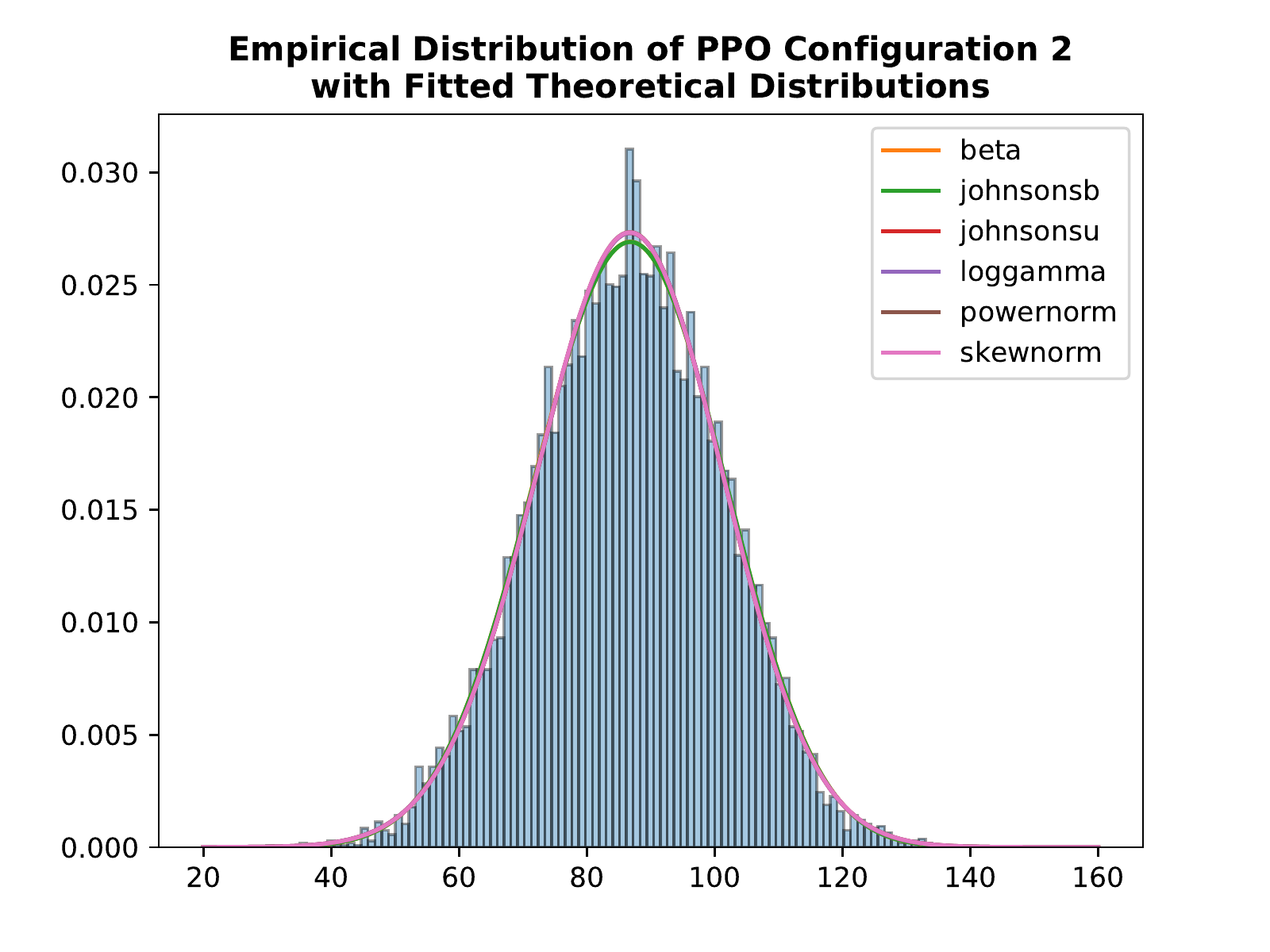}
    \end{minipage}
    \begin{minipage}{0.49\textwidth}
        \centering
        \includegraphics[width=1.0\textwidth,trim={10pt 0 35pt 0},clip]{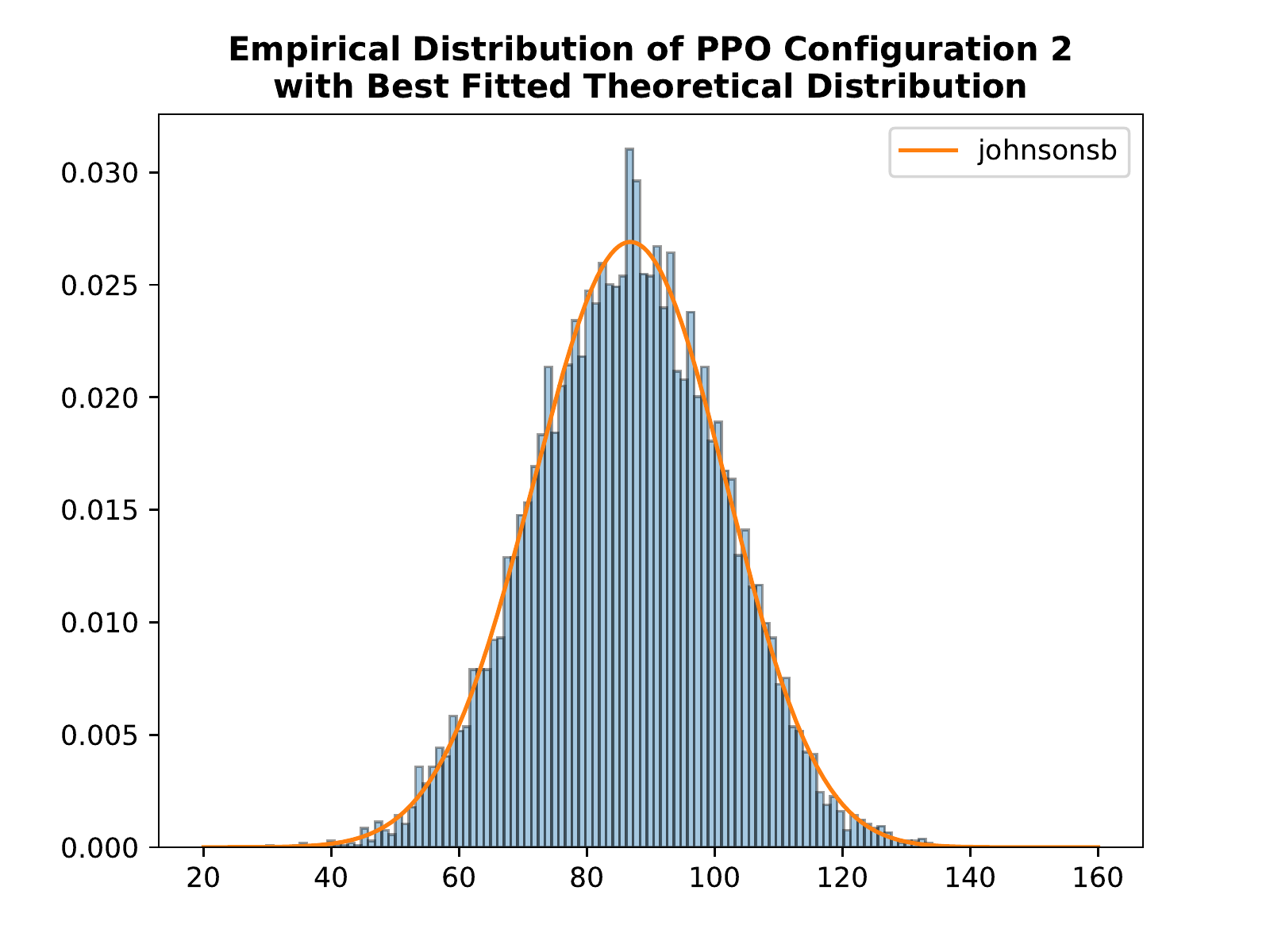}
    \end{minipage}
    \caption{\ppocaption {\tiny(figure continues on next page)}}
\end{figure}

\begin{figure}[p]\ContinuedFloat
    \centering
    \begin{minipage}{0.49\textwidth}
        \centering
        \includegraphics[width=1.0\textwidth,trim={10pt 0 35pt 0},clip]{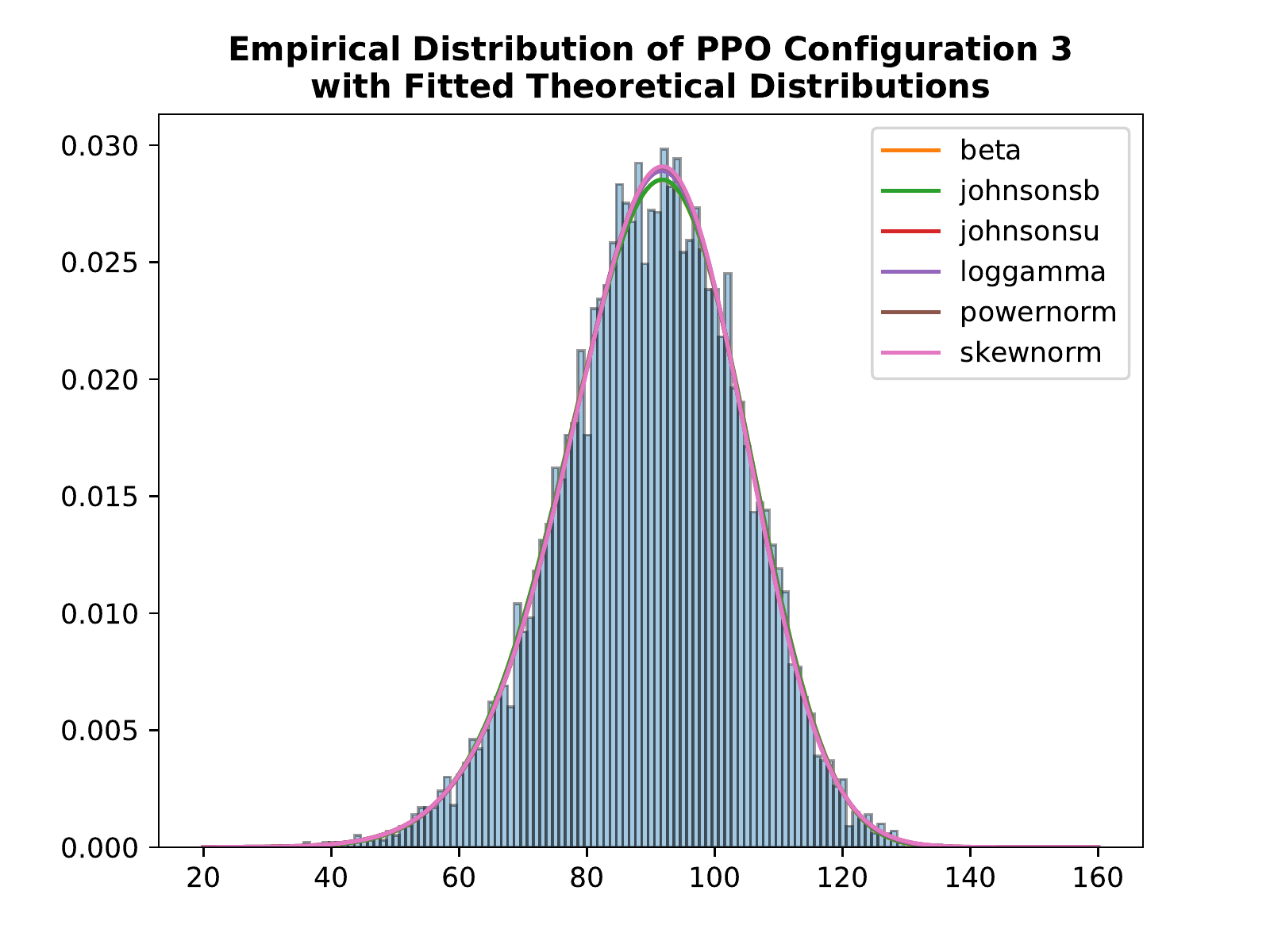}
    \end{minipage}
    \begin{minipage}{0.49\textwidth}
        \centering
        \includegraphics[width=1.0\textwidth,trim={10pt 0 35pt 0},clip]{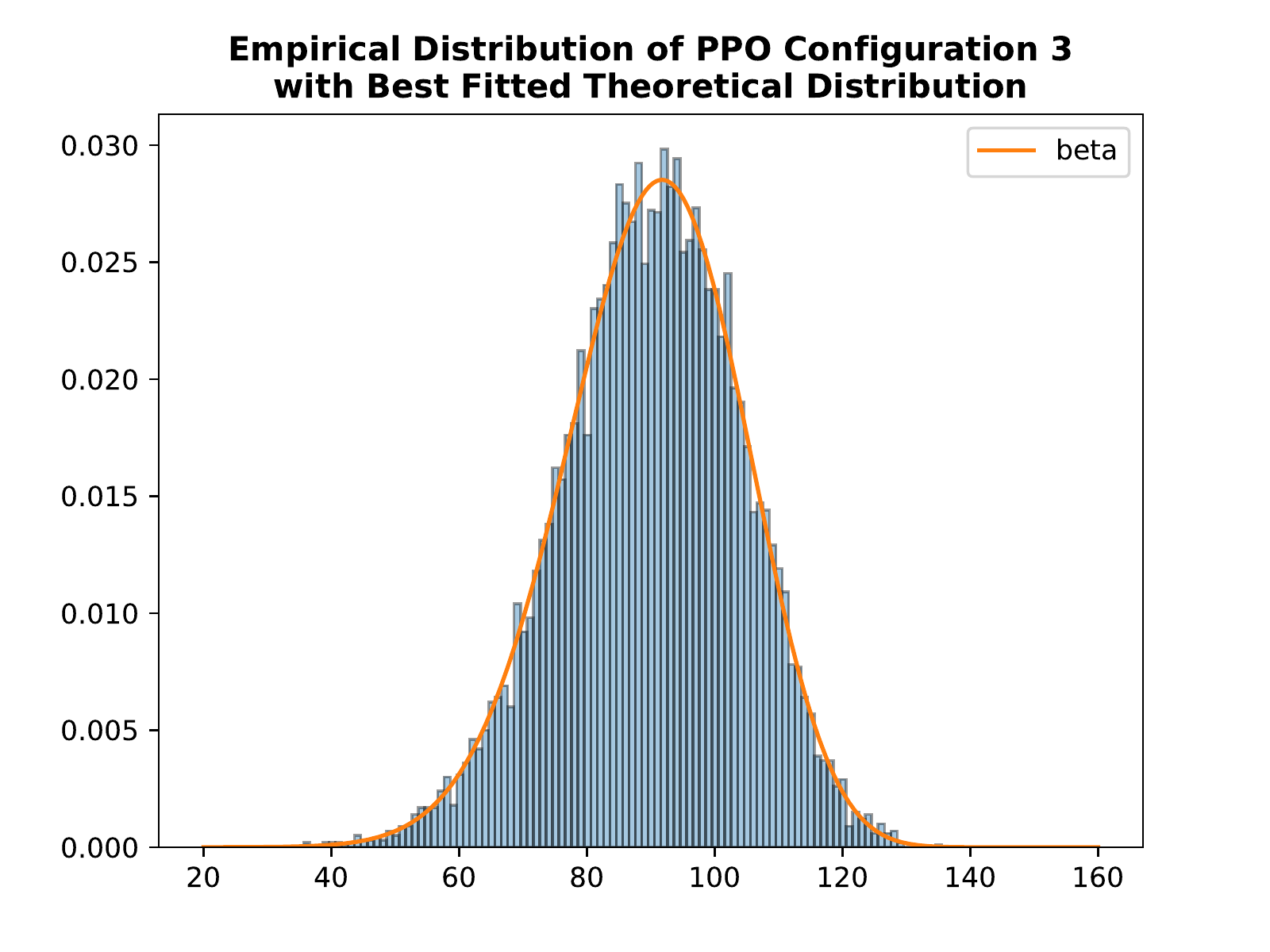}
    \end{minipage}
    \begin{minipage}{0.49\textwidth}
        \centering
        \includegraphics[width=1.0\textwidth,trim={10pt 0 35pt 0},clip]{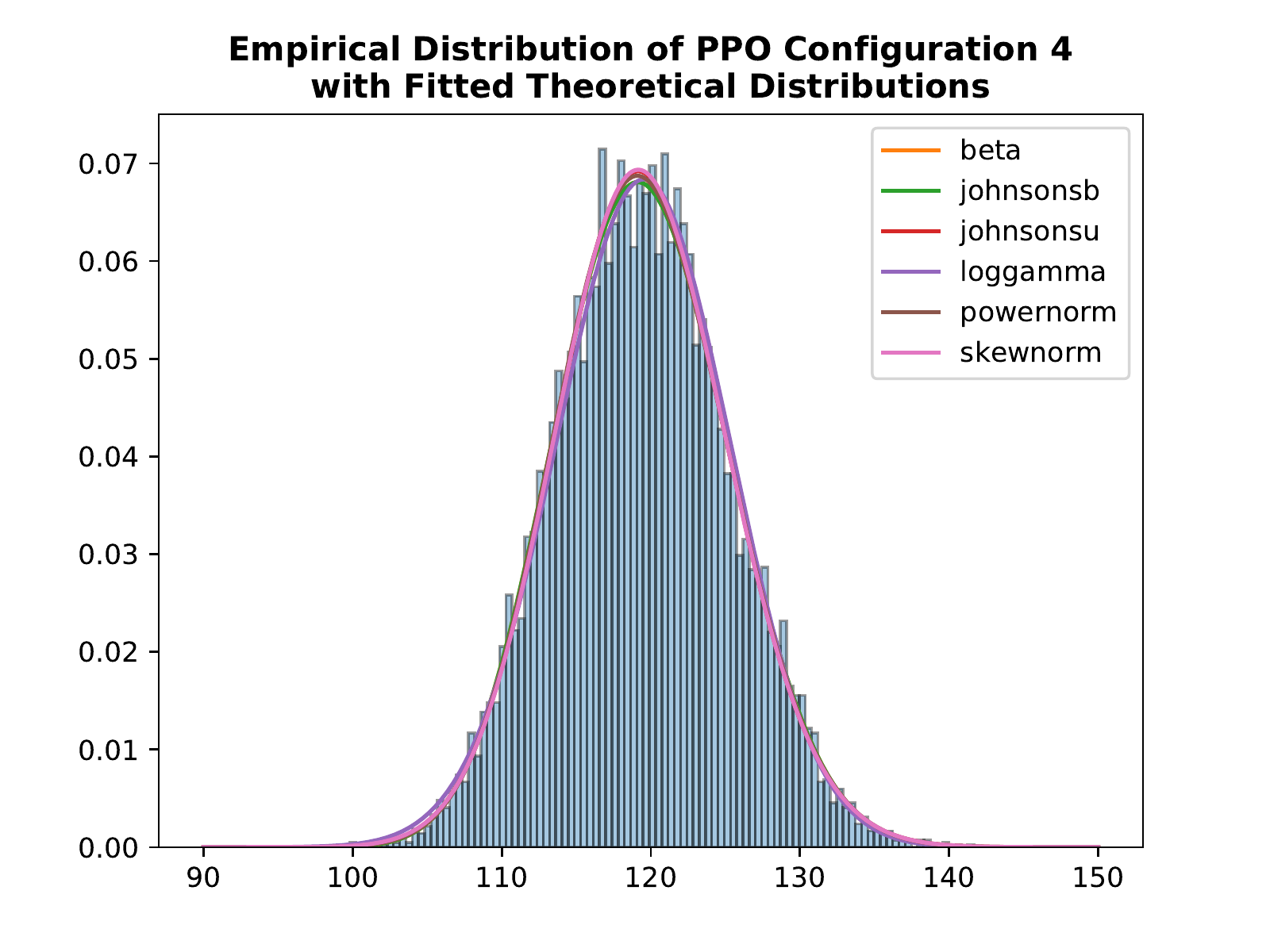}
    \end{minipage}
    \begin{minipage}{0.49\textwidth}
        \centering
        \includegraphics[width=1.0\textwidth,trim={10pt 0 35pt 0},clip]{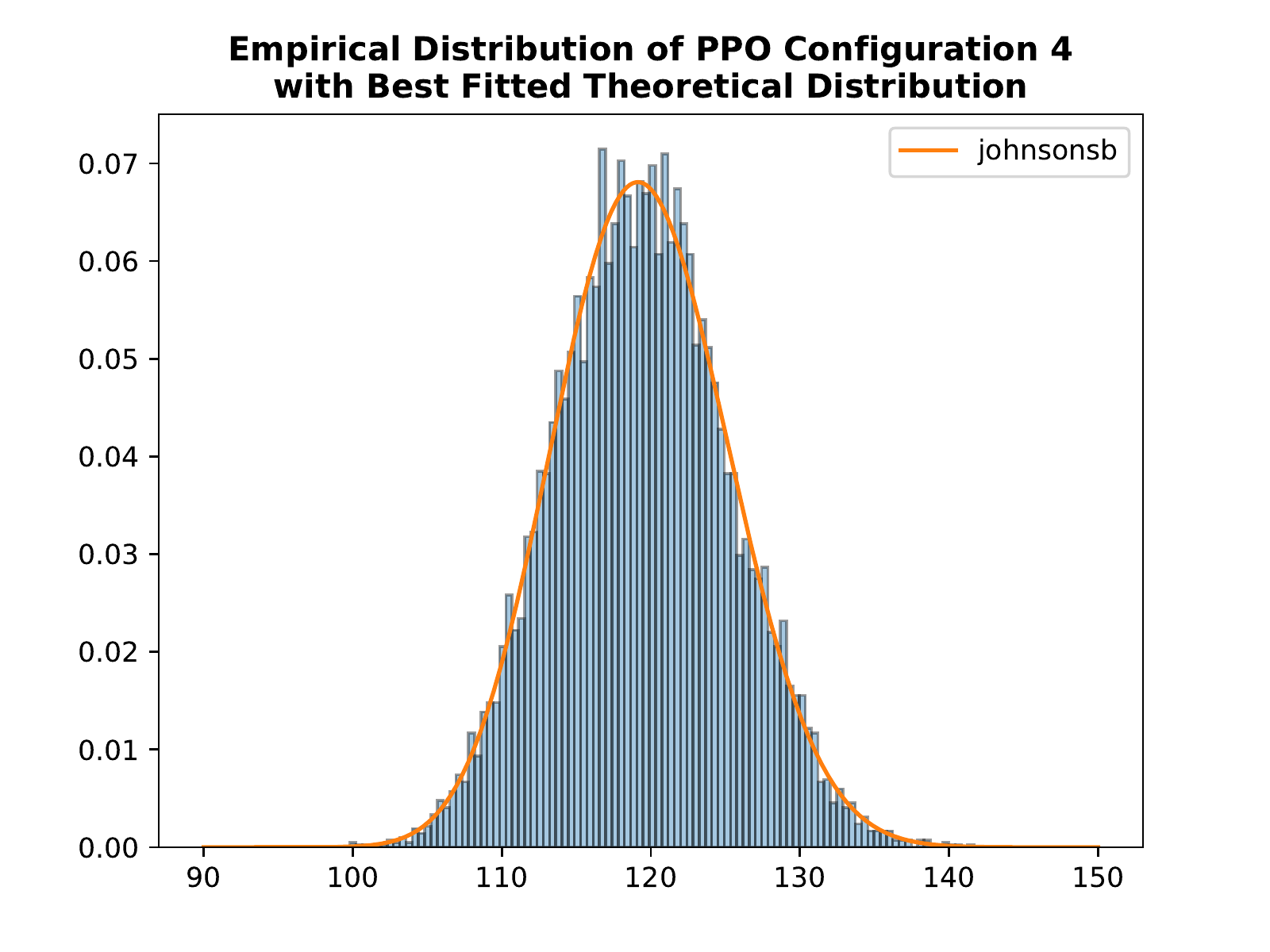}
    \end{minipage}
    \begin{minipage}{0.49\textwidth}
        \centering
        \includegraphics[width=1.0\textwidth,trim={10pt 0 35pt 0},clip]{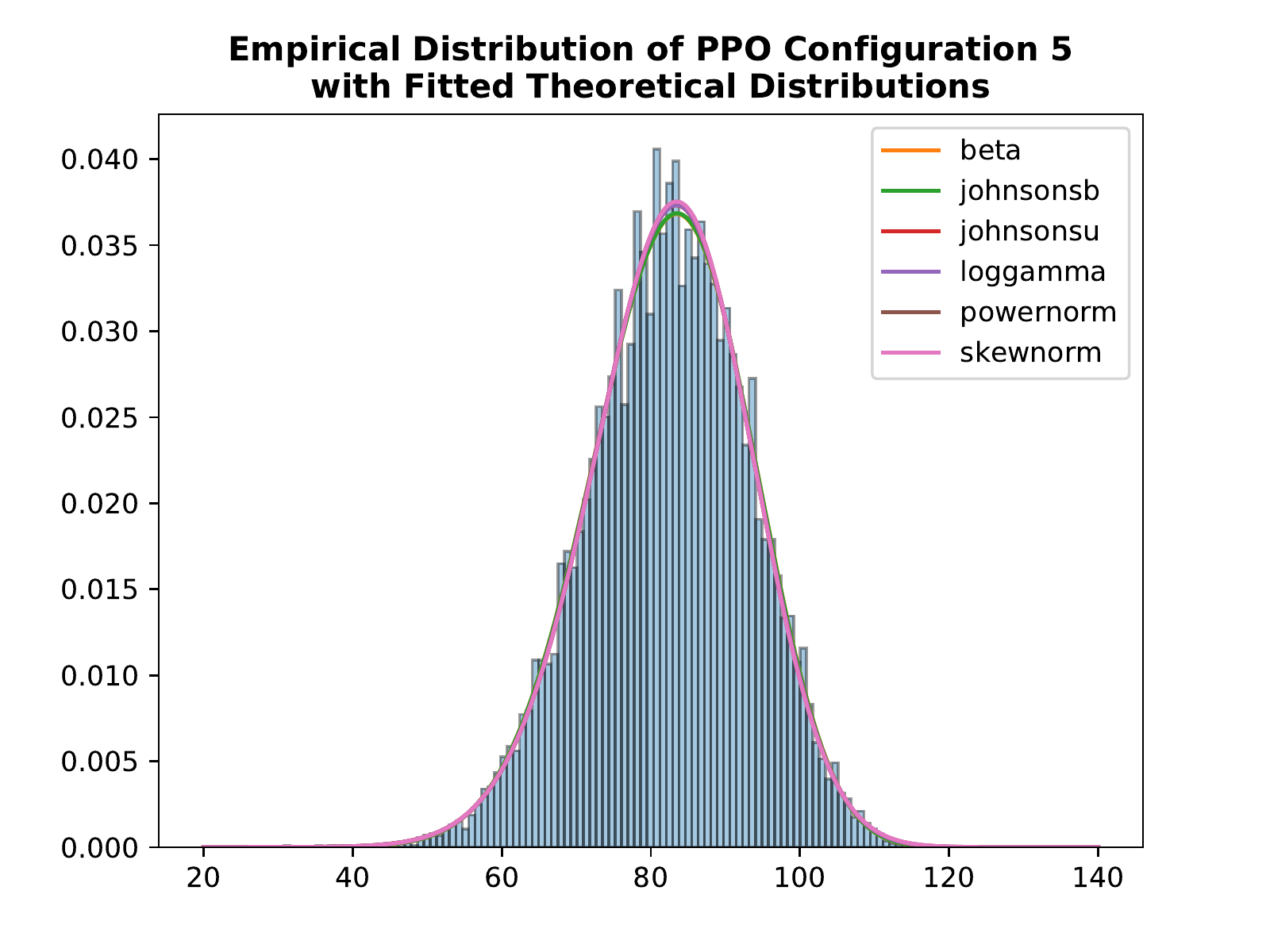}
    \end{minipage}
    \begin{minipage}{0.49\textwidth}
        \centering
        \includegraphics[width=1.0\textwidth,trim={10pt 0 35pt 0},clip]{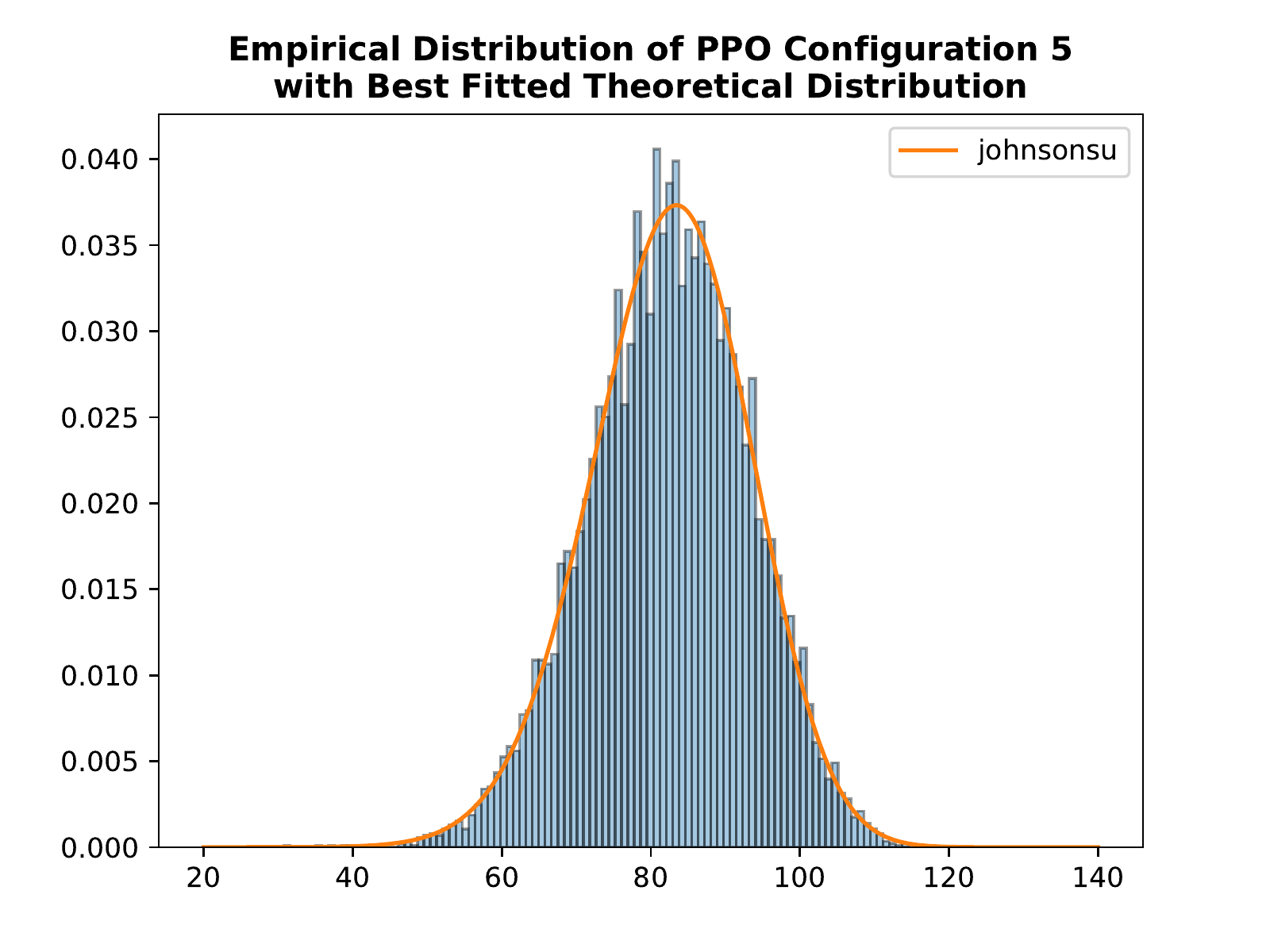}
    \end{minipage}
    \caption{\ppocaption}\label{fig:dist_fitting_ppo}
\end{figure}

%===============================================================================

\begin{table}[p]
    \centering
    \begin{tabular}{l l l Q L{7cm}}
        \toprule
        \multicolumn{5}{c}{\textbf{KS Test on TRPO}}\\
        & \multicolumn{1}{c}{Distribution}  &  \multicolumn{1}{c}{Statistics}  &  \multicolumn{1}{c}{$p$-value}  &  Distribution Parameters\\
        \midrule
        \parbox[t]{2mm}{\multirow{6}{*}{\rotatebox[origin=c]{90}{Configuration 1}}}
        & beta & 0.0082 & 0.5188 & (824.65, 167.66, -175.37, 374.38) \\
        & johnsonsb & 0.0082 & 0.5087 & (-7.13, 9.58, 3.31, 195.55) \\
        & johnsonsu & 0.0079 & 0.5644 & (10.94, 15.94, 178.02, 56.88) \\
        & loggamma & 0.0080 & 0.5406 & (79.15, -36.56, 39.48) \\
        & powernorm & 0.0075 & 0.6235 & (1.77, 138.22, 5.22) \\
        & skewnorm & 0.0075 & 0.6189 & (-0.92, 138.62, 5.29) \\
        \midrule
        \parbox[t]{2mm}{\multirow{6}{*}{\rotatebox[origin=c]{90}{Configuration 2}}}
        & beta & 0.0044 & 0.9911 & (18.83, 8.83, 89.22, 74.13) \\
        & johnsonsb & 0.0044 & 0.9903 & (-1.62, 2.71, 89.67, 78.02) \\
        & johnsonsu & 0.0075 & 0.6204 & (12.79, 8.57, 189.57, 23.46) \\
        & loggamma & 0.0074 & 0.6443 & (10.59, 92.24, 20.53) \\
        & powernorm & 0.0085 & 0.4630 & (5.39, 151.53, 9.81) \\
        & skewnorm & 0.0088 & 0.4167 & (-1.53, 145.51, 8.69) \\
        \midrule
        \parbox[t]{2mm}{\multirow{6}{*}{\rotatebox[origin=c]{90}{Configuration 3}}}
        & beta & 0.0058 & 0.8847 & (456.27, 282.03, -269.74, 618.35) \\
        & johnsonsb & 0.0083 & 0.4996 & (12132.2, 16581.66, -270975.92, 834554.97) \\
        & johnsonsu & 0.0061 & 0.8466 & (13.44, 32.78, 253.12, 333.52) \\
        & loggamma & 0.0060 & 0.8605 & (645.48, -1703.51, 280.7) \\
        & powernorm & 0.0063 & 0.8245 & (1.2, 114.22, 11.64) \\
        & skewnorm & 0.0062 & 0.8343 & (-0.58, 117.21, 12.05) \\
        \midrule
        \parbox[t]{2mm}{\multirow{6}{*}{\rotatebox[origin=c]{90}{Configuration 4}}}
        & beta & 0.0064 & 0.8108 & (22.36, 12.28, 77.62, 31.58) \\
        & johnsonsb & 0.0064 & 0.8028 & (-1.5, 3.17, 76.79, 34.56) \\
        & johnsonsu & 0.0071 & 0.6874 & (14.6, 11.91, 123.33, 16.21) \\
        & loggamma & 0.0073 & 0.6683 & (22.52, 61.28, 11.88) \\
        & powernorm & 0.0079 & 0.5575 & (2.92, 100.79, 3.36) \\
        & skewnorm & 0.0134 & 0.0555 & (0.0, 98.01, 2.53) \\
        \midrule
        \parbox[t]{2mm}{\multirow{6}{*}{\rotatebox[origin=c]{90}{Configuration 5}}}
        & beta & 0.0053 & 0.9402 & (41.71, 27.17, 45.53, 100.9) \\
        & johnsonsb & 0.0053 & 0.9391 & (-1.53, 4.6, 41.09, 112.69) \\
        & johnsonsu & 0.0073 & 0.6597 & (18.73, 20.62, 194.25, 84.29) \\
        & loggamma & 0.0073 & 0.6580 & (88.61, -141.39, 55.38) \\
        & powernorm & 0.0078 & 0.5762 & (1.67, 109.54, 6.81) \\
        & skewnorm & 0.0077 & 0.5887 & (-0.87, 110.27, 6.93) \\
        \midrule
    \end{tabular}
    \caption{\textbf{Probability values obtained by \gls*{ks} test:} The table shows the $p$-values from fitting six theoretical distributions to our \glspl*{edf}.}
\end{table}

\begin{table}[p]
    \centering
    \begin{tabular}{l l l Q L{7cm}}
        \toprule
        \multicolumn{5}{c}{\textbf{KS Test on PPO}}\\
        & \multicolumn{1}{c}{Distribution}  &  \multicolumn{1}{c}{Statistics}  &  \multicolumn{1}{c}{$p$-value}  &  Distribution Parameters\\
        \midrule
        \parbox[t]{2mm}{\multirow{6}{*}{\rotatebox[origin=c]{90}{Configuration 1}}}
        & beta & 0.0032 & 1.0000 & (33.33, 26.22, 46.19, 162.37) \\
        & johnsonsb & 0.0031 & 1.0000 & (-0.83, 4.37, 35.88, 185.12) \\
        & johnsonsu & 0.0045 & 0.9886 & (17.75, 27.5, 299.04, 234.23) \\
        & loggamma & 0.0043 & 0.9925 & (240.56, -742.66, 160.51) \\
        & powernorm & 0.0047 & 0.9782 & (1.35, 139.96, 11.29) \\
        & skewnorm & 0.0048 & 0.9762 & (-0.7, 142.42, 11.66) \\
        \midrule
        \parbox[t]{2mm}{\multirow{6}{*}{\rotatebox[origin=c]{90}{Configuration 2}}}
        & beta & 0.0058 & 0.8880 & (27.36, 24.02, -26.37, 211.93) \\
        & johnsonsb & 0.0058 & 0.8937 & (-0.42, 4.05, -40.1, 240.88) \\
        & johnsonsu & 0.0082 & 0.5115 & (18.66, 37.94, 339.16, 493.35) \\
        & loggamma & 0.0082 & 0.5114 & (637.2, -2293.83, 368.68) \\
        & powernorm & 0.0086 & 0.4428 & (1.17, 88.64, 15.31) \\
        & skewnorm & 0.0086 & 0.4487 & (-0.55, 92.63, 15.85) \\
        \midrule
        \parbox[t]{2mm}{\multirow{6}{*}{\rotatebox[origin=c]{90}{Configuration 3}}}
        & beta & 0.0067 & 0.7663 & (53.3, 20.5, -104.13, 268.91) \\
        & johnsonsb & 0.0068 & 0.7434 & (-3.04, 4.28, -91.49, 271.58) \\
        & johnsonsu & 0.0079 & 0.5618 & (13.09, 10.61, 214.53, 78.87) \\
        & loggamma & 0.0076 & 0.6067 & (18.11, -77.65, 58.47) \\
        & powernorm & 0.0084 & 0.4859 & (3.43, 108.01, 19.21) \\
        & skewnorm & 0.0086 & 0.4449 & (-1.3, 101.47, 17.99) \\
        \midrule
        \parbox[t]{2mm}{\multirow{6}{*}{\rotatebox[origin=c]{90}{Configuration 4}}}
        & beta & 0.0049 & 0.9699 & (20.79, 29.82, 84.78, 84.4) \\
        & johnsonsb & 0.0048 & 0.9733 & (1.12, 3.95, 78.74, 94.5) \\
        & johnsonsu & 0.0068 & 0.7430 & (-15.88, 19.61, 43.62, 84.03) \\
        & loggamma & 0.0101 & 0.2577 & (877.15, -1051.47, 172.8) \\
        & powernorm & 0.0063 & 0.8236 & (0.57, 116.73, 4.84) \\
        & skewnorm & 0.0074 & 0.6364 & (0.86, 115.92, 6.77) \\
        \midrule
        \parbox[t]{2mm}{\multirow{6}{*}{\rotatebox[origin=c]{90}{Configuration 5}}}
        & beta & 0.0067 & 0.7584 & (42.62, 20.27, -42.17, 183.85) \\
        & johnsonsb & 0.0067 & 0.7532 & (-2.49, 4.28, -46.6, 201.73) \\
        & johnsonsu & 0.0063 & 0.8247 & (13.87, 12.67, 191.14, 81.67) \\
        & loggamma & 0.0064 & 0.8040 & (27.53, -101.83, 55.89) \\
        & powernorm & 0.0072 & 0.6841 & (2.61, 92.9, 13.91) \\
        & skewnorm & 0.0071 & 0.6902 & (-1.15, 90.53, 13.46) \\
        \midrule
    \end{tabular}
    \caption{\textbf{Probability values obtained by \gls*{ks} test:} The table shows the $p$-values from fitting six theoretical distributions to our \glspl*{edf}.}
\end{table}

  \newpage
  \section{Normality Test of Empirical Distributions}\label{app:norm_empirical_dists}
  To test if the empirical distributions, created by bootstrapping, are normally distributed we conduct a normality test (based on D'Agostino and Pearson's test \cite{d1971omnibus, d1973tests}) which rejects 8/10 of our null--hypothesis, $\mathcal{H}_0$. Thus we cannot assume that the empirical distributions in figure \ref{fig:empirical_distributions}. The resulting $p$-values from the normality test is presented in table \ref{tab:normality_hypothesis_tests}.

\def\trponormalcaption{
    \textbf{Empirical distributions obtained by bootstrapping:} (\textit{top}) \gls*{trpo} and (\textit{bottom}) \gls*{ppo}. All ten hyper-parameter configurations are represented in the histograms with a normal distribution fitted to it. Note that even though the data appears normally distributed, our normality test rejects 8/10 of our null--hypothesis, meaning that the samples were not normally distributed. See table \ref{tab:normality_hypothesis_tests} for corresponding $p$--values.
}

\begin{figure}[h!]
    \centering
    \begin{minipage}{0.49\textwidth}
        \centering
        \includegraphics[width=1.0\textwidth,trim={10pt 0 35pt 0},clip]{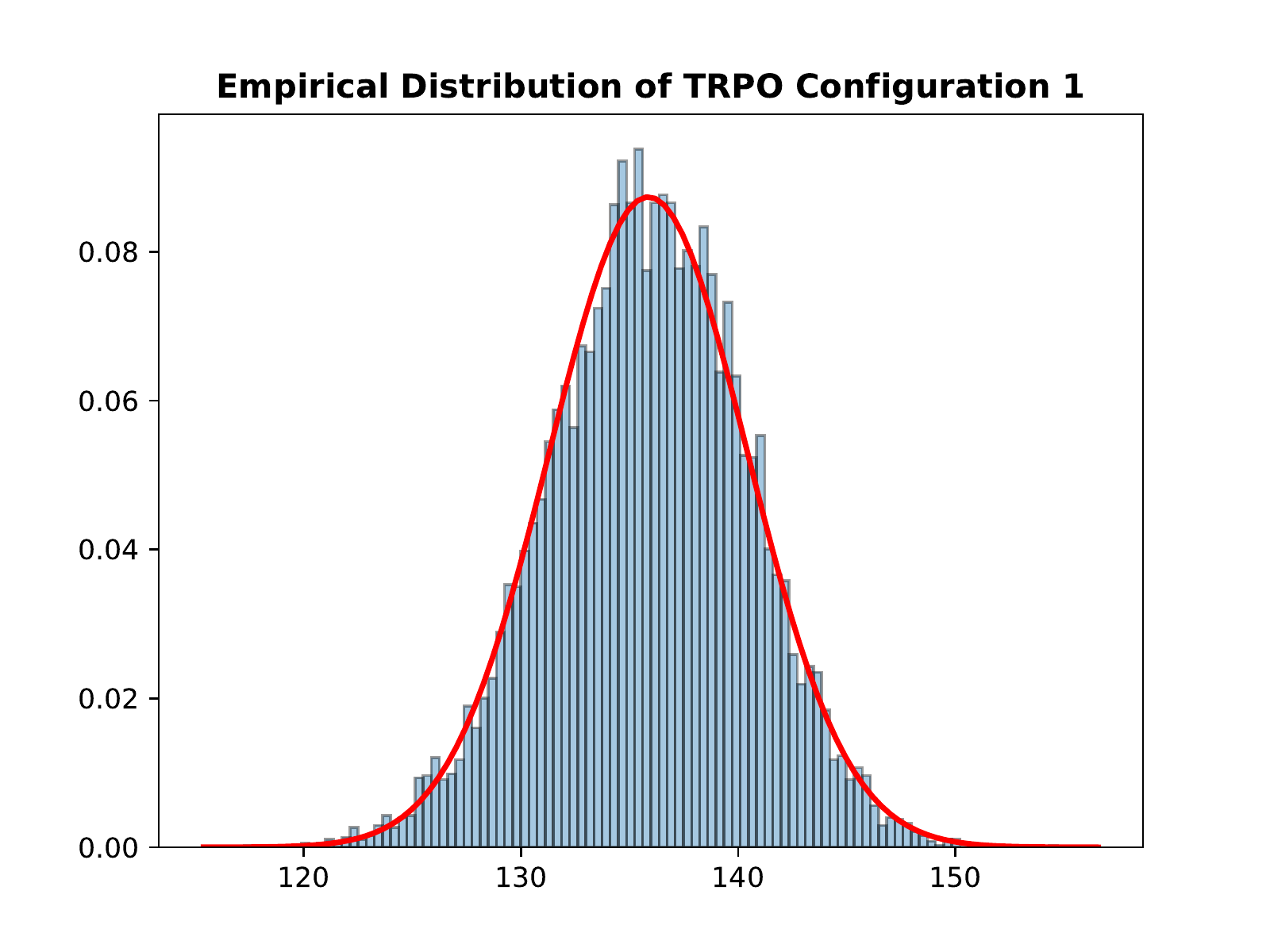}
    \end{minipage}\hfill
    \begin{minipage}{0.49\textwidth}
        \centering
        \includegraphics[width=1.0\textwidth,trim={10pt 0 35pt 0},clip]{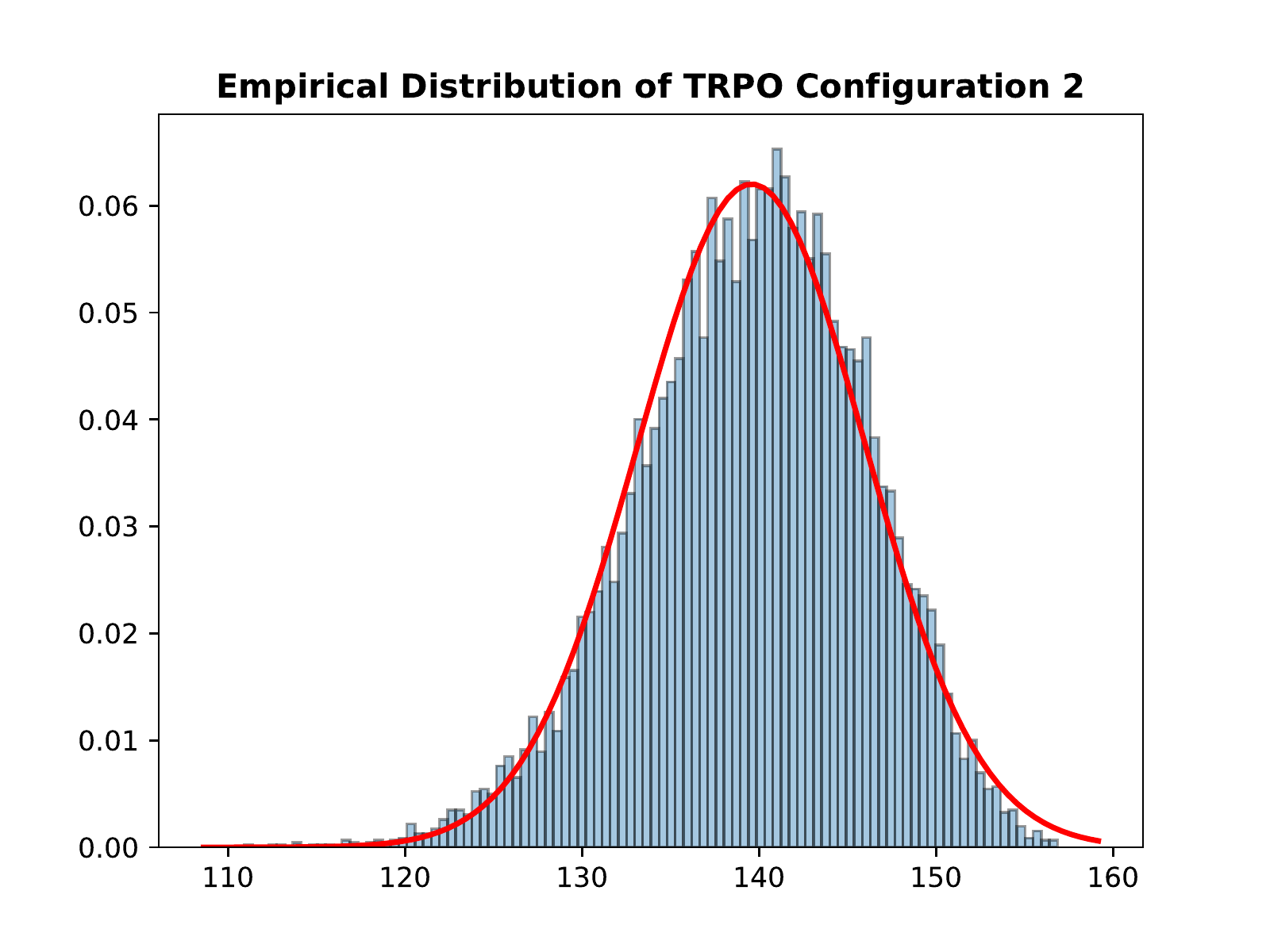}
    \end{minipage}
    \begin{minipage}{0.49\textwidth}
        \centering
        \includegraphics[width=1.0\textwidth,trim={10pt 0 35pt 0},clip]{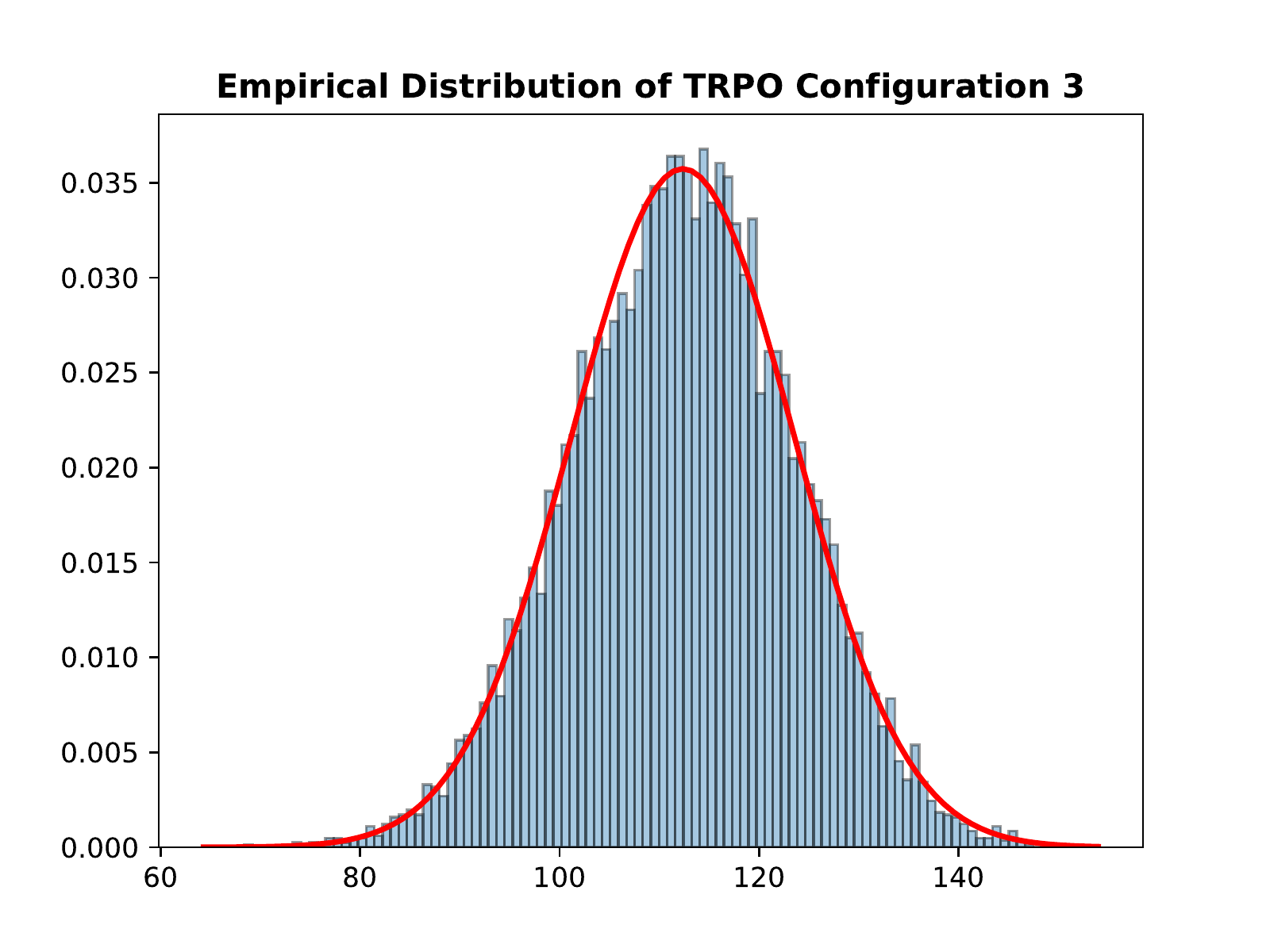}
    \end{minipage}\hfill
    \begin{minipage}{0.49\textwidth}
        \centering
        \includegraphics[width=1.0\textwidth,trim={10pt 0 35pt 0},clip]{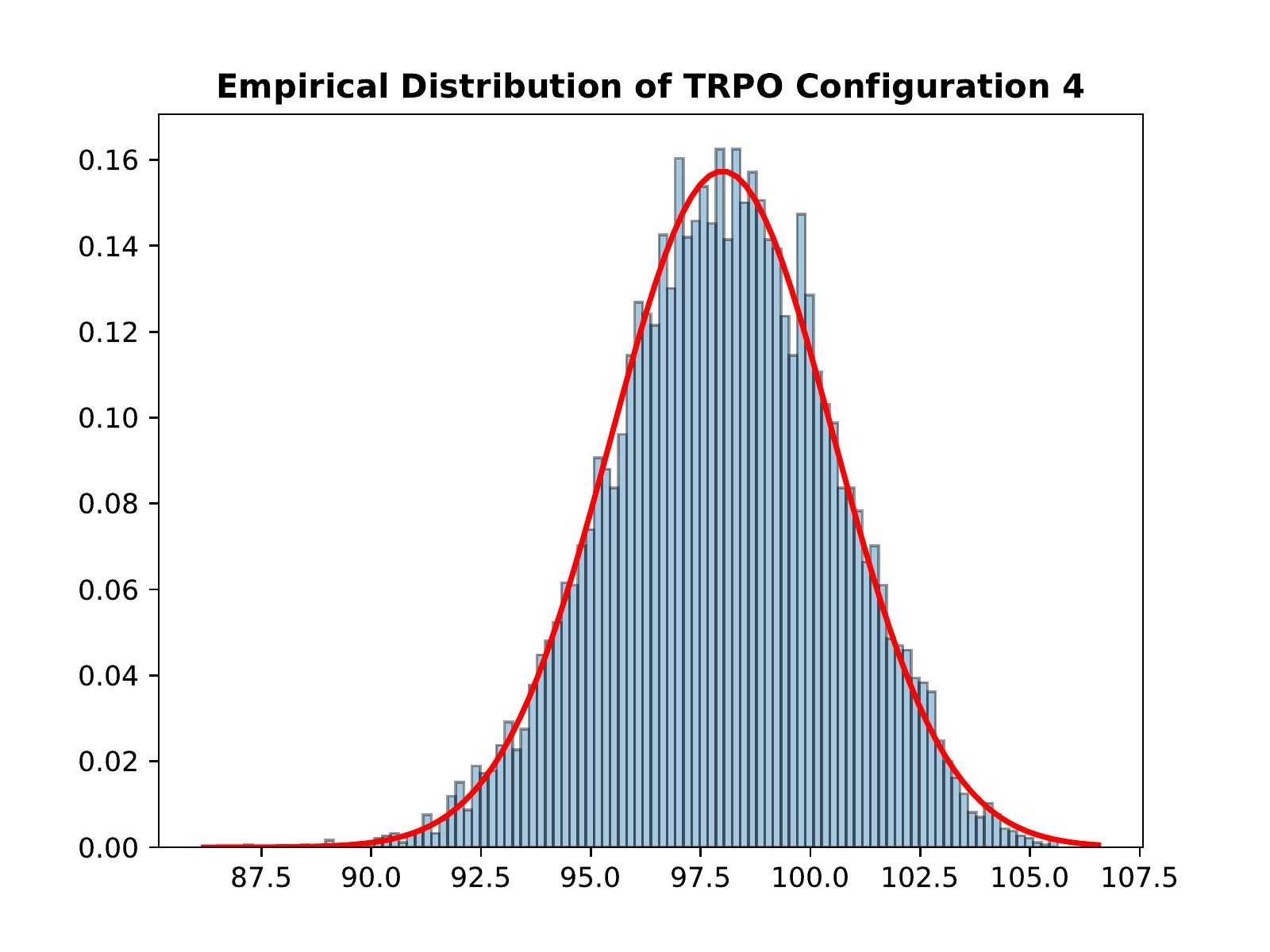}
    \end{minipage}
    \begin{minipage}{0.49\textwidth}
        \centering
        \includegraphics[width=1.0\textwidth,trim={10pt 0 35pt 0},clip]{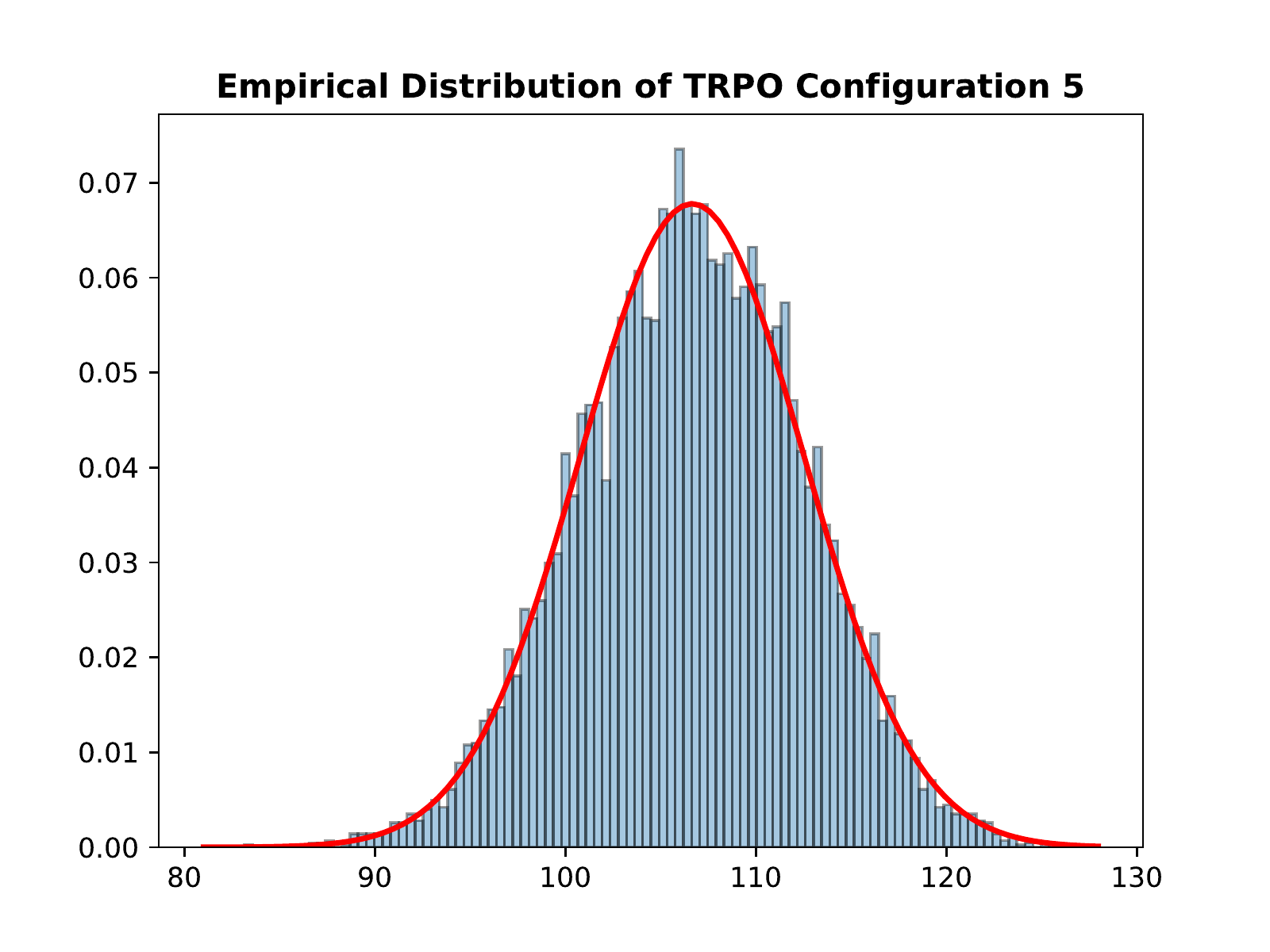}
    \end{minipage}
    \caption{\trponormalcaption {\tiny(figure continues on next page)}}
\end{figure}

\begin{figure}[p]\ContinuedFloat
    \centering
    \begin{minipage}{0.49\textwidth}
        \centering
        \includegraphics[width=1.0\textwidth,trim={10pt 0 35pt 0},clip]{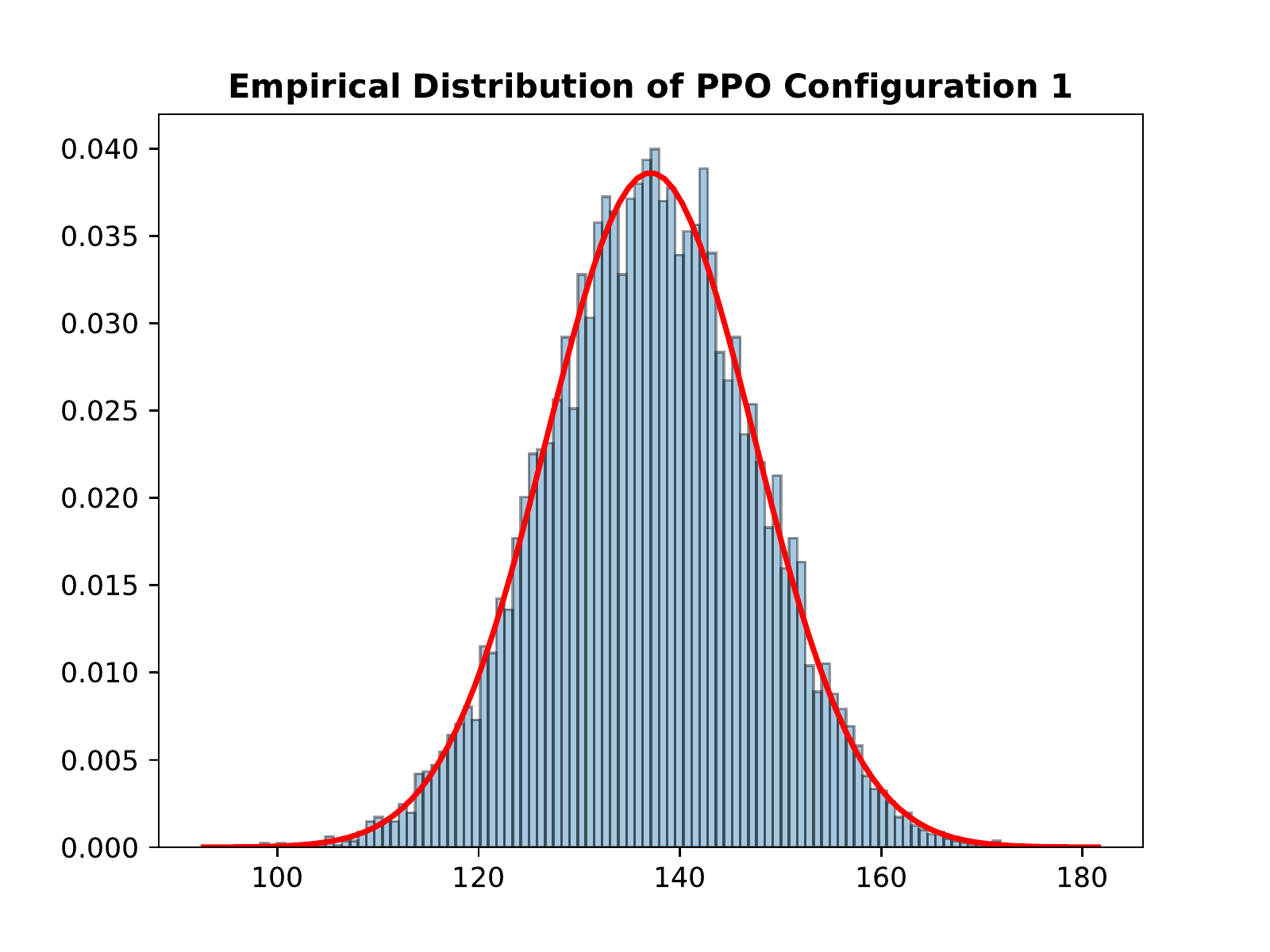}
    \end{minipage}\hfill
    \begin{minipage}{0.49\textwidth}
        \centering
        \includegraphics[width=1.0\textwidth,trim={10pt 0 35pt 0},clip]{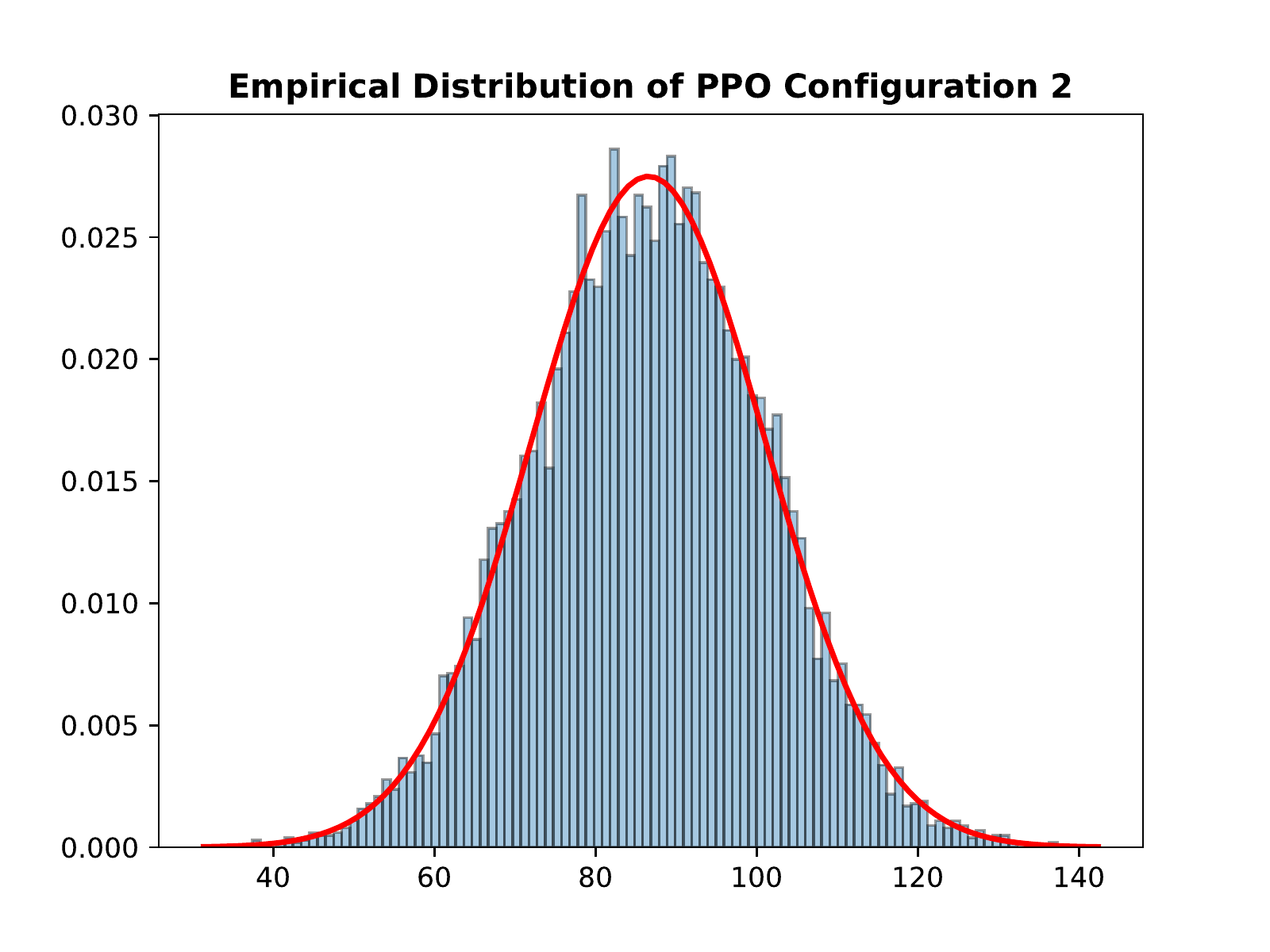}
    \end{minipage}
    \begin{minipage}{0.49\textwidth}
        \centering
        \includegraphics[width=1.0\textwidth,trim={10pt 0 35pt 0},clip]{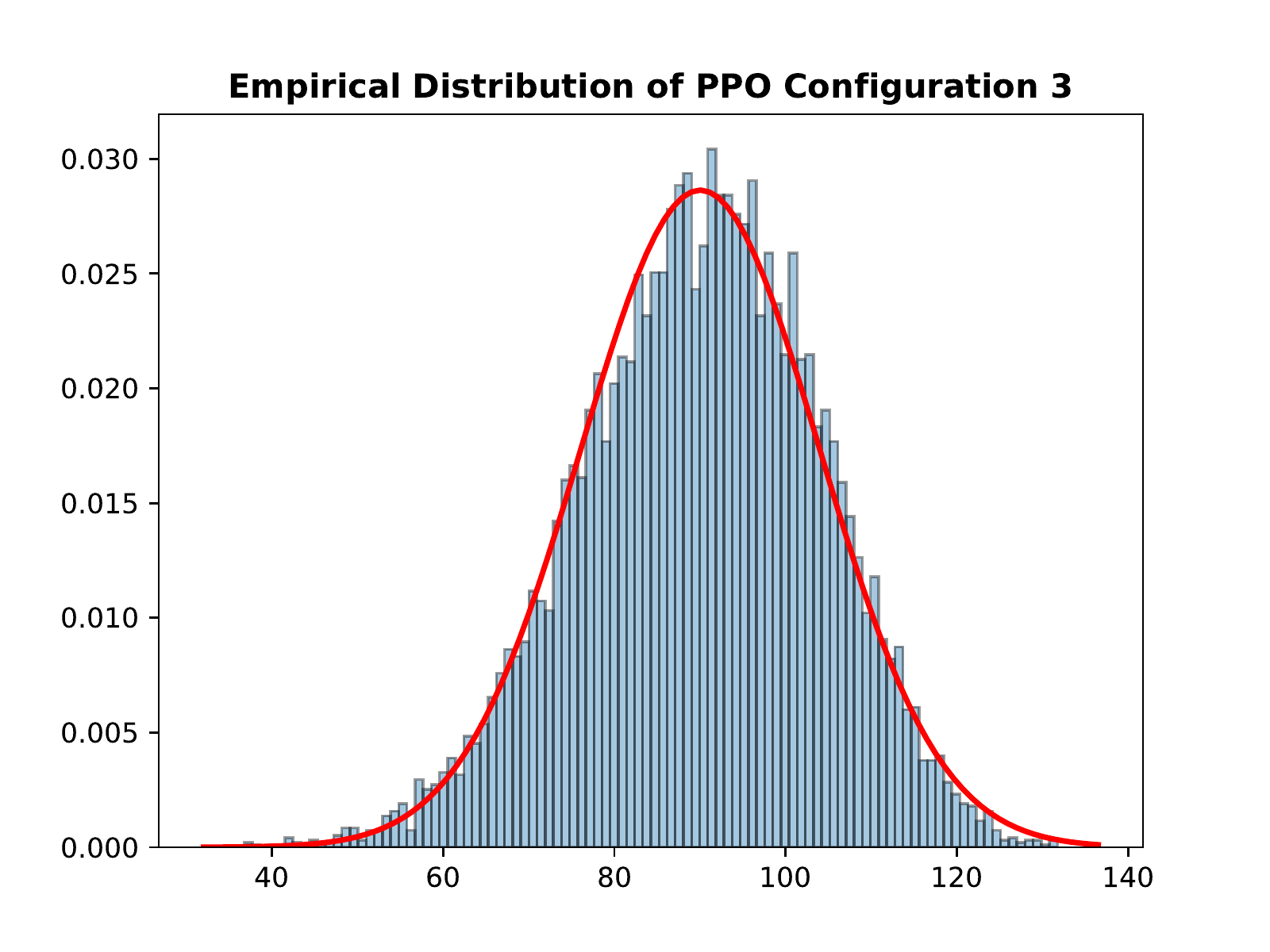}
    \end{minipage}\hfill
    \begin{minipage}{0.49\textwidth}
        \centering
        \includegraphics[width=1.0\textwidth,trim={10pt 0 35pt 0},clip]{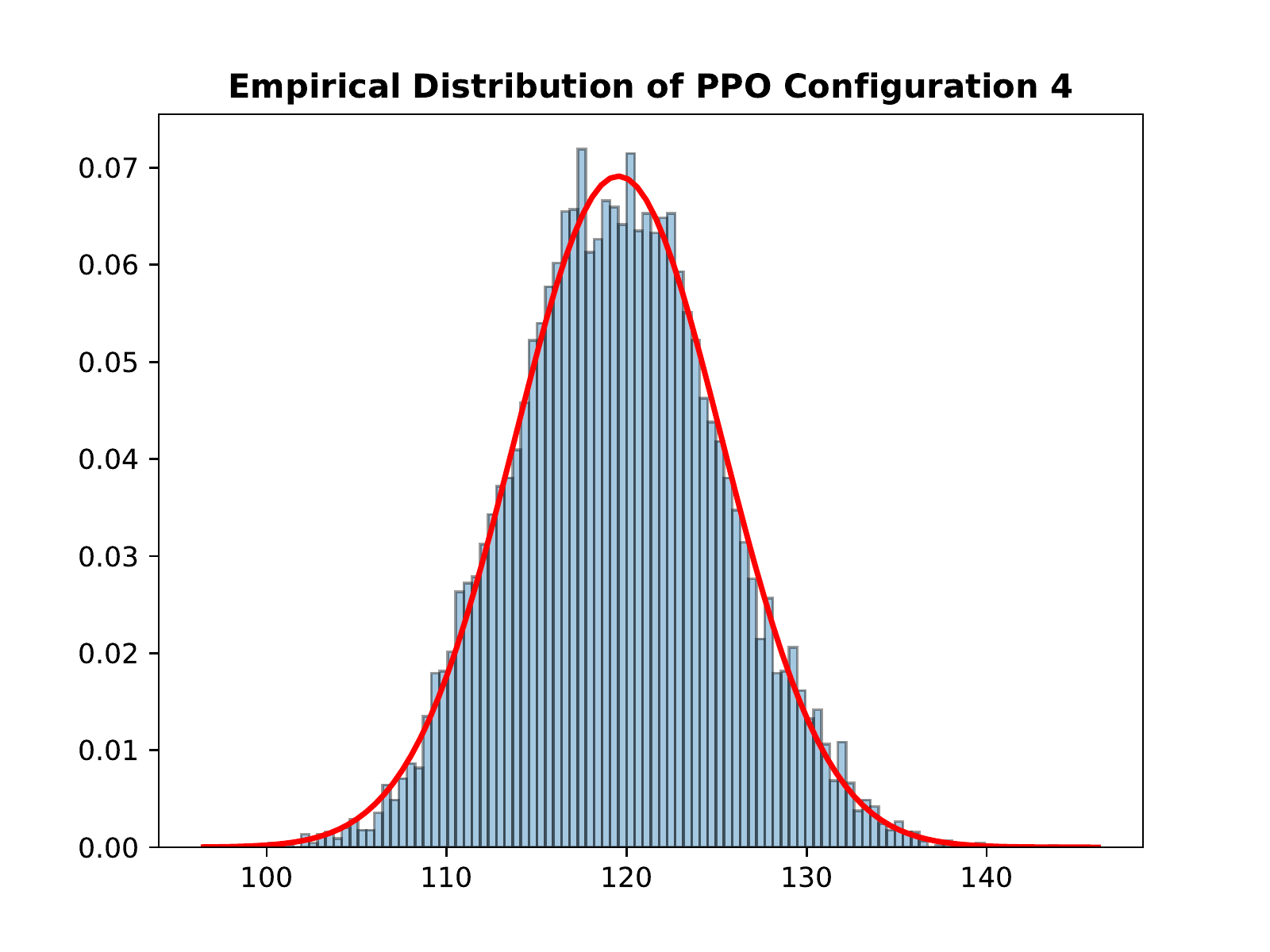}
    \end{minipage}
    \begin{minipage}{0.49\textwidth}
        \centering
        \includegraphics[width=1.0\textwidth,trim={10pt 0 35pt 0},clip]{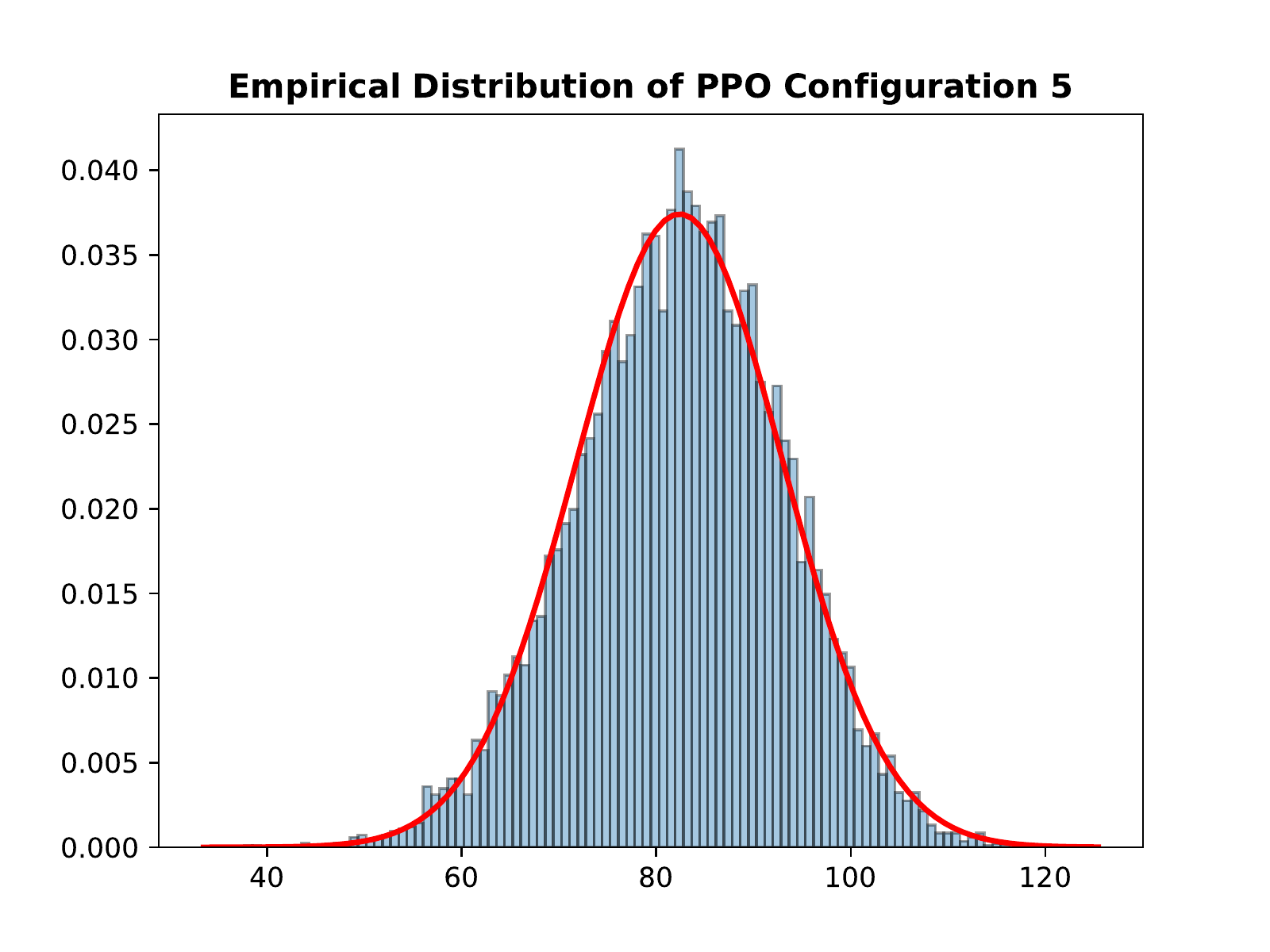}
    \end{minipage}
    \caption{\trponormalcaption}\label{fig:empirical_distributions}
\end{figure}

\def\num#1{\numx#1}\def\numx#1e#2{{#1}\mathrm{e}{#2}}

\begin{table}[!ht]
      \centering
      \begin{tabular}{c|C{3cm}C{3cm}}
      \toprule
        \multirow{2}{3cm}{Hyper-parameter configuration}   & \multicolumn{2}{c}{Algorithms} \\
                                        & TRPO                          & PPO               \\
      \midrule
        1                               & Rejected \newline ${\scriptstyle(p = 0.0002)}$ & Failed to reject \newline ${\scriptstyle(p = 0.1604)}$   \\
        2                               & Rejected \newline ${\scriptstyle(p = \num{5.17e-33})}$ & Failed to reject  \newline ${\scriptstyle(p = 0.1212)}$    \\
        3                               & Rejected \newline ${\scriptstyle(p = 0.0034)}$ & Rejected  \newline ${\scriptstyle(p = \num{7.24e-17})}$    \\
        4                               & Rejected \newline ${\scriptstyle(p = \num{1.41e-6})}$ & Rejected  \newline ${\scriptstyle(p = 0.0012)}$    \\
        5                               & Rejected \newline ${\scriptstyle(p = 0.0005)}$ & Rejected  \newline ${\scriptstyle(p = \num{4.69e-09})}$    \\
      \bottomrule
      \end{tabular}
      \caption{\textbf{Hypothesis testing for normality:} The table depicts the decision based on the normality test and the resulting $p$--values to support those decisions. Formally, the test rejects 8 out of 10 hypotheses, thus failing to reject 2 of them. From this we derive that none of the 10 empirical distributions are normally distributed. }
      \label{tab:normality_hypothesis_tests}
\end{table}

  \newpage
  \section{Sources of Non-Determinism in Machine Learning}\label{app:non_determinism}
  In \gls*{ml}, in contrast to studies in e.g. chemistry and social sciences, we have the ability to create code (methods) and datasets (observations) that, if open-sourced, other scientists can use to \textit{reproduce} the results from the original published research. From this overly simplified statement it should seem that reproducibility shouldn't be a problem in \gls*{ml}. There do, however, exists a reproducibility crisis in the field of \gls*{ml}. This crisis is often caused by the nature of \gls*{ml} but also non-rigorous testing approaches and sparsely documented hyperparameters have part in the crisis \cite{sandve2013ten, henderson2018deep}.

\subsection{Common Causes of Non-Determinism}
Some of the most important causes for general non-determinism in \gls*{ml} are listed below. It should be noted that this list assumes that we are attempting to \textit{reproduce} an \gls*{ml} experiment, hence these problems can occur \textit{even} when we have obtained the original code and data.

\begin{itemize}
    \item \textbf{GPU:} GPU floating point calculations -- made through Nvidia's neural network library; CuDNN -- are not guaranteed to generate the bit-wise reproducibility across different GPU versions and architectures, but \textit{should} generate the same bit-wise results across runs when executed on GPUs with the same architecture and number of SMs\footnote{\url{https://docs.nvidia.com/deeplearning/sdk/cudnn-developer-guide/index.html\#reproducibility}}.
    \item \textbf{Third-Party Libraries} Often, the libraries used are using other libraries which might in turn use stochastic processes needing a seed to a different random number generator.
\end{itemize}

\subsection{Non-Determinism in Deep Learning}
In addition to the general sources of non-determinism we here present sources specific for \gls*{dl}.

\begin{itemize}
\item \textbf{Random Initialization of Weights:} Often, the layer weights of a neural network is initialized by sampling from a particular distribution to aspire faster convergence \cite{glorot2010understanding, he2015delving}. The initialization must be the same from run-to-run in order to expect \textit{same} results and not \textit{similar} results.

\item \textbf{Shuffling of the Datasets:} To avoid the optimization functions getting stuck in local minima, the training of neural networks often occurs by dividing the dataset into mini-batches. It is also shown that shuffling the data after each epoch reduces the bias between gradient updates making the model more general as it tends to overfit less.

\item \textbf{Random Sampling:} If we are in that luxurious position that we have to much data to reasonably work with we draw a random subsample from the dataset to train our model with. 

\item \textbf{Random Train/Test/Validation Splits:} When data availability is low the go-to validation method is $k$--fold cross validation where the dataset is stochastically split into two or three sets of data.

\item \textbf{Stochastic Attributes of the Hidden Layers:} One of the most often used techniques for preventing overfitting is dropout \cite{srivastava2014dropout} which is inherently random during the training process. To \textit{reproduce} or \textit{repeat} the training process of a certain neural network originally trained with dropout one must know which neurons are excluded at what times through the original training process. 
\end{itemize}

\subsection{Non-Determinism in (Deep) Reinforcement Learning}

Here we describe the sources of non-determinism related to both \gls*{rl} and \gls*{drl}.
\begin{itemize}
    \item \textbf{Environment:} Especially when dealing with real-world robotic \gls{rl}, sensor delays, etc. supports the statement that our world is stochastic.
    \item \textbf{Network initialisation:} As in \gls{dl} the initialisation of the neural networks' weights are a stochastic process and must thus be controlled for to ensure reproducibility.  
    \item \textbf{Minibatch sampling:} Several algorithms within \gls*{drl} includes sampling randomly from the training data and from replay buffers \cite{schaul2015prioritized}.
\end{itemize}

Although some of the aforementioned sources of non-determinism can be successfully managed, many researchers does not control, or report how they have controlled for the non-determinism in their experiments. We find it necessary to advocate for presetting seeds for the random processes in the code, in order to remove or at least reduce the stochasticity of the reported experiments.

%===============================================================================

\end{document}